\documentclass[10pt,twocolumn,letterpaper]{article}

\usepackage{iccv}
\usepackage{times}
\usepackage{epsfig}
\usepackage{graphicx}
\usepackage{amsmath}
\usepackage{amssymb}
\usepackage{caption}
\usepackage{microtype}
\usepackage{booktabs}
\usepackage{multirow}
\usepackage{xspace}
\usepackage{tabularx}

\usepackage{caption}
\usepackage[breaklinks=true,bookmarks=false]{hyperref}

\iccvfinalcopy
\def\iccvPaperID{3383} %

\newcommand{\rev}[1]{{\textcolor{black}{#1}}}

\newcommand\blfootnote[1]{%
  \begingroup
  \renewcommand\thefootnote{}\footnote{#1}%
  \addtocounter{footnote}{-1}%
  \endgroup
}

\newenvironment{packed_item}{
\begin{itemize}
  \setlength{\itemsep}{1pt}
  \setlength{\parskip}{2pt}
  \setlength{\parsep}{0pt}
}{\end{itemize}}

\newcommand{\whitetxt}[1]{{\color{white}#1}\normalfont}

\newbox\jsavebox
\newcommand{\jsubfig}[2]{%
	\sbox\jsavebox{#1}%
	\parbox[t]{\wd\jsavebox}{\centering\usebox\jsavebox\\#2}%
	}

\long\def\ignorethis#1{}

\begin{document}

\title{Towers of Babel: Combining Images, Language, and 3D Geometry for \\ Learning Multimodal Vision}

\author{
Xiaoshi Wu$^{1*}$ \ \ \
Hadar Averbuch-Elor$^{2*}$ \ \ \
Jin Sun$^{2}$ \ \ \
Noah Snavely$^{2}$
\\[2mm]
\vspace{1em}
$^1$Tsinghua University \ \ \
$^2$Cornell Tech, Cornell University
}

\twocolumn[{%
\renewcommand\twocolumn[1][]{#1}%
\maketitle
\thispagestyle{empty}
\begin{center}
\vspace{-4pt}
  \centering
      \includegraphics[width = 0.99\linewidth]{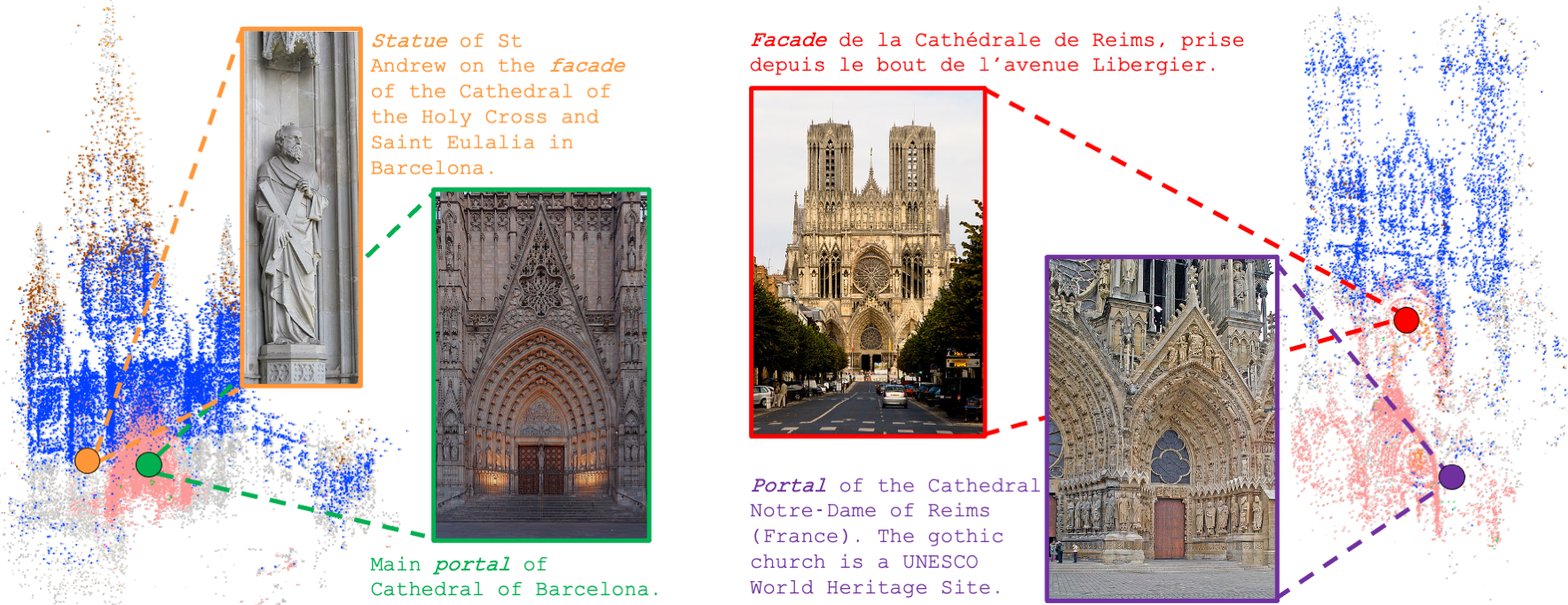}
      \vspace{-3pt}
    \captionof{figure} {Our \emph{WikiScenes} dataset combines 3D reconstructions, %
    images, and language descriptions for dozens of landmarks, like the Barcelona and Reims Cathedrals pictured above. WikiScenes enables new tasks that combine different modalities, such as associating semantic concepts like ``portal'', ``facade'', and ``tower'' (colored in pink, blue and brown, respectively) with 3D structure across 
    all cathedrals.
    }
  \label{fig:teaser}
\end{center}
}]

\blfootnote{$^*$: indicates equal contribution.}

\begin{abstract}
 The abundance and richness of Internet photos of landmarks and cities has led to significant progress in 3D vision over the past two decades, including automated 3D reconstructions of the world's landmarks from tourist photos.
However, a major source of information available for these 3D-augmented collections---namely \textit{language}, e.g., from image captions---has been virtually untapped.
In this work, we present \textit{WikiScenes}, a new, large-scale dataset of landmark photo collections that contains descriptive text in the form of captions and hierarchical category names. WikiScenes forms a new testbed for multimodal reasoning involving images, text, and 3D geometry. 
We demonstrate the utility of WikiScenes for learning semantic concepts over 
images and 3D models. 
Our weakly-supervised framework connects images, 3D structure, and semantics---utilizing the strong constraints provided by 3D geometry---to associate semantic concepts to image pixels and 3D points.\footnote{\url{https://www.cs.cornell.edu/projects/babel/}}

\end{abstract}

 \section{Introduction}

Internet photos capturing tourist landmarks around the world have driven research in 3D computer vision
for over a decade \cite{snavely2006photo,goesele2007multi,furukawa2010towards,agarwal2011building,schonberger2016structure,martinbrualla2020nerfw}.
Diverse photo collections of landmarks are unified by the underlying 3D scene geometry, 
despite the fact that a scene can look dramatically different from one image to the next due to varying illumination, alternating seasons, or special events.
This geometric anchoring can be exploited when learning a range of geometry-related vision tasks, such as novel view synthesis~\cite{meshry2019neural,li2020crowdsampling}, single-view depth prediction~\cite{li2018megadepth}, and relighting~\cite{yu2019inverserendernet,yu2020self}, that require large amounts of diverse training data. 
However, prior work on tourist photos of landmarks has focused almost exclusively on lower-level reconstruction tasks, and not on higher-level scene understanding or recognition tasks.

We seek to connect such 3D-augmented image collections to a new domain: \textit{language}.
Natural language is an effective 
way to describe
the complexities of the 3D world; 
3D scenes exhibit features such as compositionality and physical and functional relationships that are 
easily captured
by language.
For instance, consider the images of the Barcelona and Reims Cathedrals in Fig.~\ref{fig:teaser}. Cathedrals like these have common elements, such as facades, columns, arches, portals, domes, etc., that tend to be physically assembled in consistent ways across all 
cathedrals (and related buildings like basilicas). Using modern structure from motion methods, we can reconstruct 3D models of the world's cathedrals, but we cannot directly infer such rich semantic connections that exist \emph{across} all cathedrals. Such reasoning calls for methods that jointly consider language, images, and 3D geometry. %

However, despite impressive progress connecting images to natural language descriptions across tasks such as image captioning \cite{you2016image,lu2017knowing,anderson2018bottom} and visual grounding \cite{xiao2017weakly,hong2019learning,gupta2020contrastive}, little attention has been given to joint analysis of 3D vision and language.
In this work, we facilitate such multimodal analysis with a new framework for creating 3D-augmented datasets from Wikimedia Commons, a diverse, crowdsourced and freely-licensed large-scale data source. We use this framework to create \textbf{WikiScenes}, a new dataset that contains 63K paired images and textual descriptions capturing 99 cathedrals, along with their associated 3D reconstructions, illustrated in Fig.~\ref{fig:teaser}. WikiScenes enables a range of new explorations at the intersection of language, vision, and 3D.

We demonstrate the utility of WikiScenes for the specific task of mining and learning semantic concepts over collections of images and 3D models.
Our key insight is that while raw textual descriptions represent a weak, noisy form of supervision for semantic concepts,
the underlying 3D structure of scenes yields powerful physical constraints that grants robustness to data noise and can ground models.
In particular, we devise a novel 3D contrastive loss 
that leverages scene geometry 
to regularize learning of semantic representations. 
\rev{We also show that 3D scene geometry leads to improved vision-language models in a caption-based image retrieval task, where geometry helps in augmenting the training data with semantically-related samples.} %

In summary, our key contributions are:
\begin{packed_item}
    \item \textbf{WikiScenes}, a large-scale dataset combining language, images, and 3D models, which %
    can facilitate research that jointly considers these modalities.
    \item A contrastive learning method 
    for learning 
    semantic image representations leveraging 3D models. 
    \item Results that demonstrate that our proposed model can associate semantic concepts with images and 3D models, even for never-before-seen locations.
\end{packed_item}

 \section{Related Work}
\noindent \textbf{Joint analysis of 3D and language.}
We have recently seen pioneering efforts to jointly analyze 3D and language. Chen \etal~\cite{chen2018text2shape} learn a joint embedding of text and 3D shapes belonging to the ShapeNet dataset \cite{chang2015shapenet}, and demonstrate these embeddings on 
text-to-shape retrieval and text-to-shape generation. Achlioptas \etal~\cite{achlioptas2019shapeglot} learn language for differentiating between shapes. To do so, they generate a dataset consisting of triplets of ShapeNet chairs with utterances distinguishing one chair from the other two. In contrast to these object-centric works,
Chen \etal~\cite{chen2019scanrefer} consider full 3D scenes. They construct a multimodal dataset for indoor scenes and localize 3D objects in the scene using natural language. We also consider 3D scenes, but in our case, the 3D scenes capture complex architectural landmarks, and their images and textual descriptions are gathered from Wikimedia Commons. %

\medskip
\noindent \textbf{Vision and language.} 
Many recent works connect images to natural language descriptions. Popular tasks include instruction following \cite{Anderson:18r2r,Misra:17instructions,Blukis:18drone}, visual question answering \cite{antol2015vqa,Fukui:16bilinearpoolvqa,Hu:17compnetqa,anderson2018bottom}, and phrase localization \cite{Mao2015:refexp,Yu:16refmscoco,Wang:16phraselocalize}.
However, 
prior work has shown that 
models combining vision and language often rely on simple signals or fail to jointly consider both modalities. For instance, visual question answering techniques often ignore the image content~\cite{agrawal2016analyzing}, and visually-grounded syntax acquisition methods essentially learn a simple noun classifier~\cite{kojima2020learned}.
We assemble Internet collections that are grounded to a 3D model, providing physical constraints that can better connect language and vision.

\begin{figure*} %
    \centering
    \jsubfig{\includegraphics[height=3.36cm]{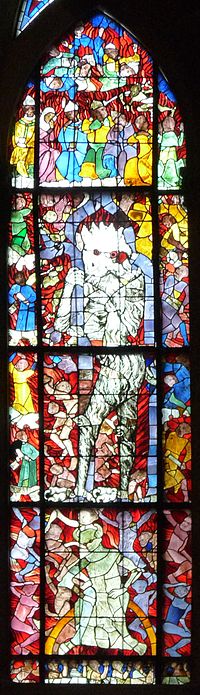}}{}
\jsubfig{\includegraphics[height=3.36cm]{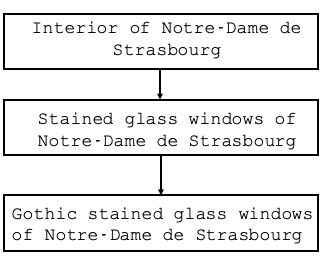}}{}
\hfill
\jsubfig{\includegraphics[height=3.36cm]{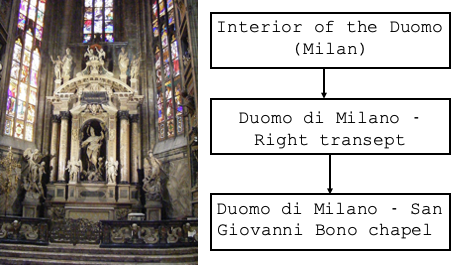}}{}
\hfill
\jsubfig{\includegraphics[height=3.36cm]{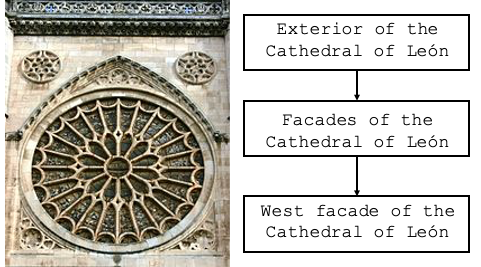}}{}
    \caption{Images paired with hierarchical WikiCategories from the root (top) to the leaf (bottom). 
   }
    \label{fig:tags}
\end{figure*}

\medskip
\noindent \textbf{Distilling information from Internet collections.}
Several works mine Internet collections capturing famous landmarks for objects \cite{gammeter2009know,philbin2008object},  events \cite{quack2008world}, or named parts~\cite{weyand2013discovering} using image clustering techniques. 
Other work analyzes camera viewpoints in large-scale tourist imagery to automatically summarize a scene~\cite{simon2007scene} or segment it into components~\cite{simon2008scene}.

Other prior work analyzes image content together with textual tags, geotags, 
and other metadata to organize image collections. Crandall \etal use image features and user tags from geotagged Flickr images to discover and classify world landmarks~\cite{crandall2009mapping}. 
3D Wikipedia analyzes textual descriptions of tourist landmarks, leveraging photo co-occurrences to annotate specific 3D models like the Pantheon~\cite{russell20133d}. In contrast to the above methods, which operate on each location in isolation, our work aims to discover semantic concepts spanning a whole category of locations, such as all the world's cathedrals. We further use a contrastive learning framework for detecting these concepts in unseen landmarks.

\ignorethis{
Conceptually, "3D Wikipedia" by Russell \etal~\cite{russell20133d} is closely related to our work. However, while we are interested in a joint learning framework over a category of landmarks, Russell \etal analyze a single landmark, leveraging photo co-occurrences for annotating each landmark. 

Our own work was among the first to tap into the vast visual knowledge encoded in Internet collections capturing tourist %
via our Photo Tourism project \cite{snavely2006photo}. In that work, we devised methods for building 3D models from noisy, unstructured Internet collections collections. We later scaled up our 3D algorithms to work on entire cities \cite{agarwal2011building}. More recently, we have shown the benefit of using these collections as diverse training data for a variety of vision tasks, such as single-view depth prediction \cite{li2018megadepth} and neural re-rendering~\cite{meshry2019neural}. 

As we analyze in our recent work, multi-modal models combining vision and language often rely on simple signals, and do not exhibit the complex reasoning we would like them to acquire \cite{kojima2020learned}. We believe that using Internet collections that are grounded to a 3D model allows for more holistic models which jointly analyze vision, language and 3D. 
}
 \section{The WikiScenes Dataset}

Our WikiScenes dataset consists of paired images and language descriptions capturing world landmarks and cultural sites, with associated 3D models and camera poses. WikiScenes is derived from the massive public catalog of freely-licensed crowdsourced data available in Wikimedia Commons,\footnote{\url{https://commons.wikimedia.org}} 
which contains a large variety of images with captions and other metadata.
Within Wikimedia Commons, landmarks are organized into a hierarchy of semantic categories. In this work, we focus on cathedrals as a showcase of our framework, although our methodology is general and can be applied to other types of landmarks. We will also release companion datasets featuring mosques and synagogues. %

To create WikiScenes, we first assembled a list of cathedrals using prior work on mining landmarks from geotagged photos~\cite{crandall2009mapping}. Each cathedral corresponds to a specific \emph{category} on Wikimedia Commons, at which is rooted a hierarchy of sub-categories that each contain photos and other relevant information. 
We refer to a Wikimedia Commons category as a \emph{WikiCategory}. 
For example, ``Cath\'{e}drale Notre-Dame de Paris''\footnote{\url{https://commons.wikimedia.org/wiki/Category:Cath\%C3\%A9drale_Notre-Dame_de_Paris}} is the name of a WikiCategory corresponding to the Notre Dame Cathedral in Paris. 
It has a descendent WikiCategory called ``Nave of Notre-Dame de Paris''\footnote{\url{https://commons.wikimedia.org/wiki/Category:Nave_of_Notre-Dame_de_Paris}} that features photos of the nave (a specific region of a cathedral interior), as well as yet more detailed WikiCategories.
Each landmark's root WikiCategory node contains ``Exterior'', ``Interior'' and ``Views'' subcategories. %
We download all images and associated descriptions under these subcategories. %
We extract two forms of textual descriptions for each image:
\begin{packed_item}
  \item \textbf{Captions} associated with images, describing the image using free-form language (Figure~\ref{fig:teaser}).
  \item The \textbf{WikiCategory hierarchy} %
  associated with each image. %
  Example hierarchies are shown in Figure \ref{fig:tags}.
\end{packed_item}
Because data stored 
in Wikimedia Commons is not specific to any single language edition of Wikipedia, our dataset contains text in numerous languages, 
allowing for future multilingual tasks like learning of cross-lingual representations~\cite{suris2020globetrotter}. However, 
one can also train with text from a single language, such as English. %
Overall, WikiScenes contains 
$63$K images of cathedrals with textual descriptions. %

\begin{figure*}
    \centering
\jsubfig{\includegraphics[height=4.55cm]{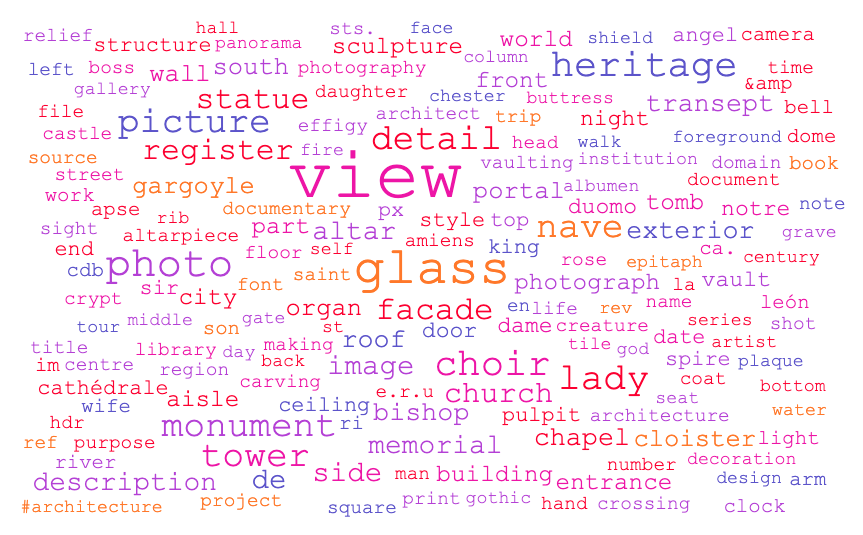}}{Candidates \emph{from captions}}
\jsubfig{\includegraphics[height=4.55cm]{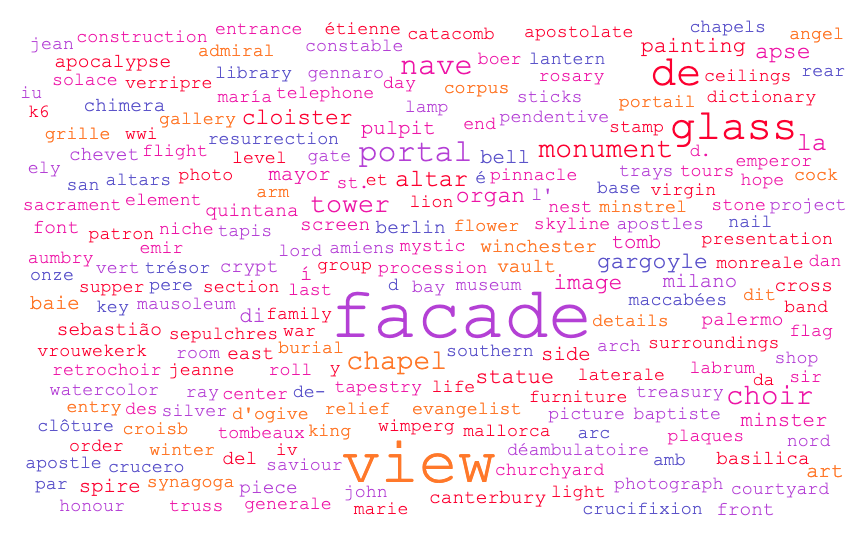}}{Candidates \emph{from the leaf categories}}
\jsubfig{\hspace{3pt} \includegraphics[height=4.45cm]{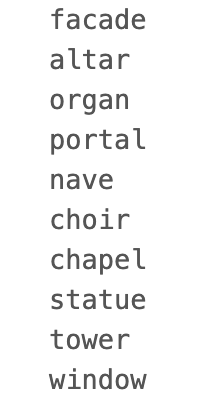} \hspace{3pt}}{Distilled concepts
}

    \caption{We visualize the raw text captured in WikiScenes captions (left) and leaf tags (center). Larger words are more frequent in the dataset. Our distilled concepts, obtained according to the algorithm described in Sec.~\ref{sec:distill}, are listed on the right.
    }
    \label{fig:cloud}
\end{figure*}

We integrate these Wikimedia Commons--sourced images with 3D reconstructions of landmarks built using COLMAP \cite{schonberger2016structure}, a state-of-the-art SfM system that reconstructs camera poses and sparse point clouds. For each 3D point in the reconstructed scene, we track all its associated images and corresponding pixel locations. In total, $26$K images  of cathedrals were successfully registered in 3D. Example 3D reconstructions are shown in Figure~\ref{fig:teaser}.

\medskip
\noindent \textbf{Dataset statistics.} %
WikiScenes is assembled from $99$ cathedrals spanning five continents 
and 23 countries. The languages most common in the captions are English ($45.8\%$), French ($11.1\%$) and Spanish ($10.9\%$). 
The Notre Dame Cathedral in Paris represents the largest subset, with 5,700 images-description pairs. The median number of words in a caption is seven; the average is significantly higher as some captions contain detailed excerpts about their associated landmark. $8.39\%$ of all captions contain at least one spatial connector,\footnote{The spatial connectors we consider are: above, over, below, under, beside, behind, from, towards, left, right, east and west.
} suggesting that our captions describe rich relationships between different parts of a structure. %
Please see the supplemental material for detailed distributions over attributes including language and collection size. 

\section{Mining WikiScenes for Semantic Concepts}
\label{sec:labels}
To demonstrate the 
semantic knowledge encoded in our dataset, we mine WikiScenes for semantic concepts associated with the Cathedral 
landmark category. 
While the raw textual descriptions are noisy, we show that we can distill a clean set of concepts by exploiting 
within-scene 3D constraints (Sec.~\ref{sec:distill}). We then associate these concepts to images (Sec.~ \ref{sec:association}), and 
show that these concepts can be used to train neural networks to visually recognize these concepts.

\subsection{Distilling semantic concepts}
\label{sec:distill}
To determine a set of candidate concepts, we first assemble a list of all nouns found in the leaf nodes of the WikiCategories, hereby denoted as the \emph{leaf categories}, as empirically we found that the leaf categories are most representative of the image content. Since we are interested in a list of \emph{abstract} concepts and not in detecting specific places and objects, we filter out nouns detected as entities using an off-the-shelf Named Entity Recognition (NER) tagger \cite{qi2020stanza}.  
Figure~\ref{fig:cloud} (middle) visualizes the initial candidate list as a word cloud (more frequent words appear larger).
As the figure illustrates, this list contains nouns that indeed describe semantic regions in the ``Cathedral'' category, but also contains many outliers, or nouns  
not specifically related to the ``Cathedral'' category, such as ``view'' or ``photograph''. 

As an alternative, we can also extract nouns directly from the captions (Figure \ref{fig:cloud}, left). This results in a noisier list, %
as the captions are generally longer 
with more diverse and detailed descriptions.
In addition, leveraging category names leads to more images with noun descriptions---over $56$K images have at least one noun in their leaf category, whereas only $22$K images have an English caption with a noun.

To distill a clean set of semantic concepts from the initial list, we identify and select concepts that pass two tests: they are (1) \emph{well-supported} in the collection (i.e., they occur frequently in the textual descriptions) and (2) \emph{coherent}, in the sense that they consistently reference identical or visually similar elements.
While well-supported concepts can be determined by simple frequency measurements, coherence is more difficult to assess from noisy Internet images and their descriptions. However, because these images are physically grounded via a 3D model, we can measure coherence in 3D.

For each candidate concept, e.g., ``facade'', we construct multiple visual adjacency graphs (one per landmark) over the images associated with that concept. Note that an image can be associated with multiple concepts, according to the nouns detected in its leaf category. For each graph, nodes $v \in V$ correspond to images and two images are connected by an edge $e \in E$ if they share at least $K$ common keypoints in the 3D model (where $K$ is empirically set to $10$). 
We are interested in measuring the degree to which the images of the candidate concept are clustered together in 3D. Therefore, for each landmark $\ell$, we compute the graph density:
\begin{equation}
\rho^\ell = \frac{2|E|}{|V|(|V|-1)}.
\end{equation}
The coherence of the candidate concept is measured as the average graph density $\rho$, obtained by taking the average over all corresponding landmark graphs with at least $10$ nodes.

Finally, candidate concepts that appear in at least $25$ landmarks (roughly a quarter of the ``Cathedral'' category) and have a coherency score $\rho \geq 0.08$ are added to our distilled set (Figure \ref{fig:cloud}, right).

\subsection{Associating images with distilled concepts}
\label{sec:association}
Although the distilled set of semantic concepts is constructed only from text appearing in the leaf categories, we utilize both the image captions and leaf categories when generating labels: an image is associated with a concept if the concept is present either in the caption or in its leaf categories.
An image can be associated with multiple concepts.

One exception is that text often includes concepts that are spatially related to the main concept present in an image using \emph{spatial} connectors such as ``beside'', ``next'', ``from'', ``towards''. For example, an image associated with the text ``nave looking towards portal'' should be associated with ``nave'', but not necessarily with ``portal''. Hence, we do not associate concepts with images if the concept appears anywhere after a spatial connector.

 \section{Learning Semantic Representations}
\label{sec:learning}

\begin{figure}
    \centering
    \includegraphics[width = 0.99\columnwidth]{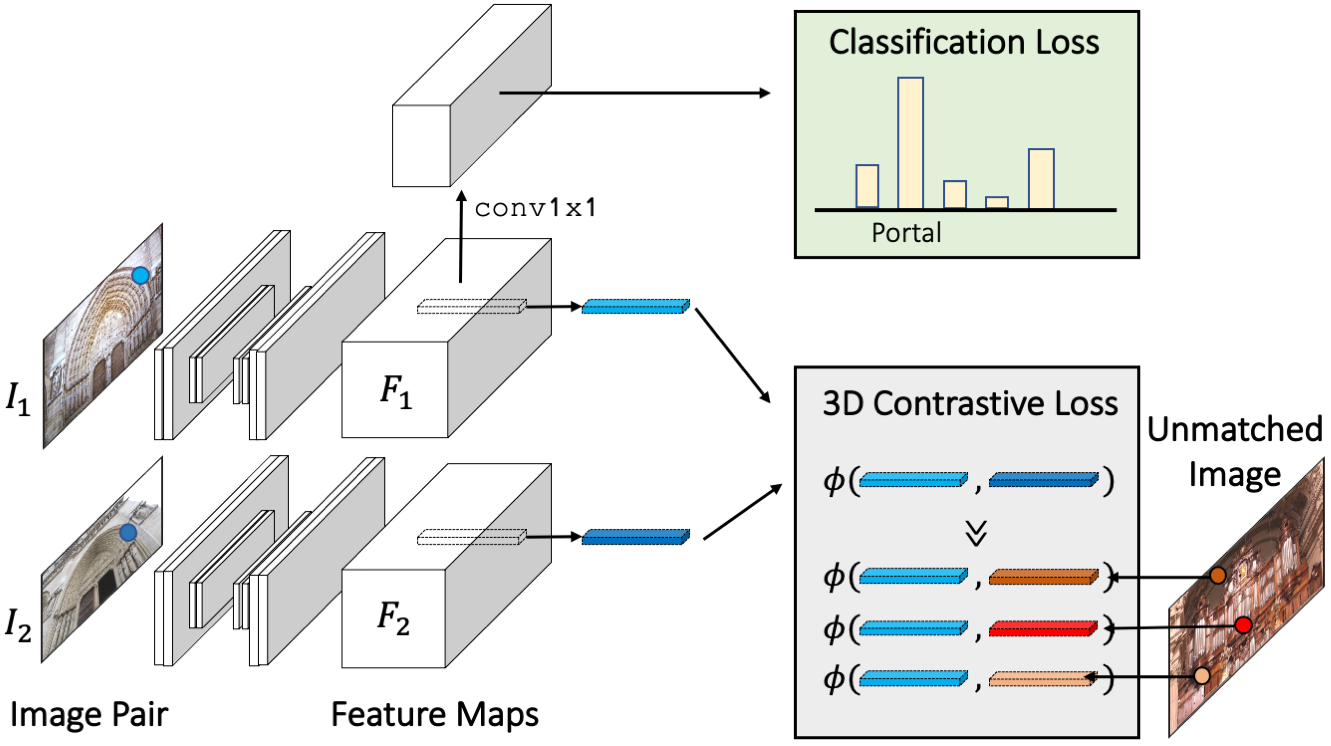}
    \caption{Overview of our contrastive learning framework. Given an image pair with shared keypoints (left), we jointly train a model to classify the images into one of the $C$ concepts from the learned score maps and to output a higher similarity for pixels mapping to the same point in 3D (in blue). Negative pairs are constructed by sampling non-corresponding points from other images in the batch.} 
    \label{fig:overview}
\end{figure}

WikiScenes can be used to study a range of different problems. Here, we focus on semantic reasoning over 2D images and 3D models.
In the previous section, we proposed a technique for discovering semantic concepts and associating these with images in WikiScenes. Now, we show how these image-level pseudo-labels can provide a supervision signal for learning semantic feature representations over an entire category of landmarks. %

We seek to learn pixel-wise representations (in contrast to whole-image representations), because we wish to easily map knowledge from 2D to 3D and vice versa. We would also like our learned representations to be semantically meaningful. In other words, our distilled concepts should be identifiable from these pixel-wise representations. To this end, we devise a contrastive learning framework that computes a feature descriptor for every pixel in the image. We also show how our trained model can be directly utilized to estimate feature descriptors for 3D points through their associated images.

\subsection{Training objectives}

Our training data consists of image pairs $(I_1,I_2)$ with shared keypoints, obtained from the corresponding SfM model. We use convolutional networks with shared weights to extract dense feature maps $F_1$ and $F_2$ whose width and height match those of the original images. For simplicity of notation, we assume both images have dimensions $w \times h$. To train a feature descriptor model with such data, we propose to use two complimentary loss terms: a novel 3D contrastive loss that utilizes within-scene physical constraints and a classification loss (Figure \ref{fig:overview}).

\medskip
\noindent \textbf{3D contrastive loss.}
We design a new 3D contrastive loss to encourage within-scene consistency, such that pixels from different images corresponding to the same 3D point should have similar features. \rev{This is unlike prior works on contrastive learning that use handcrafted data augmentations~\cite{chen2020simple,he2020momentum} or synthetic images~\cite{park2020contrastive} to generate positive pairs---in our case the positive pairs are 2D pixels that are projections of the same point in 3D.}
This loss 
relates images with different characteristics, such as lighting and scale, %
allowing to better focus on semantics and providing higher robustness against such nuisance factors. 

Our learning method works as follows: %
For each point $p$ in $I_1$ corresponding to point $p^+$ in $I_2$ (i.e., they are both projections of the same 3D point $P$), we formulate a contrastive loss to maximize the mutual information between their descriptors $F_1(p)$ and $F_2(p^+)$. 
We consider a noise contrastive estimation framework \cite{oord2018representation}, consisting of the positive pair $(p,p^+)$ and $m$ negative pairs $\{(p, p_i^-)\}$:
\begin{equation}
\mathcal{L}_\mathsf{3D} = -\log \left[\frac{e^{\phi (p,p^+)} }
{ e^{\phi (p,p^+)} + 
\sum_{i=1}^m e^{\phi (p,p_i^-)} } \right],  
\end{equation}
where the similarity $\phi(p,p^*)$ is computed as the dot product of feature descriptors scaled by a temperature $\tau$:
\begin{equation}
    \phi (p, p^*)=  F_1(p) \cdot F_2(p^*) \mathbin{/} \tau.
\end{equation}
This loss can be 
interpreted as the log loss of a ($m+1$)-way softmax classifier that learns to classify $p$ as $p^+$.
\rev{The points $p_i^-$ are sampled uniformly from other images in the same batch.}
To avoid collapsing the feature space, we normalize all feature descriptors to unit length.

\medskip
\noindent \textbf{Semantic classification loss.} For each image we also compute a semantic classification loss. Given $C$ unique semantic concepts, we obtain unnormalized score maps from the feature descriptors using a simple \texttt{conv1x1} layer. That is, we map the $[K\times h \times w]$ feature descriptor tensor to a $[C\times h \times w]$ score map tensor, where each slice corresponds to one of the semantic concepts.

Following the design proposed by Araslanov \etal~\cite{araslanov2020single}, we add a background channel and compute a pixel-wise $\texttt{softmax}$ to obtain normalized score maps $y_{\textsf{pix}} \in \mathbb{R}^{C+1\times h \times w}$ and image-level classification scores $y \in \mathbb{R}^C$, derived from the %
score maps using the method of Araslanov \etal. Our semantic classification loss is defined as
\begin{equation}
\mathcal{L}_\mathsf{cls} = \mathcal{L}_\mathsf{cls}^\mathsf{im} + \mathcal{L}_\mathsf{cls}^\mathsf{pix},   
\end{equation}
where $\mathcal{L}^\mathsf{cls}_\mathsf{im}$ is a classification loss on image-level scores $y$ %
and $\mathcal{L}^\mathsf{cls}_\mathsf{pix}$ is a self-supervised semantic segmentation loss over pixel-wise predictions (where high-confidence pixel predictions serve as self-supervised labels). For both training and evaluation, we only consider images labeled with a single concept and the one-hot class label is set according to our pseudo image label.
We minimize a cross-entropy loss for both image-level and pixel-level predictions.

\subsection{Inference}
\label{sec:inference}

\definecolor{facade}{RGB}{0, 0, 255}%
\definecolor{window}{RGB}{255,128, 0}%
\definecolor{chapel}{RGB}{0,153, 0}%
\definecolor{organ}{RGB}{255, 0, 0} %
\definecolor{nave}{RGB}{101,0,204}%
\definecolor{tower}{RGB}{102, 51, 0}%
\definecolor{choir}{RGB}{255,51, 255}%
\definecolor{portal}{RGB}{255,153, 153}%
\definecolor{altar}{RGB}{236, 227, 102} %
\definecolor{statue}{RGB}{57,218,250}%

\begin{figure}
    \centering
\jsubfig{\includegraphics[height=3.5cm]{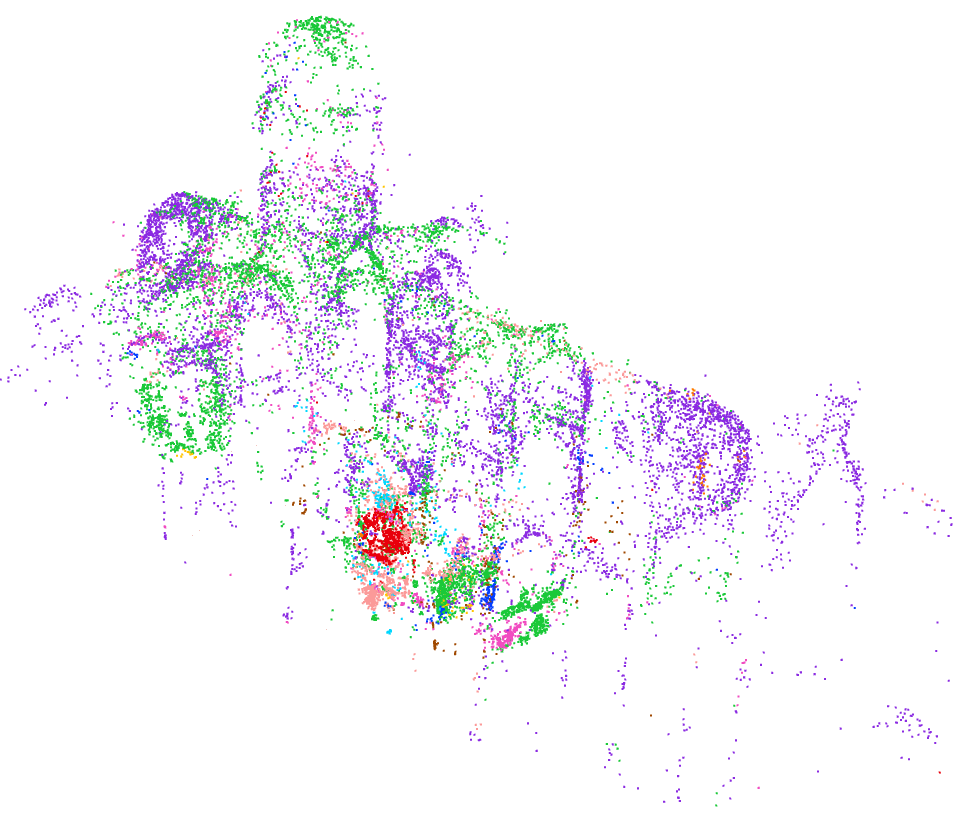}}{Baseline (w/o $\mathcal{L}_{\mathit{3D}}$)}
        \hfill
\jsubfig{\includegraphics[height=3.5cm]{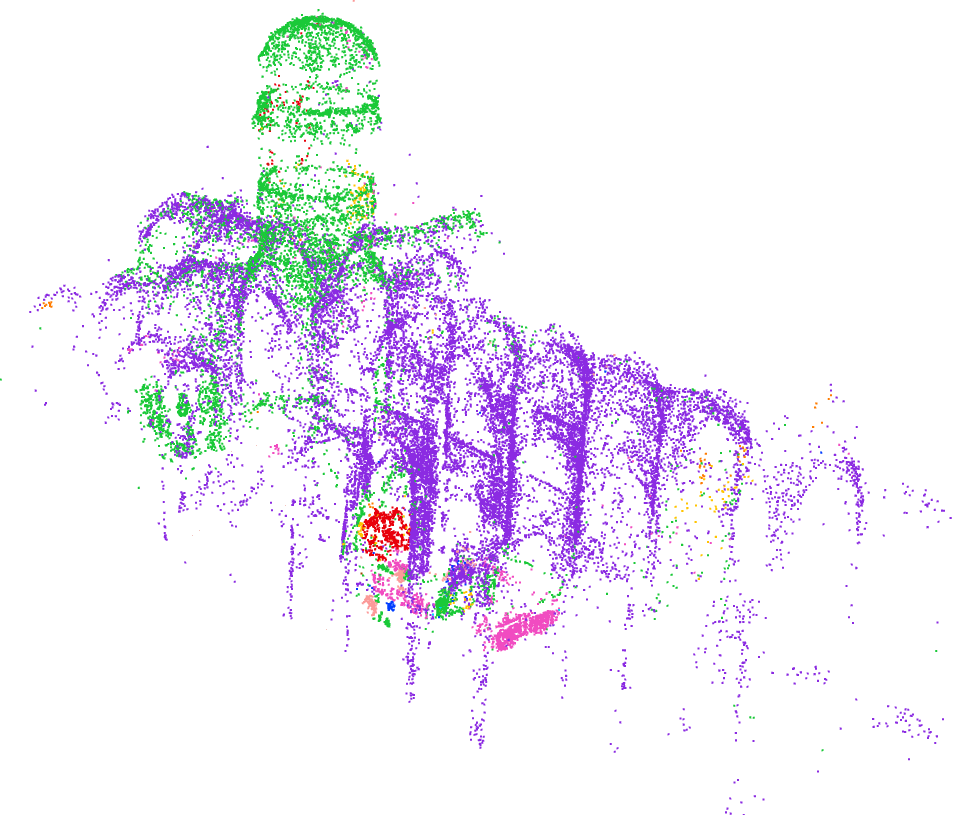}}{Ours}
    \caption{Segmenting an unseen 3D model of the interior of the Aachen Cathedral in Germany.
    Color legend: \color{nave}{\emph{nave}}, \color{chapel}{\emph{chapel}}, \color{organ}{\emph{organ}}, \color{altar}{\emph{altar}},  \color{choir}{\emph{choir}}, \color{statue}{\emph{statue}}, \color{portal}{\emph{portal}}, \color{facade}{\emph{facade}}.}
    \label{fig:coloring}
\end{figure}

At inference time, we can feed an image from a never-before-seen location into our model (Figure \ref{fig:coloring}). 
The model outputs pixel-wise feature descriptors and probability scores over the semantic concepts for each pixel (and also for the full image, if that is desired). 
We follow the procedure described in \cite{araslanov2020single} to extract 2D segmentations. To output probability scores for a 3D point in the scene, we process all the images associated with this 3D point. The feature descriptors of all its 2D projections are averaged, and we process this average descriptor to output its associated probability scores. We associate a 3D point with one of the semantic concepts if its corresponding confidence score is greater than $\varphi = 0.5$. %

 \section{Evaluation}
\label{sec:results}
In this section, we demonstrate our ability to learn semantic concepts shared across multiple landmarks. Specifically, we seek to answer the following questions:
\begin{packed_item}
    \item Is WikiScenes suitable for learning these concepts? 
    \item How important is the 3D contrastive loss? 
    \item How well does our model generalize to Internet photos from never-before-seen locations?
\end{packed_item}
We perform a variety of experiments to evaluate performance across multiple tasks, including classification, segmentation, and a caption-based image retrieval task that operates on the raw captions directly.
These experiments are complemented with a visual analysis that highlights the unique characteristics and challenges of our data.

\subsection{Implementation details}
\noindent \textbf{Data.} Out of the 99 WikiScenes landmarks, 70 landmarks contain sufficient labeled data that can serve for training and evaluating our models (images are labeled using the approach described in Section \ref{sec:association}). 
We create a 9:1 split at the landmarks level, forming a test set for landmarks \emph{unseen} during training (WS-U). For the 63 landmarks in the training set, we create a 9:1 split at the images level, forming a test set for \emph{known} landmarks (WS-K) to evaluate how well our model can classify unseen images in familiar locations.
Overall, we use almost 9K labeled images for training, with balanced class frequencies across the ten semantic concepts.

\medskip
\noindent \textbf{Training.} We use a batch size of 32, corresponding to 16 image pairs. Only half of these are \emph{real} pairs with shared keypoints, as we also want to consider labeled images that are not associated with any 3D reconstruction, possibly due of a sparse sampling of views in these regions. Please refer to the supplementary for additional implementation details.

\newcolumntype{Y}{>{\centering\arraybackslash}X}
\newcolumntype{L}{>{\arraybackslash}X}
\begin{table*}[h]
\centering
\setlength{\tabcolsep}{2.8pt}
\def\arraystretch{0.8}
\begin{tabularx}{\textwidth}{lllcccccccccccccccccc}
\toprule
Test Set     & &Model &&  mAP &&  mAP$^\star$ &&           facade & window & chapel & organ & nave & tower & choir & portal & altar & statue    \\ \midrule
\multirow{2}{*}{WS-K} && Baseline (w/o $\mathcal{L}_{\mathit{3D}}$) && 70.8 && 77.7 && 87.2&	\textbf{89.2}&	60.2&	89.7&	\textbf{85.8}&	\textbf{64.1}&	61.5&	68.0&	50.0&	52.0 \\
    && Ours &&  \textbf{75.3} && \textbf{81.0} && \textbf{90.0}&	88.5&	\textbf{68.7}&	\textbf{90.7}&	85.7&	61.1&	\textbf{77.2}&	\textbf{76.5}&	\textbf{54.4}&	\textbf{59.9}\\
\midrule
\multirow{2}{*}{WS-U} && Baseline (w/o $\mathcal{L}_{\mathit{3D}}$) && 48.3 && 64.0 && 71.0&	92.2&	10.7&	\textbf{57.3}&	71.0&	\textbf{53.4}&	43.6&	31.1&	25.8&	27.1\\
    && Ours && \textbf{52.0} && \textbf{67.3} && \textbf{77.7} &	\textbf{93.4}&	\textbf{16.5}&	49.4&	\textbf{77.3}&	46.1&	\textbf{44.1}&	\textbf{35.2} &	\textbf{39.9} &	\textbf{40.0}&\\
\bottomrule
\end{tabularx}
\vspace{-8pt}
\caption{Classification Performance. We report mean average precision (mAP, $^\star$ indicates averaging over all images, and not per class), and per distilled concept average precision (AP). Results of our model are compared against a model trained without our 3D contrastive loss. Performance is reported on images from known landmarks (WS-K) and unseen landmarks (WS-U). The best
results are highlighted in bold. %
}
\label{tab:classification}
\end{table*}

\subsection{Label quality} 
We assess the accuracy of our pseudo-labels by manually inspecting 50 randomly sampled training images for each concept, and identifying images with incorrect labels (i.e., the image does not picture all or part of the semantic concept). We found an accuracy greater than 98\%, 
suggesting that our pseudo-labels are highly accurate. We found that most errors are due to images that contain schematic diagrams or scans of the concept (and not natural images capturing it). Please refer to the supplementary material for visualizations of our training samples.

\subsection{3D-consistency guided classification}
Next we evaluate to what extent semantic concepts can be learned across a multitude of landmarks, and the effect of the 3D consistency regularization allowed by our dataset on classification results. 
We perform an image classification evaluation using our pseudo-labels, which we consider ground-truth for evaluation purposes. We compare our model to a model with the same architecture, trained using the semantic classification loss but without our 3D contrastive loss, hereby denoted as the \emph{baseline} model---adapted from the one proposed in Araslanov \etal~\cite{araslanov2020single}.

For each model, we report the overall mean average precision (mAP), as well as a breakdown of AP per concept, in Table \ref{tab:classification}. Results are reported for test images from known locations (WS-K) and unseen locations (WS-U).
As the table illustrates, our model outperforms the baseline model in most of the concepts and yields significant gains in mAP, boosting overall performance by $4.5\%$ and  $3.7\%$, 
when evaluating on WS-K and WS-U,
respectively (and an improvement of 3.3$\%$ when averaging across images, which is less affected by class frequencies). We provide additional experiments and an analysis of errors in the supplementary material. %

\ignorethis{
To further validate the effectiveness of our 3D loss, we take off-the-shelf networks dedicated for classification and repeat the experiment of testing classification performance with and without our 3D contrastive loss. Results are reported in Table \ref{tab:classification2}. As illustrated in the table, our 3D contrastive loss consistently boosts classification performance, even for off-the-shelf models.
}

\ignorethis{
\begin{figure}
    \centering
    \includegraphics[width = 0.85\columnwidth]{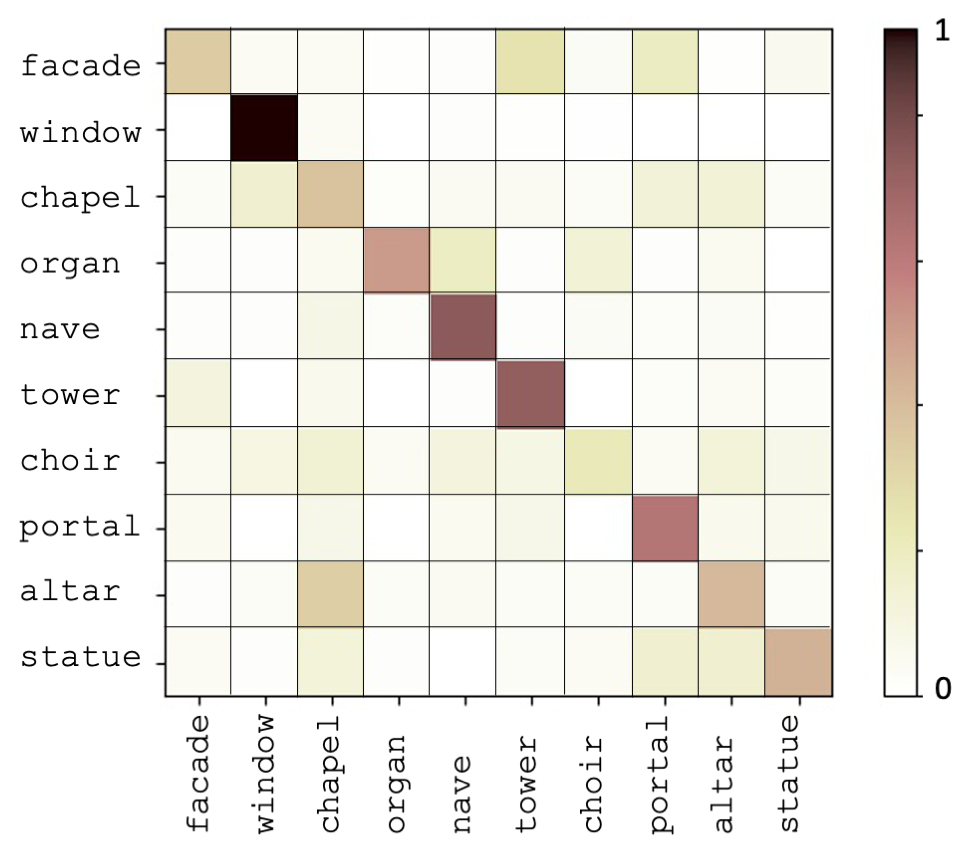} \\
    \caption{Confusion matrix of our classification model on unseen landmarks (the WS-U test set). Ground-truth concept labels correspond to rows, and predicted concept labels to columns. Each row is normalized such that a cell indicates the probability of a classification given the ground-truth label.}
    \label{fig:confusion}
\end{figure}

\medskip
\noindent \textbf{Error Analysis.}
Figure \ref{fig:confusion} shows a confusion matrix for our model.
We can see that many of the mistakes are understandable, given the hierarchical nature of our data. For example, both ``tower'' and ``portal'' are part of a ``facade'', and an ``altar'' is often placed inside a ``chapel''. As the confusion matrix and the quantitative evaluation illustrate, the ``choir'' concept exhibits lower generalization power and is often confused with other concepts found inside a cathedral. %

}

\begin{figure}
    \centering
     \rotatebox{90}{\whitetxt{ss} Input}
     \hfill
\jsubfig{
\includegraphics[height=1.65cm]{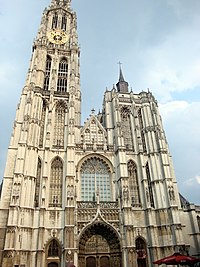}
\includegraphics[height=1.65cm]{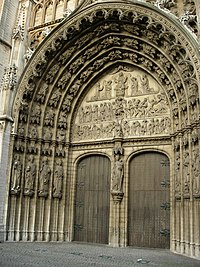}
\includegraphics[height=1.65cm]{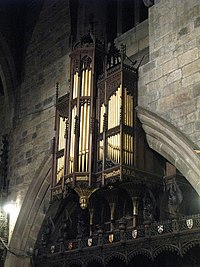}
\includegraphics[height=1.65cm]{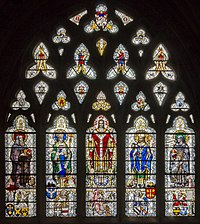}
\includegraphics[height=1.65cm]{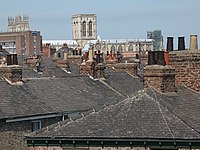}}{}
\\
\vspace{2pt}
        \rotatebox{90}{\whitetxt{s}w/o $\mathcal{L}_{\mathit{3D}}$}
        \hfill
\jsubfig{
\includegraphics[height=1.65cm]{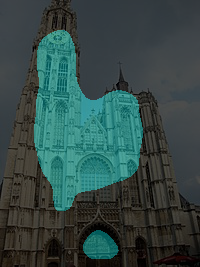}
\includegraphics[height=1.65cm]{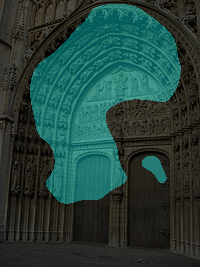}
\includegraphics[height=1.65cm]{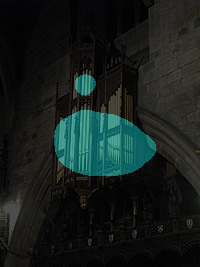}
\includegraphics[height=1.65cm]{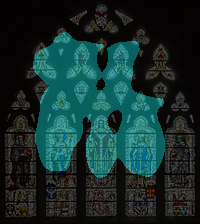}
\includegraphics[height=1.65cm]{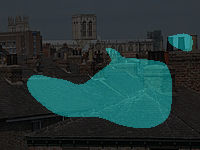}}{}
\\
\vspace{2pt}
        \rotatebox{90}{\whitetxt{ss} Ours }
        \hfill
\jsubfig{
\includegraphics[height=1.65cm]{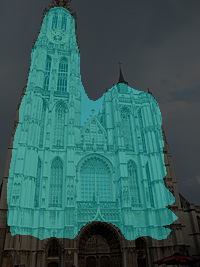}
\includegraphics[height=1.65cm]{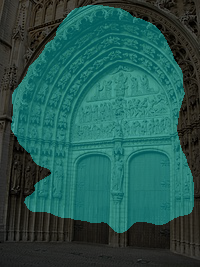}
\includegraphics[height=1.65cm]{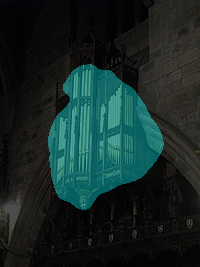}
\includegraphics[height=1.65cm]{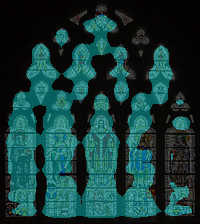}
\includegraphics[height=1.65cm]{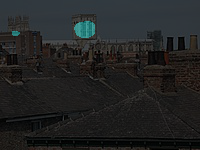}}{}
\\
\vspace{2pt}
        \rotatebox{90}{GT Masks }
        \hfill
\jsubfig{
\includegraphics[height=1.65cm]{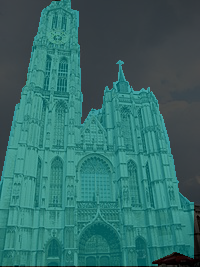}
\includegraphics[height=1.65cm]{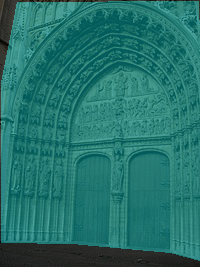}
\includegraphics[height=1.65cm]{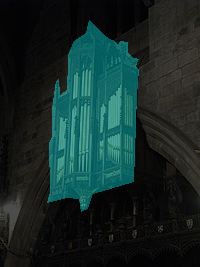}
\includegraphics[height=1.65cm]{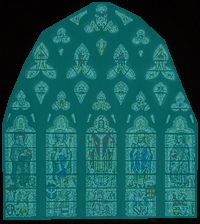}
\includegraphics[height=1.65cm]{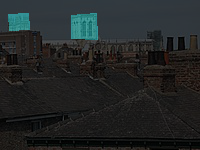}}{}
\caption{Segmenting images of unseen landmarks. Pixels are labeled \emph{facade}, \emph{portal}, \emph{organ}, \emph{window}, \emph{tower} from left to right.}
    \label{fig:seg}
\end{figure}

\subsection{2D and 3D segmentation}
Our framework learns pixel-wise features that are useful beyond classification, e.g., for producing segmentation maps for 2D images and 3D reconstructions. We show segmentation results for 2D images in Figure~\ref{fig:seg}
and for 3D reconstructions in Figures \ref{fig:teaser} and \ref{fig:coloring}. 

\rev{We manually label a random subset of test images (from unseen landmarks) for evaluating 2D segmentation performance and report standard segmentation metrics in Table \ref{tab:seg2d}. Specifically, we labeled $237$ images spanning six concepts that have definite boundaries
(\emph{facade}, \emph{portal}, \emph{window}, \emph{organ}, \emph{tower} and \emph{statue}). The distributions across these classes are roughly uniform (with 24-50 images per class).}

\rev{ 
Table \ref{tab:seg2d} shows the average intersection-over-union (IoU), precision and recall on the manually labeled set.
These results show that our 3D-contrastive loss boosts performance over all metrics. Precision is significantly higher ($81\%$ vs.\ $69\%$), with a modest increase in IoU and recall.    
}

\newcolumntype{Y}{>{\centering\arraybackslash}X}
\newcolumntype{L}{>{\arraybackslash}X}
\begin{table}[t]
\centering
\setlength{\tabcolsep}{2.8pt}
\def\arraystretch{0.8}
\begin{tabularx}{0.88\columnwidth}{lllcccccccccccccccccc}
\toprule
Model &&  IoU && Precision &&Recall     \\ \midrule
 Baseline (w/o $\mathcal{L}_{\mathit{3D}}$) && 25.4 && 68.6 && 28.4 \\
 Ours &&  \textbf{27.2} && \textbf{80.8} && \textbf{29.6}\\
\bottomrule
\end{tabularx}
\vspace{-8pt}
\caption{Image segmentation performance on manually labeled set. %
}
\label{tab:seg2d}
\end{table}

\ignorethis{
\begin{table*}[]
\begin{tabular}{lllllllll}
\toprule
Metric                   & model                               & avg.          & facade        & window        & organ         & tower         & portal        & statue        \\ \hline
\multirow{2}{*}{IoU(\%)} & Baseline(w/o 3D loss)               & 25.4          & 13.4          & \textbf{41.6} & 24.1          & 35.2          & 24.4          & \textbf{13.6} \\
                         & w/ L\_\{3D\} (inter-image sampling) & \textbf{27.2} & \textbf{18.8} & 40.9          & \textbf{25.6} & \textbf{39.5} & \textbf{26.8} & 11.7          \\ \hline
\multirow{2}{*}{Pr.(\%)} & Baseline(w/o 3D loss)               & 68.6          & 50.7          & 91.4          & 84.0          & 66.3          & 63.4          & 55.9          \\
                         & w/ L\_\{3D\} (inter-image sampling) & \textbf{80.8} & \textbf{92.1} & \textbf{96.9} & \textbf{92.2} & \textbf{67.3} & \textbf{74.9} & \textbf{61.0} \\ \hline
\multirow{2}{*}{Re.(\%)} & Baseline(w/o 3D loss)               & 28.4          & 15.4          & \textbf{43.4} & 25.3          & 42.8          & 28.4          & \textbf{15.2} \\
                         & w/ L\_\{3D\} (inter-image sampling) & \textbf{29.6} & \textbf{19.1} & 41.4          & \textbf{26.2} & \textbf{48.8} & \textbf{29.5} & 12.6          \\ 
                         \bottomrule
\end{tabular}
\end{table*}
}

\label{sec:seg}
To evaluate 3D segmentation performance, as it is difficult to obtain ground-truth 3D segmentations for large-scale landmarks whose reconstructions span thousands of points, we design two proxy metrics to assess both \emph{completeness} and \emph{accuracy} of the 3D  results. These metrics are (i) the fraction of ambiguous points $\theta_\varphi$, and (ii) the interior-exterior error $\Delta_\varphi$ (both dependent on the confidence scores $\varphi$).

The fraction of ambiguous points $\theta_\varphi$ quantifies the extent to which the model associates concepts to 3D points with high confidence. 
To compute $\theta_\varphi$, we measure the fraction of points that are not associated with a concept, averaging over all landmarks. For example, $\theta_\varphi=0$ means that for all points, the model's predictions across all images was consistent in 3D space, and thus the points were successfully associated with concepts, and $\theta_\varphi=1$ means that all points are ambiguous in their semantic association.

\begin{table}[t]
\centering
\setlength{\tabcolsep}{2.0pt}
\def\arraystretch{0.8}
\begin{tabularx}{1.0\columnwidth}{lcccccccccccc}
\toprule
 &           & \multicolumn{4}{c}{WS-K} &  & \multicolumn{4}{c}{WS-U} 
                          \\ 
\cmidrule(lr){3-6} \cmidrule(lr){8-11} 
Method && $\theta_{0.5}$& $\theta_{0.75}$ & $\Delta_{0.5}$ & $\Delta_{0.75}$ &&$\theta_{0.5}$ &$\theta_{0.75}$ & $\Delta_{0.5}$ & $\Delta_{0.75}$ \\ \midrule
Baseline && 0.50 & 0.78& \textbf{0.10}& 0.09 &&0.56 &0.83 & 0.13& 0.10\\ 
Ours && \textbf{0.43} & \textbf{0.70}& \textbf{0.10} & \textbf{0.06} &&\textbf{0.40} &\textbf{0.69}& \textbf{0.11}& \textbf{0.06}\\ 
\bottomrule
\end{tabularx}
\vspace{-8pt}
\caption{3D Segmentation Evaluation. %
Proxy metrics $\theta$ and $\Delta$ are described in detail in Section \ref{sec:seg}. For both metrics, lower is better.   }
\label{tab:3d}
\end{table}

Due to limited visual connectivity, 3D reconstructions of landmarks typically are broken into one or more exterior reconstructions and one or more interior reconstructions. Thus, we devise
the interior-exterior error $\Delta_\varphi$ to 
quantify to what extent concepts that should be exclusively found in either an exterior reconstruction or an interior reconstruction are mixed into a single reconstruction. For example, for the interior 3D reconstruction shown in Figure \ref{fig:coloring}, we do not expect to see points labeled as ``facade'' or ``tower'', since those concepts appear outdoors. 
\emph{Interior} concepts include ``organ'', ``nave'', ``altar'', and ``choir'', and  \emph{exterior} concepts include ``portal'', ``facade'', and ``tower''. For each 3D reconstruction $m$, the error $\Delta_\varphi$ is defined as
\begin{equation}
\Delta^m_\varphi = \min \left(p_\mathsf{ext}, 1-p_\mathsf{ext} \right),
\end{equation}
where $p_\mathsf{ext}$ is the probability of an exterior concept in the 3D reconstruction (normalized over the sum of exterior and interior concepts in the reconstruction).
We perform a weighted averaging over all the reconstructions, such that larger 3D reconstructions affect the average accordingly.

We report results for both $\theta_\varphi=0.5$ and $\theta_\varphi=0.75$ in Table \ref{tab:3d} (note that all our qualitative results are generated using $\theta_\varphi=0.5$). 
As illustrated in the table, our model surpasses the baseline model (trained without the 3D contrastive loss) on both metrics, demonstrating that more points are consistently associated with concepts, and that each point cloud is more consistently segmented into exterior or interior concepts.
Note that some structural parts
are inherently more ambiguous (for example, a ``statue'' is often placed on a ``facade''), hence many 3D points are not associated with concepts (also for our model). We explore this further in the supplementary material, showing a confusion matrix for our image classification model as well as the ancestor labels associated with each concept.

\newcolumntype{Y}{>{\centering\arraybackslash}X}
\newcolumntype{L}{>{\arraybackslash}X}
\begin{table}[tb]
\centering
\setlength{\tabcolsep}{1.7pt}
\def\arraystretch{0.8}
\begin{tabularx}{\columnwidth}{lllcccccccccccccccccc}
\toprule
Model &&  R1 & R5 & R10 && S1 & S5 & S10 && S1$^\star$ & S5$^\star$ & S10$^\star$    \\ \midrule
Pretrained && 1.2 & 4.3 & 6.6 && 22.9 & 51.0 & 67.2 && 44.2 & 73.9 & 85.8 \\
 Baseline  && 3.2 & 11.9 & 19.2 && 51.9 & 80.6 & 88.0 && 69.2 & 89.3 & 94.6 \\
 Ours  && \textbf{4.0} & \textbf{13.9} & \textbf{22.5} && \textbf{64.0} & \textbf{81.9} & \textbf{91.2} && \textbf{76.0} & \textbf{91.2} & \textbf{96.3} \\
\bottomrule
\end{tabularx}
\vspace{-8pt}
\caption{Caption-Based Image Retrieval Performance. We report performance using a standard retrieval metric and our proposed semantic metric ($^\star$ indicates averaging over all images, and not per class). Results of our model are compared against a model trained without our 3D augmentations (baseline) and on the pretrained model~\cite{lu202012}. Performance is reported on images from unseen landmarks (WS-U). The best
results are highlighted in bold. 
}
\label{tab:retrieval}
\end{table}

\setlength{\fboxrule}{1pt}

\begin{figure}[tb] %
\jsubfig{\fbox{\includegraphics[height=1.22cm]{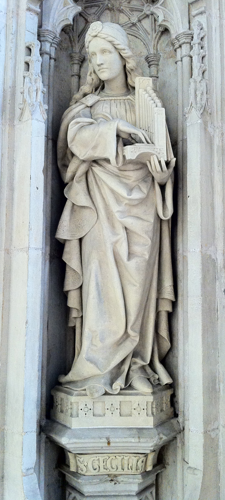}} \hfill \includegraphics[height=1.22cm]{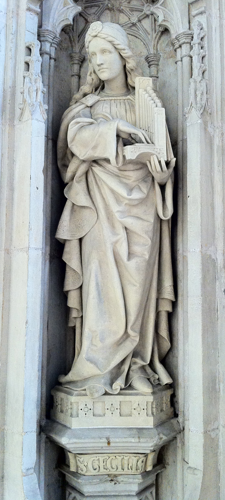}
\includegraphics[height=1.22cm]{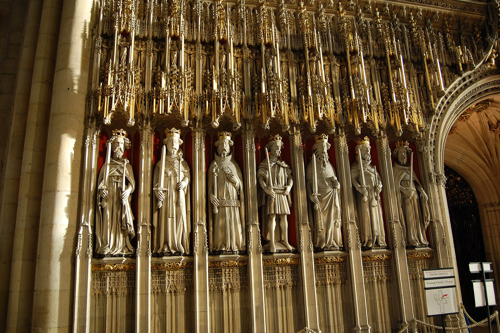}
\includegraphics[height=1.22cm]{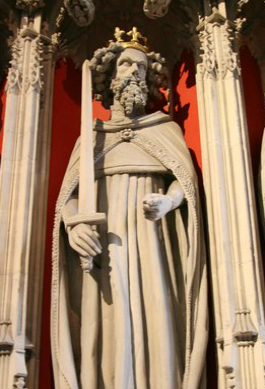}}
{\leftskip=0.1pt   \footnotesize{``Statue of Saint Cecilia in the south transept of York Minster.''} } 
\hspace{4pt}
\jsubfig{\fbox{\includegraphics[height=1.22cm]{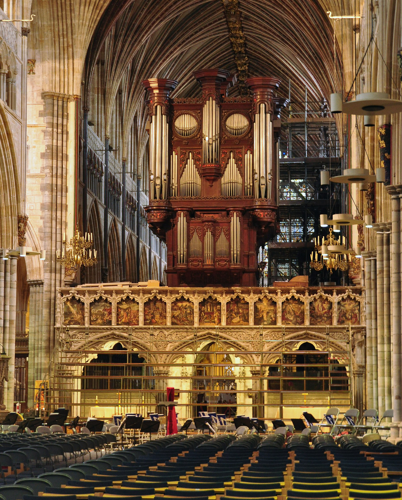}} \hfill  \includegraphics[height=1.22cm]{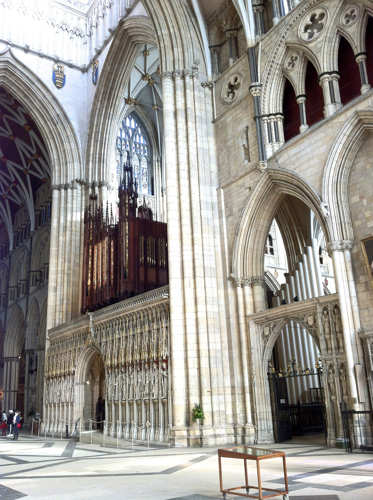}
\includegraphics[height=1.22cm]{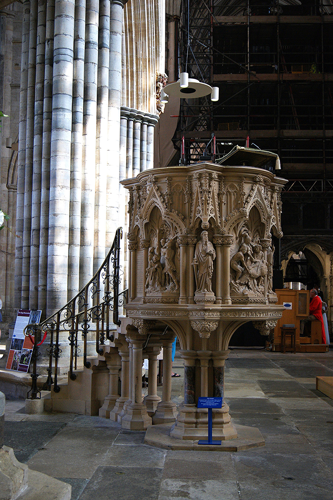}
\includegraphics[height=1.22cm]{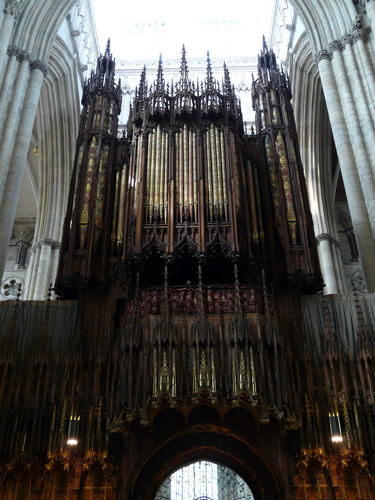}}
{\leftskip=0.1pt   \footnotesize{``The organ in Exeter Cathedral in Devon.''} } 
\\
\jsubfig{\fbox{\includegraphics[height=2.0cm]{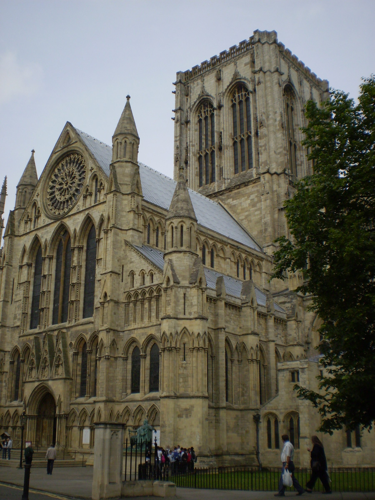}} \hfill \includegraphics[height=2.0cm]{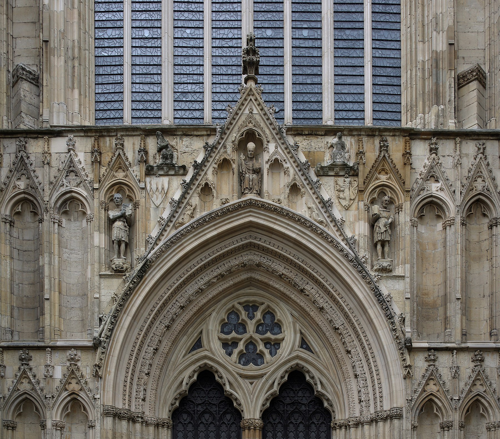}
\includegraphics[height=2.0cm]{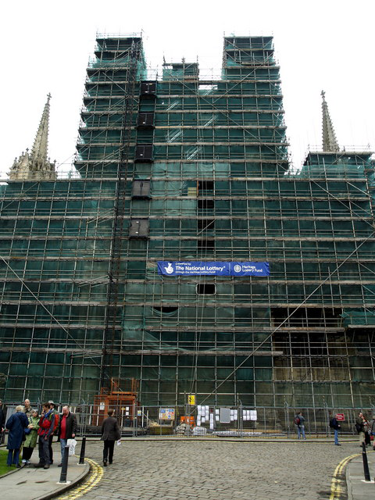}
\includegraphics[height=2.0cm]{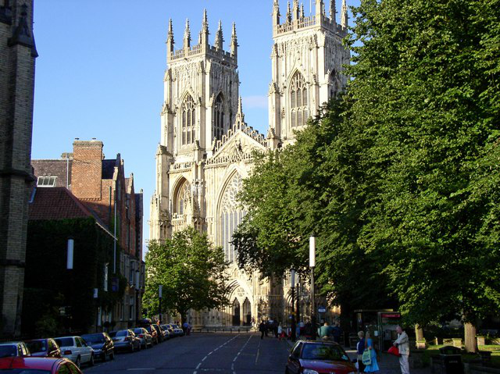}}
{\leftskip=0.1pt   \footnotesize{``York Minster as seen from across the street, York, England.''} } 
\vspace{-8pt}
\caption{Retrieving images from captions of unseen landmarks. Above we show the top three retrievals next to the target image (left), corresponding to the caption below.
}
\label{fig:retrieval}
\end{figure}

\subsection{Learning semantics from raw captions}
\rev{ 
To explore the utility of the raw captions without first distilling concepts, we train a joint vision-language model on images and their raw captions and evaluate it on a caption-based image retrieval task. As with other tasks like classification, we explore the benefit of having 3D geometry in this experiment, showing that geometry can be used to perform data augmentation and boost 
retrieval performance.
}

We finetune a state-of-the-art multi-task joint visual and textual representation model~\cite{lu202012} using the same landmarks-level splits as above, training on landmarks from WS-K and testing on unseen landmarks in WS-U. We compare models finetuned on two different subsets: (1) a \emph{baseline} subset, provided with pairs of English-only captions and their corresponding images, and (2) a \emph{3D-augmented} subset, where, in addition to the real image-caption pairs, we create new image-caption pairs by associating images with captions from other images with a large visual overlap (measured by thresholding on an IoU ratio of 3D keypoints, set empirically to 0.3). Performing such 3D-aware augmentation enables use of additional images---for which a caption may be unavailable---but whose 
content is similar to the original image (while appearance and viewpoint may vary). 
\ignorethis{
Augmenting with such 
additional 3D-augmented (and thus semantically-related) pairs 
can help models 
disentangle nuisance factors from the semantic content. 
}
Our 3D-augmentation strategy yields a training dataset with roughly 1.5K more images and 9K more image-caption pairs (the original training set contains nearly 20K pairs).

\rev{Table \ref{tab:retrieval} shows caption-based image retrieval performance using Recall@K (R1,R5,R10 in the table), which is a standard metric that measures the percentage of successful retrievals for which the target image is among the top-K retrievals.
Additionally, to quantify how \emph{semantically} accurate these retrievals are, we use our semantic labels (obtained according to the method described in Section \ref{sec:association}) as a proxy and propose a semantic measure S that measures the percentage of retrievals containing at least one image labeled correctly. %
All metrics are reported for the two models and also for the pretrained model~\cite{lu202012} (without finetuning). For our semantic metric, we report an average per class and average over all the images in the test set.}

\rev{Using 3D augmentations gives a boost in performance across all metrics. Figure~\ref{fig:retrieval} illustrates several retrieval results from our model. As illustrated in the bottom row, the model can also align general concepts to our images, such as what a cathedral should look like ``from across the street''. We show additional qualitative results 
in the supplementary material.  }

 \section{Conclusion}
We have presented a new large-scale dataset at the intersection of vision, language, and 3D. We demonstrated the use 
of our dataset for mining semantic concepts and for learning 
to associate these concepts with images and 3D models from never-before-seen locations. 
We show that these tasks benefit from having access to 3D geometry, 
allowing robust distillation of semantics from 
noisy Internet collections.

\medskip
\noindent
\textbf{Future applications.}
We believe our dataset could spark research into many new problems. Automatic captioning of images capturing tourist attractions is one interesting avenue for future research. The rich textual descriptions in our dataset could allow users to virtually explore any tourist attraction, serving as a virtual ``tour guide''. 
Our dataset 
could also enable automatic generation of new 3D scenes and language-guided scene editing. While 
text-based 2D image generation is a very active research area \cite{dong2017semantic,nam2018text,li2019controllable}, the problem of generating 
and modifying 3D scenes 
using language 
is largely unexplored. 
Finally, our focus was on discovery of well-supported concepts, but our dataset can also benefit 
zero- or few-shot settings 
via the detailed descriptions present in image captions, enabling rich conceptualization of general visual concepts.

\ignorethis{
\begin{enumerate}
    \item 3D scene generation and editing (language-guided editing). With the rise of generative models for images and 3D models, we are witnessing growing interest among various research communities in controlling the generated output. %
    As language can freely describe a desired visual result, much attention has been devoted to text-based image generation \cite{dong2017semantic,nam2018text,li2019controllable}.
    We plan to explore v. 

    \item Automatic virtual tour guide (image captioning). This will allow users to virtually explore any tourist attraction in a descriptive and informative manner from just a single (or a collection) of unlabeled images.
    \item Geometric relationships graph (e.g., tymapanum is part of a facade or tymapanum is above portal) using a joint language and 3D analysis
    \item Boosting SfM techniques (based on semantics / language cues)
    \item Few-shot learning (we focused on classes with large support but there are many more concepts that have less support) with 3D consistency
    \item    Navigation in complex 3D enviroments. Our dataset could challenge models trained to follow natural language instructions. %
   
\end{enumerate}
}

\medskip
\noindent \textbf{Acknowledgments.}
This work was supported by the National Science Foundation (IIS-2008313), by the generosity of Eric and Wendy Schmidt by recommendation of the Schmidt Futures program, by the Zuckerman STEM leadership program, and by an AWS ML Research Award.

{\small
\bibliographystyle{ieee_fullname}
\bibliography{main}

\begin{thebibliography}{10}\itemsep=-1pt

\bibitem{achlioptas2019shapeglot}
Panos Achlioptas, Judy Fan, Robert Hawkins, Noah Goodman, and Leonidas~J
  Guibas.
\newblock Shape{G}lot: Learning language for shape differentiation.
\newblock In {\em ICCV}, 2019.

\bibitem{agarwal2011building}
Sameer Agarwal, Yasutaka Furukawa, Noah Snavely, Ian Simon, Brian Curless,
  Steven~M Seitz, and Richard Szeliski.
\newblock Building {R}ome in a day.
\newblock {\em Communications of the ACM}, 54(10), 2011.

\bibitem{agrawal2016analyzing}
Aishwarya Agrawal, Dhruv Batra, and Devi Parikh.
\newblock Analyzing the behavior of visual question answering models.
\newblock {\em arXiv preprint arXiv:1606.07356}, 2016.

\bibitem{anderson2018bottom}
Peter Anderson, Xiaodong He, Chris Buehler, Damien Teney, Mark Johnson, Stephen
  Gould, and Lei Zhang.
\newblock Bottom-up and top-down attention for image captioning and visual
  question answering.
\newblock In {\em CVPR}, 2018.

\bibitem{Anderson:18r2r}
Peter Anderson, Qi Wu, Damien Teney, Jake Bruce, Mark Johnson, Niko
  S{\"u}nderhauf, Ian Reid, Stephen Gould, and Anton van~den Hengel.
\newblock Vision-and-language navigation: Interpreting visually-grounded
  navigation instructions in real environments.
\newblock In {\em CVPR}, 2018.

\bibitem{antol2015vqa}
Stanislaw Antol, Aishwarya Agrawal, Jiasen Lu, Margaret Mitchell, Dhruv Batra,
  C.~Lawrence Zitnick, and Devi Parikh.
\newblock {VQA}: {V}isual {Q}uestion {A}nswering.
\newblock In {\em ICCV}, 2015.

\bibitem{araslanov2020single}
Nikita Araslanov and Stefan Roth.
\newblock Single-stage semantic segmentation from image labels.
\newblock In {\em CVPR}, 2020.

\bibitem{Blukis:18drone}
Valts Blukis, Nataly Brukhim, Andrew Bennett, Ross~A. Knepper, and Yoav Artzi.
\newblock Following high-level navigation instructions on a simulated
  quadcopter with imitation learning.
\newblock In {\em Proceedings of the Robotics: Science and Systems Conference},
  2018.

\bibitem{chang2015shapenet}
Angel~X Chang, Thomas Funkhouser, Leonidas Guibas, Pat Hanrahan, Qixing Huang,
  Zimo Li, Silvio Savarese, Manolis Savva, Shuran Song, Hao Su, et~al.
\newblock {ShapeNet}: An information-rich 3d model repository.
\newblock {\em arXiv preprint arXiv:1512.03012}, 2015.

\bibitem{chen2019scanrefer}
Dave~Zhenyu Chen, Angel~X Chang, and Matthias Nie{\ss}ner.
\newblock Scanrefer: 3d object localization in rgb-d scans using natural
  language.
\newblock {\em arXiv preprint arXiv:1912.08830}, 2019.

\bibitem{chen2018text2shape}
Kevin Chen, Christopher~B Choy, Manolis Savva, Angel~X Chang, Thomas
  Funkhouser, and Silvio Savarese.
\newblock Text2shape: Generating shapes from natural language by learning joint
  embeddings.
\newblock In {\em ACCV}, 2018.

\bibitem{7913730}
L. {Chen}, G. {Papandreou}, I. {Kokkinos}, K. {Murphy}, and A.~L. {Yuille}.
\newblock {DeepLab}: {S}emantic image segmentation with deep convolutional
  nets, atrous convolution, and fully connected crfs.
\newblock {\em PAMI}, 40(4):834--848, 2018.

\bibitem{chen2020simple}
Ting Chen, Simon Kornblith, Mohammad Norouzi, and Geoffrey Hinton.
\newblock A simple framework for contrastive learning of visual
  representations.
\newblock In {\em Proc. Int. Conf. on Machine Learning}. PMLR, 2020.

\bibitem{crandall2009mapping}
David~J Crandall, Lars Backstrom, Daniel Huttenlocher, and Jon Kleinberg.
\newblock Mapping the world's photos.
\newblock In {\em Proc. Int. Conf. on World Wide Web}, 2009.

\bibitem{dong2017semantic}
Hao Dong, Simiao Yu, Chao Wu, and Yike Guo.
\newblock Semantic image synthesis via adversarial learning.
\newblock In {\em ICCV}, 2017.

\bibitem{Fukui:16bilinearpoolvqa}
Akira Fukui, Dong~Huk Park, Daylen Yang, Anna Rohrbach, Trevor Darrell, and
  Marcus Rohrbach.
\newblock Multimodal compact bilinear pooling for visual question answering and
  visual grounding.
\newblock In {\em Proceedings of the Conference on Empirical Methods in Natural
  Language Processing}, pages 457--468, 2016.

\bibitem{furukawa2010towards}
Yasutaka Furukawa, Brian Curless, Steven~M Seitz, and Richard Szeliski.
\newblock Towards internet-scale multi-view stereo.
\newblock In {\em CVPR}, 2010.

\bibitem{gammeter2009know}
Stephan Gammeter, Lukas Bossard, Till Quack, and Luc Van~Gool.
\newblock I know what you did last summer: object-level auto-annotation of
  holiday snaps.
\newblock In {\em ICCV}, pages 614--621, 2009.

\bibitem{goesele2007multi}
Michael Goesele, Noah Snavely, Brian Curless, Hugues Hoppe, and Steven~M Seitz.
\newblock Multi-view stereo for community photo collections.
\newblock In {\em ICCV}, 2007.

\bibitem{gupta2020contrastive}
Tanmay Gupta, Arash Vahdat, Gal Chechik, Xiaodong Yang, Jan Kautz, and Derek
  Hoiem.
\newblock Contrastive learning for weakly supervised phrase grounding.
\newblock {\em arXiv preprint arXiv:2006.09920}, 2020.

\bibitem{he2020momentum}
Kaiming He, Haoqi Fan, Yuxin Wu, Saining Xie, and Ross Girshick.
\newblock Momentum contrast for unsupervised visual representation learning.
\newblock In {\em CVPR}, 2020.

\bibitem{he2016deep}
Kaiming He, Xiangyu Zhang, Shaoqing Ren, and Jian Sun.
\newblock Deep residual learning for image recognition.
\newblock In {\em CVPR}, 2016.

\bibitem{hong2019learning}
Richang Hong, Daqing Liu, Xiaoyu Mo, Xiangnan He, and Hanwang Zhang.
\newblock Learning to compose and reason with language tree structures for
  visual grounding.
\newblock {\em PAMI}, 2019.

\bibitem{Hu:17compnetqa}
Ronghang Hu, Jacob Andreas, Marcus Rohrbach, Trevor Darrell, and Kate Saenko.
\newblock Learning to reason: End-to-end module networks for visual question
  answering.
\newblock In {\em ICCV}, pages 804--813, 2017.

\bibitem{adamoptimizer}
Diederik~P. Kingma and Jimmy Ba.
\newblock Adam: A method for stochastic optimization.
\newblock {\em CoRR}, abs/1412.6980, 2014.

\bibitem{kojima2020learned}
Noriyuki Kojima, Hadar Averbuch-Elor, Alexander~M Rush, and Yoav Artzi.
\newblock What is learned in visually grounded neural syntax acquisition.
\newblock In {\em ACL}, 2020.

\bibitem{li2019controllable}
Bowen Li, Xiaojuan Qi, Thomas Lukasiewicz, and Philip Torr.
\newblock Controllable text-to-image generation.
\newblock In {\em NeurIPS}, 2019.

\bibitem{li2018megadepth}
Zhengqi Li and Noah Snavely.
\newblock Megadepth: Learning single-view depth prediction from internet
  photos.
\newblock In {\em CVPR}, 2018.

\bibitem{li2020crowdsampling}
Zhengqi Li, Wenqi Xian, Abe Davis, and Noah Snavely.
\newblock Crowdsampling the plenoptic function.
\newblock In {\em ECCV}, 2020.

\bibitem{DBLP:journals/ijcv/Lowe04}
David~G. Lowe.
\newblock Distinctive image features from scale-invariant keypoints.
\newblock {\em Int. J. Comput. Vis.}, 60(2):91--110, 2004.

\bibitem{lu202012}
Jiasen Lu, Vedanuj Goswami, Marcus Rohrbach, Devi Parikh, and Stefan Lee.
\newblock 12-in-1: Multi-task vision and language representation learning.
\newblock In {\em CVPR}, 2020.

\bibitem{lu2017knowing}
Jiasen Lu, Caiming Xiong, Devi Parikh, and Richard Socher.
\newblock Knowing when to look: Adaptive attention via a visual sentinel for
  image captioning.
\newblock In {\em CVPR}, 2017.

\bibitem{Mao2015:refexp}
Junhua Mao, Jonathan Huang, Alexander Toshev, Oana Camburu, Alan~L. Yuille, and
  Kevin Murphy.
\newblock Generation and comprehension of unambiguous object descriptions.
\newblock In {\em CVPR}, pages 11--20, 2016.

\bibitem{martinbrualla2020nerfw}
Ricardo Martin-Brualla, Noha Radwan, Mehdi S.~M. Sajjadi, Jonathan~T. Barron,
  Alexey Dosovitskiy, and Daniel Duckworth.
\newblock {NeRF in the Wild: Neural Radiance Fields for Unconstrained Photo
  Collections}.
\newblock In {\em CVPR}, 2021.

\bibitem{meshry2019neural}
Moustafa Meshry, Dan~B Goldman, Sameh Khamis, Hugues Hoppe, Rohit Pandey, Noah
  Snavely, and Ricardo Martin-Brualla.
\newblock Neural rerendering in the wild.
\newblock In {\em CVPR}, pages 6878--6887, 2019.

\bibitem{Misra:17instructions}
Dipendra Misra, John Langford, and Yoav Artzi.
\newblock Mapping instructions and visual observations to actions with
  reinforcement learning.
\newblock In {\em Proceedings of the Conference on Empirical Methods in Natural
  Language Processing}, pages 1004--1015, 2017.

\bibitem{mo2019partnet}
Kaichun Mo, Shilin Zhu, Angel~X Chang, Li Yi, Subarna Tripathi, Leonidas~J
  Guibas, and Hao Su.
\newblock Partnet: A large-scale benchmark for fine-grained and hierarchical
  part-level 3d object understanding.
\newblock In {\em CVPR}, 2019.

\bibitem{nakatani2010langdetect}
Shuyo Nakatani.
\newblock Language detection library for java, 2010.

\bibitem{nam2018text}
Seonghyeon Nam, Yunji Kim, and Seon~Joo Kim.
\newblock Text-adaptive generative adversarial networks: manipulating images
  with natural language.
\newblock In {\em NeurIPS}, 2018.

\bibitem{oord2018representation}
Aaron van~den Oord, Yazhe Li, and Oriol Vinyals.
\newblock Representation learning with contrastive predictive coding.
\newblock {\em arXiv preprint arXiv:1807.03748}, 2018.

\bibitem{park2020contrastive}
Taesung Park, Alexei~A Efros, Richard Zhang, and Jun-Yan Zhu.
\newblock Contrastive learning for unpaired image-to-image translation.
\newblock In {\em ECCV}. Springer, 2020.

\bibitem{paszke2017automatic}
Adam Paszke, Sam Gross, Soumith Chintala, Gregory Chanan, Edward Yang, Zachary
  DeVito, Zeming Lin, Alban Desmaison, Luca Antiga, and Adam Lerer.
\newblock Automatic differentiation in pytorch.
\newblock 2017.

\bibitem{philbin2008object}
James Philbin and Andrew Zisserman.
\newblock Object mining using a matching graph on very large image collections.
\newblock In {\em 2008 Sixth Indian Conference on Computer Vision, Graphics \&
  Image Processing}, pages 738--745. IEEE, 2008.

\bibitem{qi2020stanza}
Peng Qi, Yuhao Zhang, Yuhui Zhang, Jason Bolton, and Christopher~D Manning.
\newblock Stanza: A python natural language processing toolkit for many human
  languages.
\newblock {\em arXiv preprint arXiv:2003.07082}, 2020.

\bibitem{quack2008world}
Till Quack, Bastian Leibe, and Luc Van~Gool.
\newblock World-scale mining of objects and events from community photo
  collections.
\newblock In {\em Proceedings of the 2008 international conference on
  Content-based image and video retrieval}, pages 47--56, 2008.

\bibitem{russell20133d}
Bryan~C Russell, Ricardo Martin-Brualla, Daniel~J Butler, Steven~M Seitz, and
  Luke Zettlemoyer.
\newblock {3D Wikipedia}: {U}sing online text to automatically label and
  navigate reconstructed geometry.
\newblock In {\em SIGGRAPH}, 2013.

\bibitem{sandler2018mobilenetv2}
Mark Sandler, Andrew Howard, Menglong Zhu, Andrey Zhmoginov, and Liang-Chieh
  Chen.
\newblock Mobilenetv2: Inverted residuals and linear bottlenecks.
\newblock In {\em CVPR}, pages 4510--4520, 2018.

\bibitem{schonberger2016structure}
Johannes~L Schonberger and Jan-Michael Frahm.
\newblock Structure-from-motion revisited.
\newblock In {\em CVPR}, 2016.

\bibitem{schoenberger2016vote}
Johannes~Lutz Sch\"{o}nberger, True Price, Torsten Sattler, Jan-Michael Frahm,
  and Marc Pollefeys.
\newblock A vote-and-verify strategy for fast spatial verification in image
  retrieval.
\newblock In {\em Asian Conference on Computer Vision (ACCV)}, 2016.

\bibitem{simon2008scene}
Ian Simon and Steven~M. Seitz.
\newblock Scene segmentation using the wisdom of crowds.
\newblock In {\em ECCV}, pages 541--553, 2008.

\bibitem{simon2007scene}
Ian Simon, Noah Snavely, and Steven~M. Seitz.
\newblock Scene summarization for online image collections.
\newblock In {\em ICCV}, 2007.

\bibitem{snavely2006photo}
Noah Snavely, Steven~M Seitz, and Richard Szeliski.
\newblock {Photo tourism: Exploring photo collections in 3D}.
\newblock In {\em SIGGRAPH}, 2006.

\bibitem{suris2020globetrotter}
D{\'\i}dac Sur{\'\i}s, Dave Epstein, and Carl Vondrick.
\newblock Globetrotter: Unsupervised multilingual translation from visual
  alignment.
\newblock {\em arXiv preprint arXiv:2012.04631}, 2020.

\bibitem{Wang:16phraselocalize}
Mingzhe Wang, Mahmoud Azab, Noriyuki Kojima, Rada Mihalcea, and Jia Deng.
\newblock Structured matching for phrase localization.
\newblock In {\em ECCV}, pages 696--711, 2016.

\bibitem{weyand2013discovering}
Tobias Weyand and Bastian Leibe.
\newblock Discovering details and scene structure with hierarchical iconoid
  shift.
\newblock In {\em ICCV}, 2013.

\bibitem{xiao2017weakly}
Fanyi Xiao, Leonid Sigal, and Yong Jae~Lee.
\newblock Weakly-supervised visual grounding of phrases with linguistic
  structures.
\newblock In {\em CVPR}, 2017.

\bibitem{you2016image}
Quanzeng You, Hailin Jin, Zhaowen Wang, Chen Fang, and Jiebo Luo.
\newblock Image captioning with semantic attention.
\newblock In {\em CVPR}, 2016.

\bibitem{Yu:16refmscoco}
Licheng Yu, Patrick Poirson, Shan Yang, Alexander~C Berg, and Tamara~L Berg.
\newblock Modeling context in referring expressions.
\newblock In {\em ECCV}, pages 69--85, 2016.

\bibitem{yu2020self}
Ye Yu, Abhimitra Meka, Mohamed Elgharib, Hans-Peter Seidel, Christian Theobalt,
  and William A.~P. Smith.
\newblock Self-supervised outdoor scene relighting.
\newblock In {\em ECCV}, 2020.

\bibitem{yu2019inverserendernet}
Ye Yu and William A.~P. Smith.
\newblock {InverseRenderNet}: Learning single image inverse rendering.
\newblock In {\em CVPR}, pages 3155--3164, 2019.

\end{thebibliography}
}

\newpage
\appendix
\section{Dataset Visualizations and Details} 
\begin{figure*}
  \centering
  \begin{tabular}{cccccc}
  \includegraphics[width=0.14\textwidth]{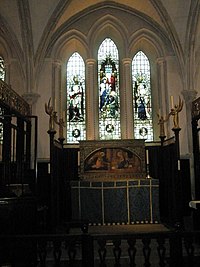} &   \includegraphics[width=0.14\textwidth]{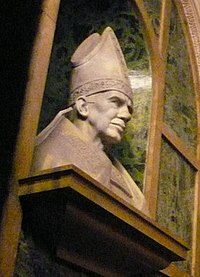} &
   \includegraphics[width=0.14\textwidth]{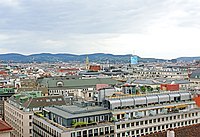} &
\includegraphics[width=0.14\textwidth]{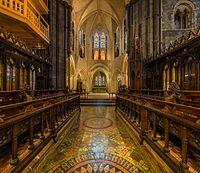} &
\includegraphics[width=0.14\textwidth]{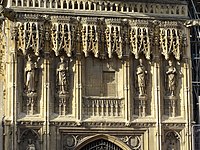} & 
\includegraphics[width=0.14\textwidth]{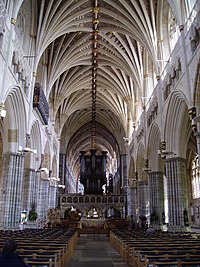} \\
  (a) & (b) & (c) & (d) & (e) & (f) \\
    \includegraphics[width=0.14\textwidth]{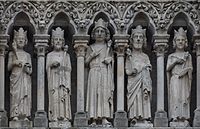} &   \includegraphics[width=0.14\textwidth]{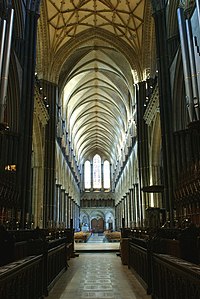} &
   \includegraphics[width=0.14\textwidth]{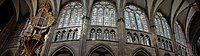} &
\includegraphics[width=0.14\textwidth]{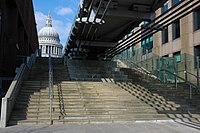} &
\includegraphics[width=0.14\textwidth]{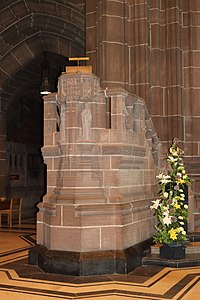} & 
\includegraphics[width=0.14\textwidth]{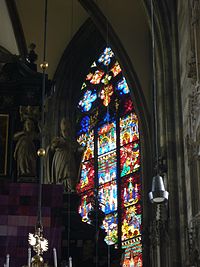} \\
  (g) & (h) & (i) & (j) & (k) & (l)
  \end{tabular}
  \caption{\textbf{Example images from WikiScenes.} Corresponding captions are: (a) \emph{Altar behind the main quire at Southwark Cathedral.} (b) \emph{Bishop Gregorio Modrego over his tomb in cathedral of Barcelona, by Fredric Marès.} (c) \emph{Went to the top of the bell tower to see the views looking over the city of Vienna.} (d) \emph{The choir of Christ Church Cathedral in Dublin, Ireland, looking east towards the sanctuary.} (e) \emph{Statues above the main entrance of Canterbury Cathedral: (left to right) Augustine of Canterbury, Lanfranc, Anselm of Canterbury and Thomas Cranmer.} (f) \emph{The nave of Exeter Cathedral From the west end of the nave looking towards the crossing with its 17th century organ.} (g) \emph{Amiens, France: Fassade detail of the Cathedrale of Amiens, showing the right group of sculptures under the rosette window.} (h) \emph{Salisbury Cathedral Looking towards the West Front, from the Quire.} (i) \emph{The Silbermann organ in Strasbourg cathedral, view from below with the nave windows.} (j) \emph{The Dome of St Paul's Cathedral viewed from the river bank below the Millennium Bridge.} (k) \emph{Sandstone pulpit next to the north transept of Liverpool Anglican Cathedral.} (l) \emph{Window with medieval glass painting behind the high altar in St. Stephen's Cathedral, Vienna.}
 }
  \label{fig:samples}
\end{figure*}

We show captions with spatial connectors and their corresponding images in Figure \ref{fig:samples} to illustrate the richness of part interactions contained within our dataset. %

\begin{figure}
    \centering
    \rotatebox{90}{\whitetxt{xxxxxx}Number of captions}
    \hspace{1pt}
    \includegraphics[width = 0.8\columnwidth]{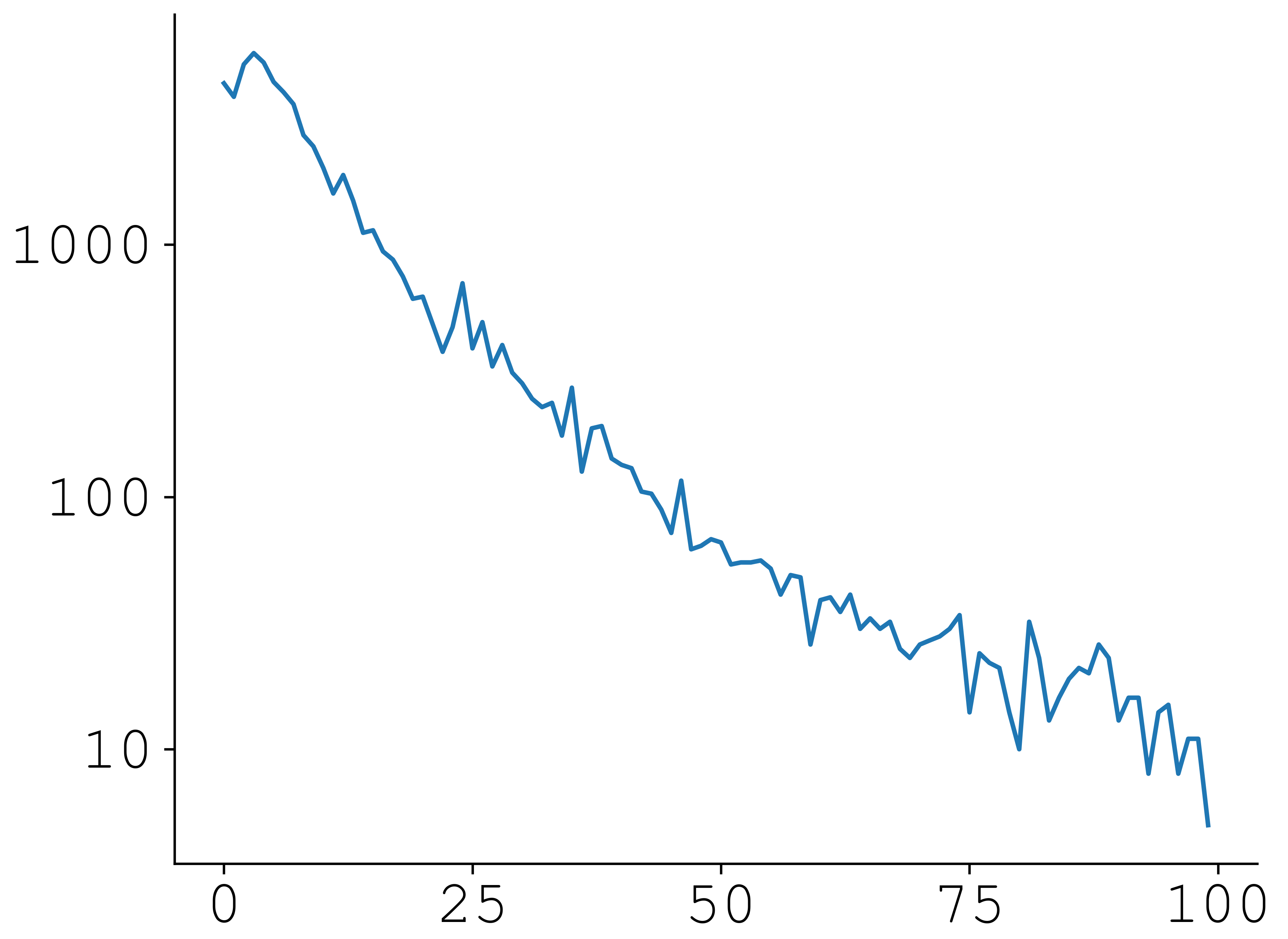} 
 \\
    Number of words
    \caption{Distribution of image captions by number of words ($y$-axis is plotted on a log scale).}
    \label{fig:caption_length}
\end{figure}

\begin{figure}
    \centering
        \rotatebox{90}{\whitetxt{xxxx}Number of data samples }
    \includegraphics[width = 0.8\columnwidth]{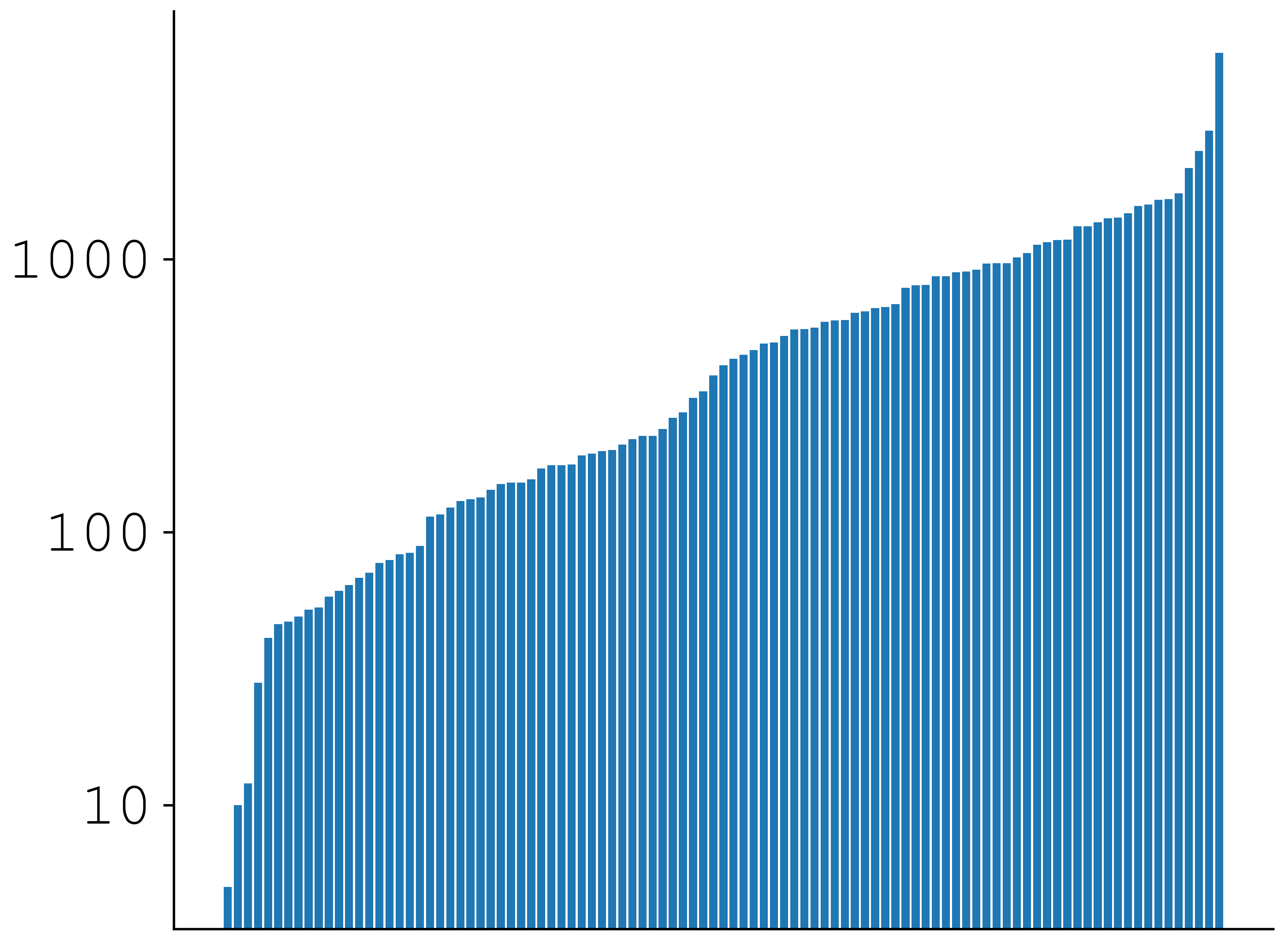} \\
    \caption{Number of images paired with textual descriptions by landmarks (sorted). The $y$-axis is plotted on a log scale.
    }
    \label{fig:landmark_size}
\end{figure}

\begin{figure}
    \centering
    \rotatebox{90}{\whitetxt{xxxxxxxxxxx}Number of captions }
    \includegraphics[width = 0.8\columnwidth]{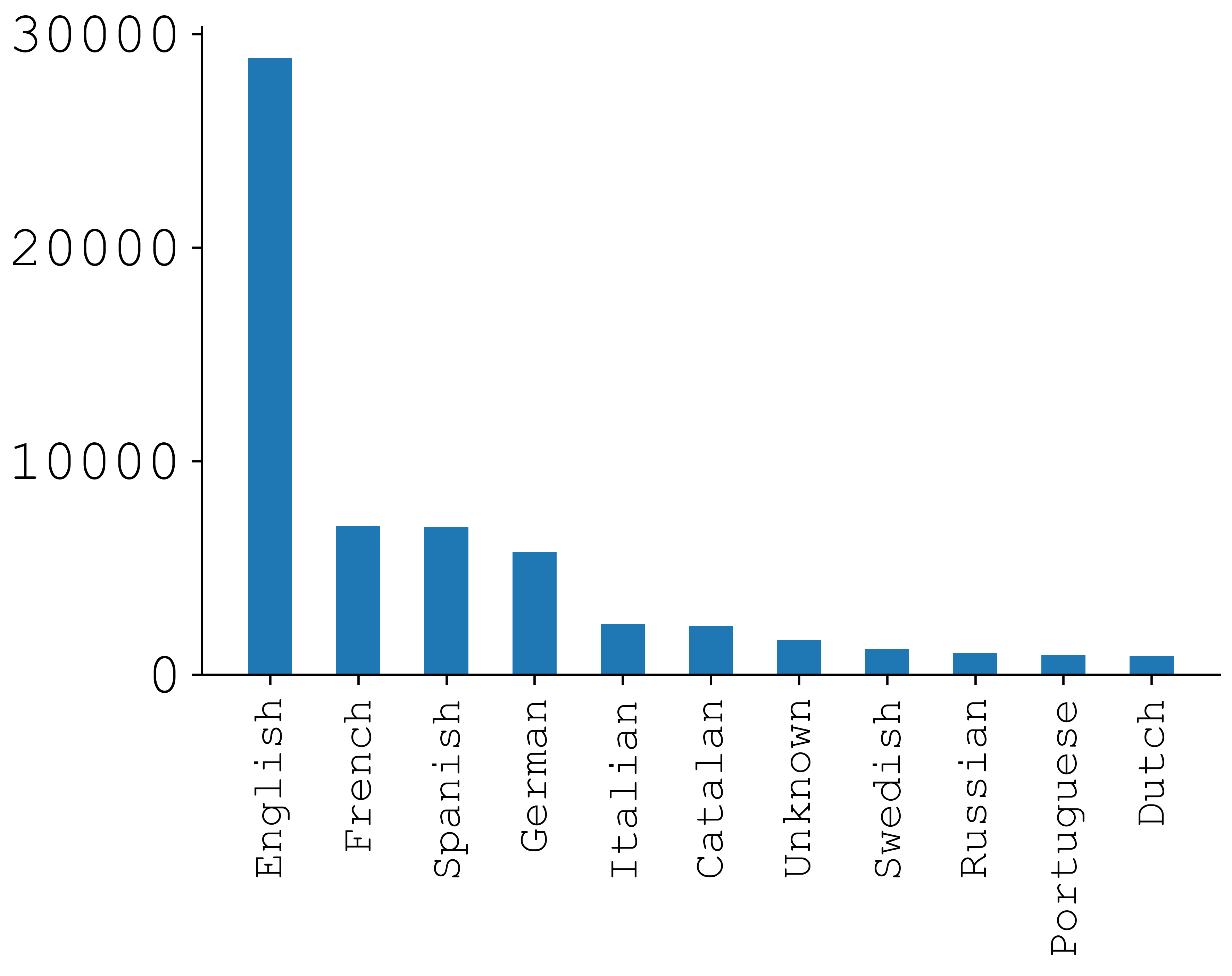} \\
    \caption{Number of captions in the top 10 languages. ``Unknown'' denotes captions that are not recognizable, such as date, URL or null strings.
}.
    \label{fig:language}
\end{figure}

\medskip \noindent \textbf{Data distributions.} Figure \ref{fig:caption_length} shows the distribution of captions by the number of words. Figure \ref{fig:landmark_size} shows the number of data samples by landmark identity sorted by size. Figure \ref{fig:language} shows the number of captions in the top 10 languages. The caption's language is detected according to~\cite{nakatani2010langdetect}.
 

\section{Implementation Details}
\subsection{Dataset construction}

We use COLMAP~\cite{schonberger2016structure} version 3.6 for building 3D reconstructions. The SIFT~\cite{DBLP:journals/ijcv/Lowe04} peak threshold is set to 0.03. To find image matches, we use vocabulary tree matching~\cite{schoenberger2016vote} using the pretrained vocabulary tree with 1M visual words. For landmarks that have reconstructions in the MegaDepth dataset~\cite{li2018megadepth} (we have 44 shared landmarks), external images from their dataset were added to assist reconstruction.  
Original high resolution images are used for reconstruction. However, for training purposes, we use resized images with the shorter dimension set to 200 pixels. We also release a higher resolution version in our dataset, where the longer dimension is set to 1200 pixels. %

\subsection{Network architecture} \label{subsec:network_structure}
\begin{figure*}
  \centering
      \includegraphics[width = 0.9\linewidth]{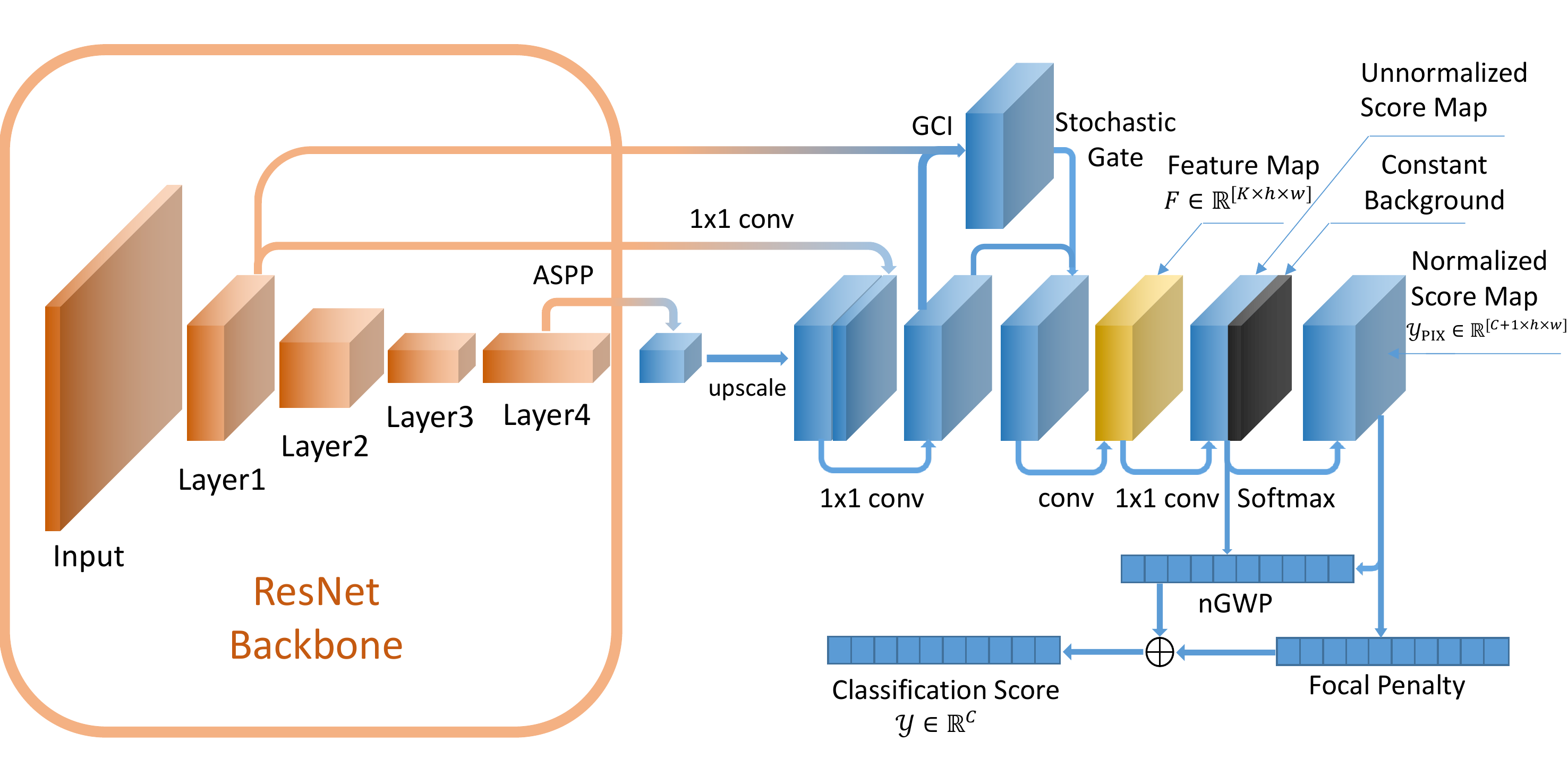}
      \vspace{-3pt}
    \captionof{figure} {Our classification network architecture. 
    }
  \label{fig:network_structure}
\end{figure*}

Figure \ref{fig:network_structure} shows the structure of our network. The structure closely follows the network proposed in Araslanov \etal~\cite{araslanov2020single}. For completeness, we briefly summarize the network architecture here. We use a Resnet-50 backbone to extract both low-level and high-level features (which is pretrained on ImageNet). Atrous Spatial Pyramid Pooling (ASPP)~\cite{7913730} augments the ResNet features by gathering information at different scales. A Global Cue Injection (GCI) module~\cite{araslanov2020single} infuses global cues from deep layers into low-level features derived from the shallow layers of ResNet. The stochastic gate~\cite{araslanov2020single} aims to mitigate overfitting introduced by errors in the pseudo-ground truth used during training. The 3D consistency loss is computed on the features before unnormalized score maps are computed. The classification score $y$ is computed according to ~\cite{araslanov2020single}, 
summing a normalized Global Weighted Pooling (nGWP) term and a focal penalty term (Equation 3 and Equation 5 in their paper).

\subsection{Training details} Our models are implemented in PyTorch~\cite{paszke2017automatic}. %
We train our model using the Adam optimizer ~\cite{adamoptimizer} with weight decay $5\times10^{-4}$, and using default Adam parameters. The model is trained for 25 epochs with learning rate decay occurring at the $15^\mathit{th}$ and $20^\mathit{th}$ epoch. Following ~\cite{araslanov2020single}, we pretrain the model without $\mathcal{L}_{cls}^{pix}$ for 5 epochs. In all our experiments, learning rate decay is performed using a factor of 0.1. For all experiments with $\mathcal{L}_\mathsf{3D}$, the balancing coefficient is set to 0.3 (i.e., the 3D constrastive loss is multiplied by this coefficient). The temperature used in the 3D constrastive loss is set to the default value of 0.07 and the number of negatives is 16. All models are pretrained on ImageNet.  %

As the images in WikiScenes are of varying resolution, we perform a random resized crop operation to convert each image to $[224\times 224]$ training samples. The scale factor of the random resized crop is sampled from the range $[0.9, 1.0]$. Random horizontal flipping and color jittering are also performed to augment the data. The brightness, contrast, saturation and hue parameters are set to $[0.3, 0.3, 0.3, 0.1]$, respectively, in the color jittering step. We balance the size of the different classes by resampling. The balanced dataset contains roughly 900 images in each class.

\subsection{Additional 2D segmentation details}
Our model predicts both image-level classification scores (\emph{i.e.}, $y$, see Figure \ref{fig:network_structure}) and pixel-wise normalized segmentation score maps (\emph{i.e.}, $y_\mathsf{pix}$, see Figure \ref{fig:network_structure}) that also include a background score, in addition to scores for each of the $C$ semantic concepts. Following~\cite{araslanov2020single}, the background score is weakened by a power function (set empirically to the $4^\mathit{th}$ power in our experiments). We first take the maximal value in $y$ to select the image-level label (we only consider images that contain a single label). Then, the 2D segmentation mask is comprised of all pixels whose score corresponding to the selected image-level concept surpasses that of the (weakened) background score.

\subsection{Additional 3D segmentation details}
For each point in a 3D model, we first gather classification scores from its 2D projection in different views. The scores are averaged before applying softmax function to obtain the classification score of the 3D point. Points with scores higher than a predetermined threshold $\varphi$ are considered foreground, and the remaining points are considered ambiguous and are therefore not rendered in our 3D visualizations. For quantitative evaluation, we provide results for $\varphi=0.5$ and $\varphi=0.75$. Our visualizations are rendered using a threshold of $\varphi=0.5$.

\subsection{Caption-based image retrieval experiment}
To perform the caption-based retrieval experiment, we run the command for fine-tuning from the multi-task trained model\footnote{Available here: \url{https://github.com/facebookresearch/vilbert-multi-task}}. We define two new tasks, one for the baseline model (that uses only original image-caption pairs) and another for the 3D-augmented model (that also uses 3D-augmented pairs). Both models are trained for 12 epochs, using their (unmodified) configurations.

Regarding evaluation, to compute our proxy semantic measure S, we followed their retrieval evaluation and construct a batch of $1000$ images from validation, however, in our case, this batch includes all labeled images from unseen images ($774$ images in total) and additional randomly-selected unlabeled images. We use these labels for evaluating whether or not the label of the retrieved images agree with the label of the target image.

\newcolumntype{Y}{>{\centering\arraybackslash}X}
\newcolumntype{L}{>{\arraybackslash}X}
\begin{table*}[h]
\centering
\setlength{\tabcolsep}{2.5pt}
\def\arraystretch{1.15}
\begin{tabularx}{1.0\textwidth}{lllcccccccccccccccccc}
\toprule
Test Set     & &Model &&  mAP  &&           facade & window & chapel & organ & nave & tower & choir & portal & altar & statue    \\ \midrule
\multirow{5}{*}{WS-K} && Baseline (w/o 3D loss) && 70.8 && 87.2&	89.2&	60.2&	89.7&	85.8&	64.1&	61.5&	68.0&	50.0&	52.0 \\
  &&w/ $\mathcal{L}_\mathsf{3D}^\mathsf{MSE}$ &&  71.4 && 86.4&	88.3&	53.1&	89.4&	86.1&	65.7&	62.0&	69.7&	52.5&	60.3\\
  && w/ $\mathcal{L}_\mathsf{3D}^\mathsf{triplet}$ &&  72.1 && 88.5&	\textbf{90.5}&	55.6&	86.0&	\textbf{86.4}&	\textbf{66.5}&	65.0&	68.4&	50.2&	\textbf{63.4}\\
  &&  w/ $\mathcal{L}_\mathsf{3D}(\textit{intra-image sampling})$ &&  73.3 && \textbf{90.4}&	87.1&	62.9&	90.3&	85.8&	62.1&	75.9&	68.4&	52.8&	57.1\\
  && w/ $\mathcal{L}_\mathsf{3D}(\textit{inter-image sampling})$ &&  \textbf{75.3} && 90.0&	88.5&	\textbf{68.7}&	\textbf{90.7}&	85.7&	61.1&	\textbf{77.2}&	\textbf{76.5}&	\textbf{54.4}&	59.9\\
\midrule
\multirow{5}{*}{WS-U} && Baseline (w/o 3D loss) && 48.3 && 71.0&	92.2&	10.7&	57.3&	71.0&	\textbf{53.4}&	43.6&	31.1&	25.8&	27.1\\
  && w/ $\mathcal{L}_\mathsf{3D}^\mathsf{MSE}$ &&  49.5 && 70.6&	94.3&	10.9&	61.8&	73.7&	50.8&	40.9&	41.3&	21.6&	28.9\\
  && w/ $\mathcal{L}_\mathsf{3D}^\mathsf{triplet}$ &&  49.9 && 73.1&	\textbf{94.9}&	9.9&	53.7&	74.7&	47.5&	40.8&	29.1&	39.4&	35.6\\
  && w/ $\mathcal{L}_\mathsf{3D}(\textit{intra-image sampling})$ &&
  \textbf{52.5} && 75.8 &	94.1&	\textbf{16.7}&	\textbf{62.5}&	75.4&	50.4&	\textbf{44.5}&	\textbf{43.0} &	24.4 &	38.4& \\
  &&  w/ $\mathcal{L}_\mathsf{3D}(\textit{inter-image sampling})$ && 52.0 && \textbf{77.7} &	93.4&	16.5&	49.4&	\textbf{77.3}&	46.1&	44.1&	35.2 &	\textbf{39.9} &	\textbf{40.0} \\

\bottomrule
\end{tabularx}
\caption{Classification performance using different types of 3D-consistency regularizations. We report mean average precision (mAP), and per distilled--concept average precision (AP). Please refer to Section \ref{sec:ablations} for more details on the different configurations. %
Performance is reported on images from two different tests sets corresponding to known landmarks (WS-K) and unseen landmarks (WS-U). The best
result for each test set and column are highlighted in bold.}
\label{tab:ablation}
\end{table*}

\section{Ablation Study}
\label{sec:ablations}
We perform an ablation study to analyze our design choices for the 3D-consistency regularization. We replace our 3D contrastive loss with the following alternatives:

\medskip \noindent \textbf{3D MSE loss.} We compute a simple MSE loss between the features of corresponding pixels:
\begin{equation*}
    \mathcal{L}_\mathsf{3D}^\mathsf{MSE} = \|F(p) - F(p^+)\|_2^2. 
\end{equation*}

\medskip \noindent \textbf{3D Triplet loss.} We select one negative pixel $p^-$ and compute the following 3D loss:
\begin{equation*}
    \mathcal{L}_\mathsf{3D}^\mathsf{triplet} = \max (0,\|F(p) - F(p^-)\|_2^2 - 
    \|F(p) - F(p^+)\|_2^2+m),
\end{equation*}
where $m$ is a margin value (set empirically to $3$).

\medskip \noindent \textbf{3D intra-image contrastive loss.} In the main paper, we introduce a 3D contrastive loss $\mathcal{L}_\mathsf{3D}(\textit{inter-image sampling})$, where the negative pixels $\{p_i^- \}$ are sampled from other images in the batch. We change the sampling strategy such that all the negative pixels are selected from other regions in the same image to obtain $\mathcal{L}_\mathsf{3D}(\textit{intra-image sampling})$. Specifically, the points $p_i^-$ are sampled uniformly in $I_2$, outside a box of size $\left[w  \mathbin{/} 4, h  \mathbin{/} 4\right]$ around $p^+$.

\medskip
Results are reported in Table \ref{tab:ablation}. As illustrated in the table, we can improve classification performance using a variety of loss configurations. Our 3D contrastive loss, using both inter-image and intra-image sampling strategies, yield the most significant improvements. %

Following prior work~\cite{araslanov2020single}, our semantic classification loss is composed of two terms, where $\mathcal{L}^\mathsf{cls}_\mathsf{pix}$ is a self-supervised loss over pixelwise predictions that is applied starting at the 6-th epoch. Classification performance is roughly the same when this self-supervised loss is not used. Specifically, mAP increases from 52.0 to 53.8 for WS-U and decreases from 75.3 to 73.4 for WS-K. The gaps to the baseline model mostly remain unchanged (3.3\% improvement for WS-U and 1.8\% improvement for WS-K).

\begin{figure}
    \centering
    \includegraphics[width = 0.85\columnwidth]{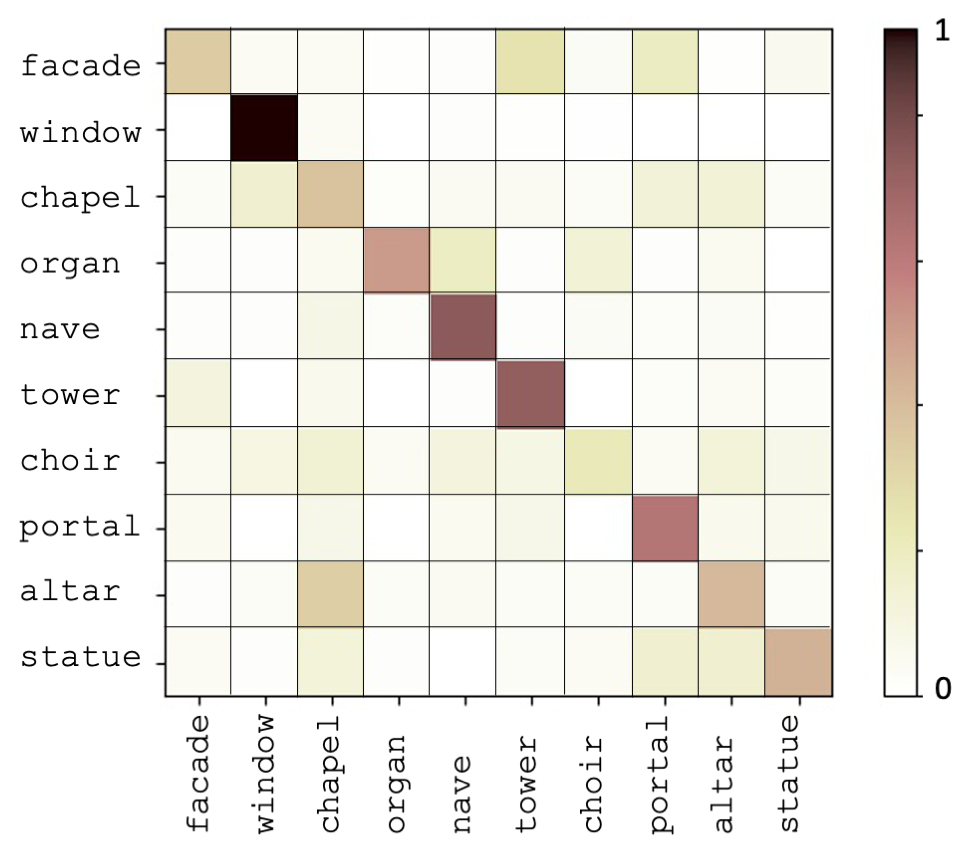} \\
    \caption{Confusion matrix of our classification model on unseen landmarks (the WS-U test set). Ground-truth concept labels correspond to rows, and predicted concept labels to columns. Each row is normalized such that a cell indicates the probability of a classification given the ground-truth label.}
    \label{fig:confusion}
\end{figure}
\section{Additional Classification Results}
Figure \ref{fig:confusion} shows a confusion matrix for our image classification model.
We observe that many of the mistakes are understandable, given the hierarchical nature of our data. For example, both ``tower'' and ``portal'' are part of a ``facade'', and an ``altar'' is often placed inside a ``chapel''. %

\begin{figure}
    \centering
    \rotatebox{90}{\whitetxt{sssssssssssssssss} Associated Labels }
    \hfill
    \includegraphics[width = 0.92\columnwidth]{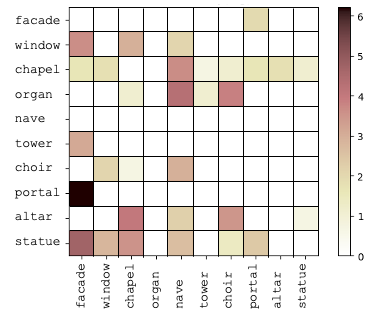} \\
    Associated Anscestor Labels
    
    \caption{Associating images with labels and anscestor labels. Above we visualize (in log scale) the co-occurrence of concepts as labels and ancestor labels. }
    \label{fig:concept_cooc}
\end{figure}
To further explore the hierarchical structure of semantics in our dataset, we associate images with \emph{ancestor} labels by considering the concepts present in its hierarchy of WikiCategories.  %
Unlike prior works that require manually annotating such hierarchical labels (e.g., \cite{mo2019partnet}), we obtain these automatically, leveraging the hierarchical structure of Wikimedia Commons. In Figure \ref{fig:concept_cooc}, we visualize these hierarchical relationships. Many of these relationships can also be observed from the confusion matrix of our model in Figure~\ref{fig:confusion}. We also observe additional intuitive connections such as an image associated with ``window'' also being associated with larger structures such as ``facade'' and ``nave''; a ``statue'' can be placed on various structures, and so on.

\newcolumntype{Y}{>{\centering\arraybackslash}X}
\newcolumntype{L}{>{\arraybackslash}X}
\begin{table}[h]
\centering
\setlength{\tabcolsep}{3.9pt}
\begin{tabularx}{\columnwidth}{lllccccccccc}
\toprule
                          &           & \multicolumn{2}{c}{Resnet-50 \cite{he2016deep}} &  & \multicolumn{2}{c}{MobileNetV2 \cite{sandler2018mobilenetv2}} 
                           \\ 
 \cmidrule(lr){3-4} \cmidrule(lr){6-7}  %
Test Set     &&   w/o $\mathcal{L}_{\mathit{3D}}$ & w/ $\mathcal{L}_{\mathit{3D}}$ &&   w/o $\mathcal{L}_{\mathit{3D}}$& w/ $\mathcal{L}_{\mathit{3D}}$\\ \midrule
WS-K && 68.5 & \textbf{73.9} && 77.1 & \textbf{79.6}\\
WS-U && 48.7 & \textbf{52.3} && 50.2 & \textbf{53.4}\\
\bottomrule
\end{tabularx}
\caption{Evaluating the effectiveness of our 3D contrastive loss on off-the-shelf classification models. For each model, we report mAP. The best results are highlighted in bold. %
}
\label{tab:classification2}
\end{table}

Finally, to further validate the effectiveness of our 3D loss, we take off-the-shelf networks dedicated for classification and repeat the experiment of testing classification performance with and without our 3D contrastive loss. For this experiment, all models are trained for 10 epochs with a learning rate decay at the $6^\mathit{th}$ epoch. Both Resnet-50 and MobileNetV2 are pretrained on ImageNet.

Results are reported in Table \ref{tab:classification2}. As illustrated in the table, our 3D contrastive loss consistently boosts classification performance, even for off-the-shelf models.

\section{Additional Qualitative Results}

\begin{figure}
    \centering
            \rotatebox{90}{\whitetxt{sss} Input}
        \hfill 
\jsubfig{\includegraphics[height=1.85cm]{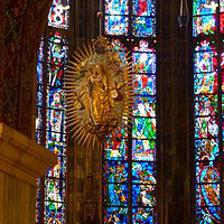}
\includegraphics[height=1.85cm]{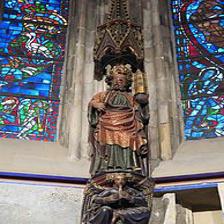}}{}
\hfill 
\jsubfig{\includegraphics[height=1.85cm]{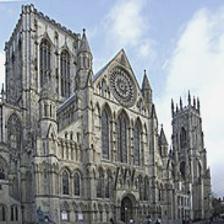}
\includegraphics[height=1.85cm]{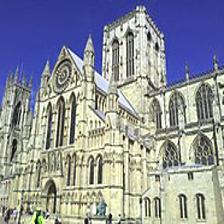}}{} 
\\ \vspace{1pt}
\rotatebox{90}{\whitetxt{ss} w/o $\mathcal{L}_{\mathit{3D}}$}
        \hfill
\hfill
\jsubfig{\includegraphics[height=1.85cm]{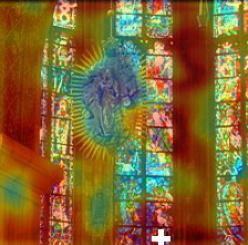}
\includegraphics[height=1.85cm]{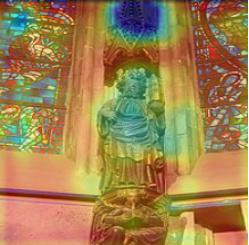}}{}
\hfill
\jsubfig{\includegraphics[height=1.85cm]{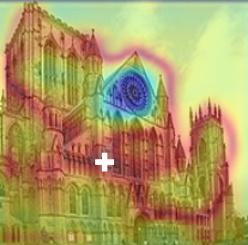}
\includegraphics[height=1.85cm]{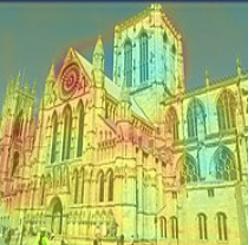}}{}
\\ \vspace{1pt}
        \rotatebox{90}{\whitetxt{sss} Ours} \hspace{1pt}
\hfill
\jsubfig{\includegraphics[height=1.85cm]{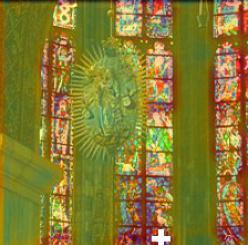}
\includegraphics[height=1.85cm]{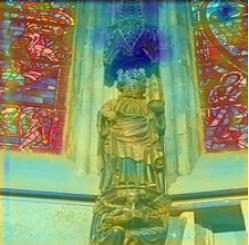}}{}
\hfill
\jsubfig{\includegraphics[height=1.85cm]{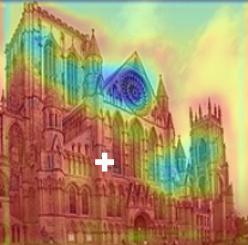}
\includegraphics[height=1.85cm]{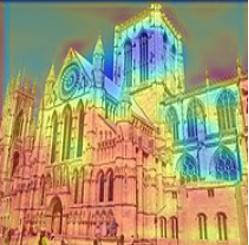}}{}
    \caption{Visualizing distances in feature space for unseen landmarks. For each image pair, we select a random pixel in the left image (marked in white) and visualize the distance to all other pixels from the selected pixel (marked in white) with and without our 3D contrastive loss. Warmer colors correspond to smaller distances. As illustrated above, distances in feature space are more semantically meaningful on the model trained with the 3D contrastive loss (see, for instance, distances on the windows in the left pair). Our model is also more robust against large motion and appearance variations between the images (as illustrated on the right).
}
    \label{fig:features}
\end{figure}
\setlength{\fboxrule}{1pt}

\begin{figure}[tb] %
\jsubfig{\fbox{\includegraphics[height=1.35cm]{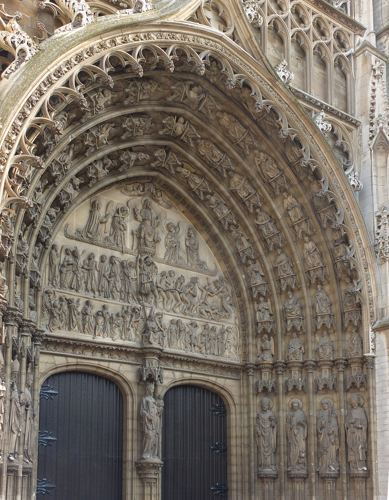}} \hfill \includegraphics[height=1.35cm]{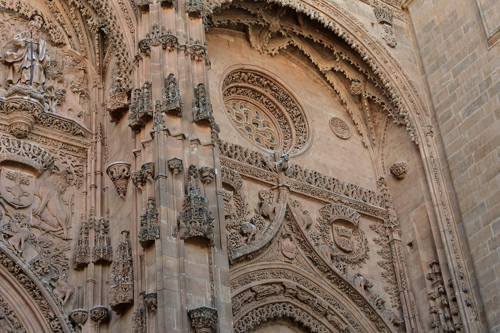}
\includegraphics[height=1.35cm]{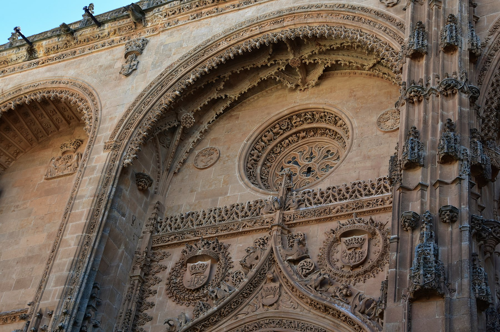}
\includegraphics[height=1.35cm]{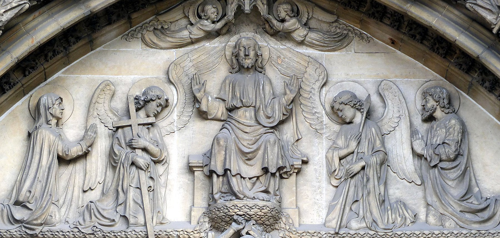}}
{\leftskip=0.1pt   \footnotesize{``Neogothic portal of Our Lady's Cathedral, Antwerp, by Jean Baptist van Wint (1829-1906). The Cathedral of Our Lady is a Roman Catholic parish church in Antwerp, Belgium.''} } 
\jsubfig{\fbox{\includegraphics[height=1.44cm]{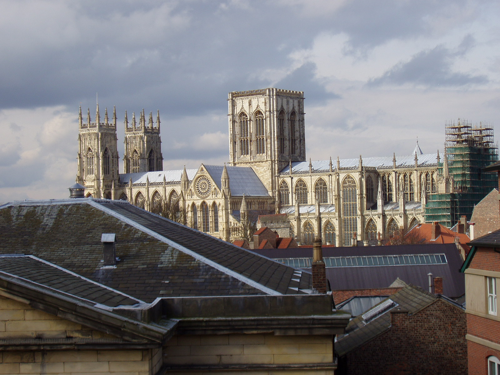}} \hfill  \includegraphics[height=1.44cm]{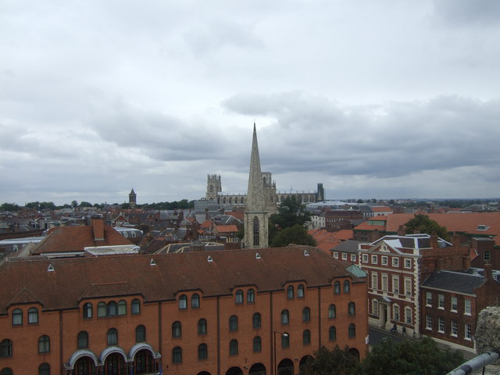}
\includegraphics[height=1.44cm]{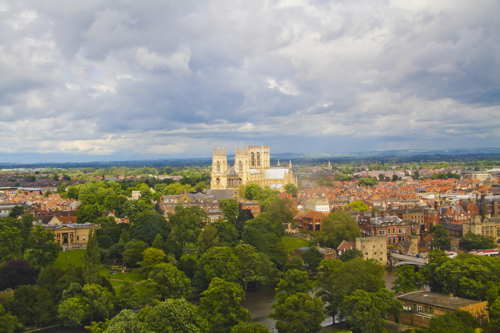}
\includegraphics[height=1.44cm]{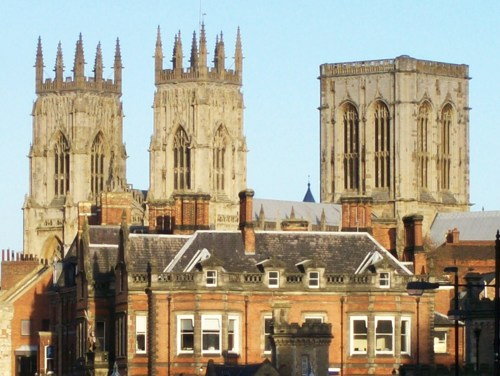}}
{\leftskip=0.1pt   \footnotesize{``York Minister across the roof-tops of York, UK.''} } 
\\
\jsubfig{\fbox{\includegraphics[height=1.45cm]{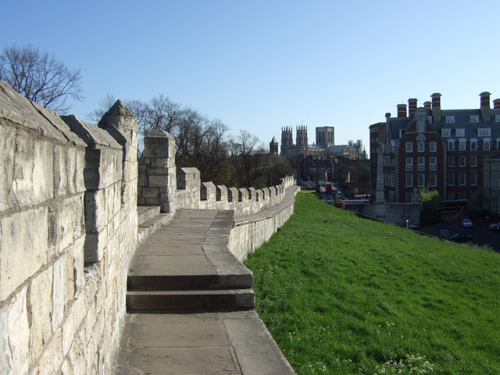}} \hfill \includegraphics[height=1.45cm]{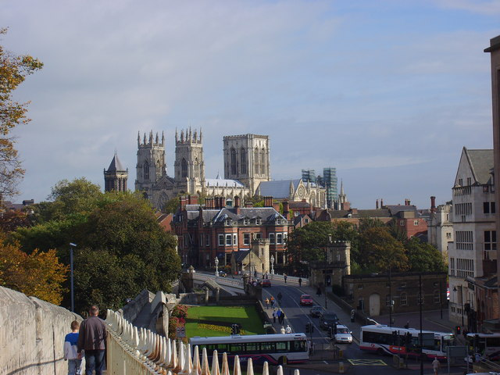}
\includegraphics[height=1.45cm]{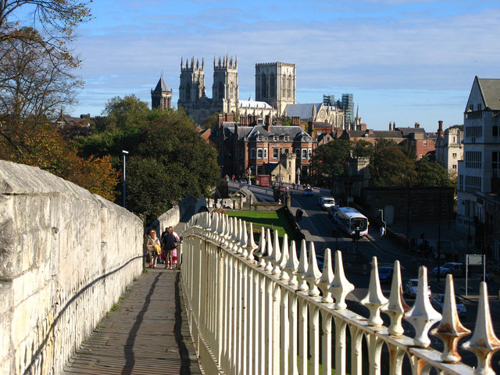}
\includegraphics[height=1.45cm]{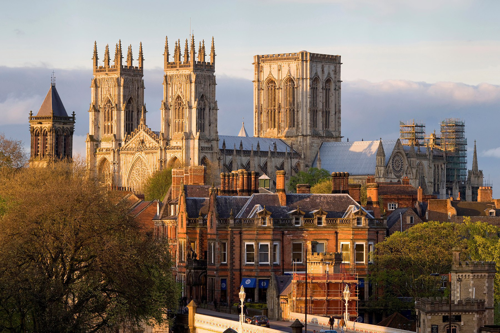}}
{\leftskip=0.1pt   \footnotesize{``York city walls pathway Looking towards Lendal Bridge and the Minster beyond.''} } 
\\
\jsubfig{\fbox{\includegraphics[height=1.78cm]{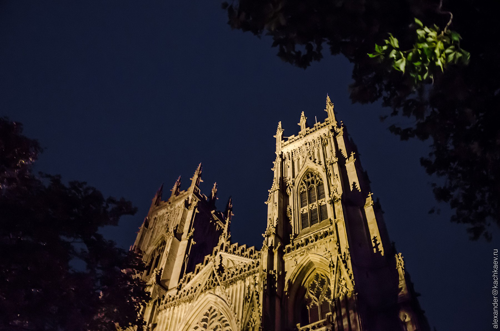}} \hfill \includegraphics[height=1.78cm]{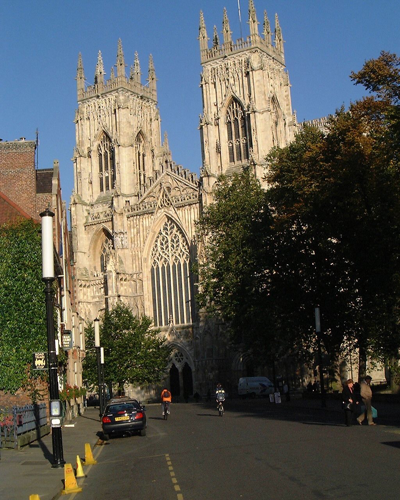}
\includegraphics[height=1.78cm]{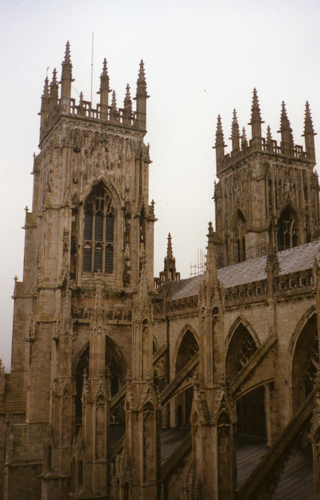}
\includegraphics[height=1.78cm]{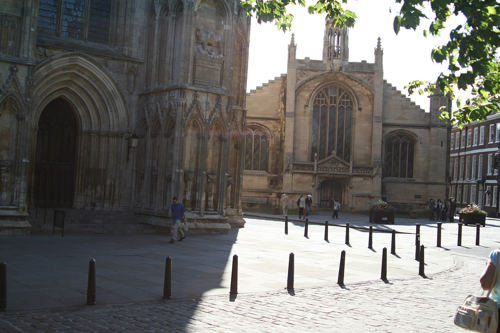}}
{\leftskip=0.1pt   \footnotesize{``York Minster at night (2012)''} } 
\vspace{-8pt}
\caption{Retrieving images from captions of The Cathedral of Our Lady and York Minister (landmarks not seen during training). Above we show the top three retrievals next to the reference image (left with black border) that corresponds to the query caption beneath. Note that this query image is not seen by the network---just the caption---and so we only show this image for reference. In the bottom row, we demonstrate that our model is less sensitive to appearance-based descriptions---in this case, the retrieved images are not captured ``at night''. This can be attributed to our 3D augmentations, which are unaware of appearance changes (thus allowing to focus on part-based \emph{scene} semantics instead).
}
\label{fig:retrieval_supp}
\end{figure}

\begin{figure*}
    \centering
    \vspace{-5pt}
    \rotatebox{90}{\whitetxt{``portal''} }
\jsubfig{\includegraphics[height=2.3cm]{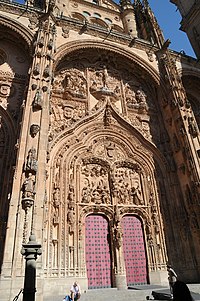}
\hspace{1pt}}{}
\jsubfig{\includegraphics[height=2.3cm]{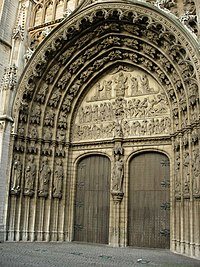}
\hspace{1pt}}{}
\jsubfig{\includegraphics[height=2.3cm]{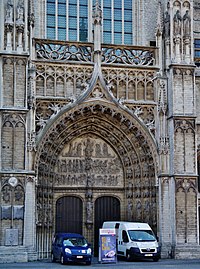}
\hspace{1pt}}{}
\jsubfig{\includegraphics[height=2.3cm]{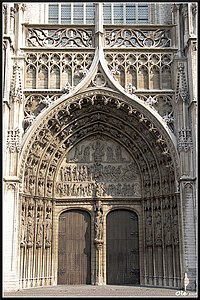}
\hspace{1pt}}{}
\jsubfig{\includegraphics[height=2.3cm]{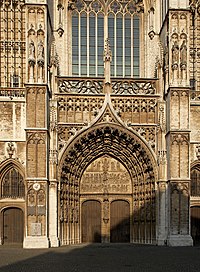}
\hspace{1pt}}{}
\jsubfig{\includegraphics[height=2.3cm]{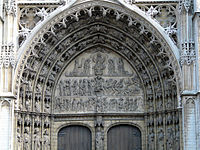}
\hspace{1pt}}{}
\jsubfig{\includegraphics[height=2.3cm]{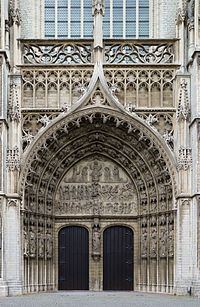}
\hspace{1pt}}{}
\jsubfig{\includegraphics[height=2.3cm]{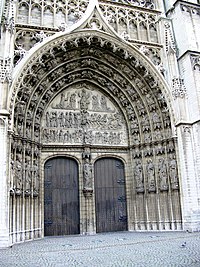}
\hspace{1pt}}{}
\\ \vspace{1pt}
\rotatebox{90}{\whitetxt{sss} ``portal''}
\jsubfig{\includegraphics[height=2.3cm]{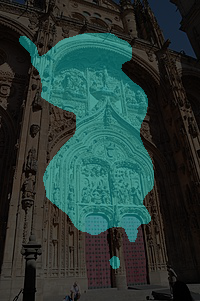}
\hspace{1pt}}{}
\jsubfig{\includegraphics[height=2.3cm]{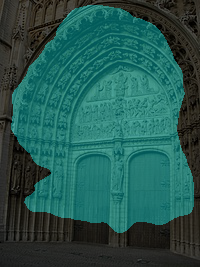}
\hspace{1pt}}{}
\jsubfig{\includegraphics[height=2.3cm]{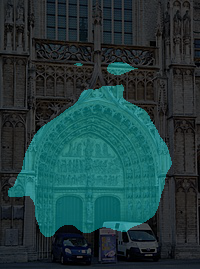}
\hspace{1pt}}{}
\jsubfig{\includegraphics[height=2.3cm]{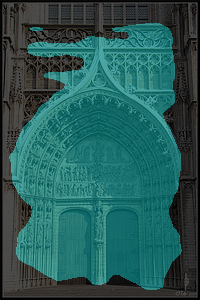}
\hspace{1pt}}{}
\jsubfig{\includegraphics[height=2.3cm]{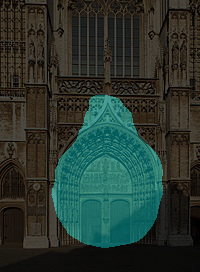}
\hspace{1pt}}{}
\jsubfig{\includegraphics[height=2.3cm]{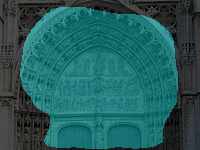}
\hspace{1pt}}{}
\jsubfig{\includegraphics[height=2.3cm]{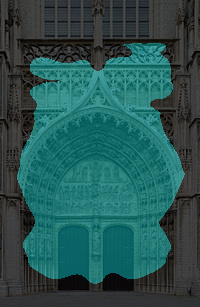}
\hspace{1pt}}{}
\jsubfig{\includegraphics[height=2.3cm]{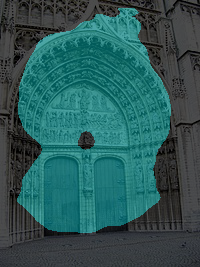}
\hspace{1pt}}{}
\\ \vspace{1pt} %
\hspace{0.5pt}
\rotatebox{90}{\whitetxt{``choir''} }
\jsubfig{\includegraphics[height=2.26cm]{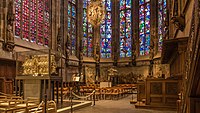}
\hspace{1pt}}{}
\jsubfig{\includegraphics[height=2.26cm]{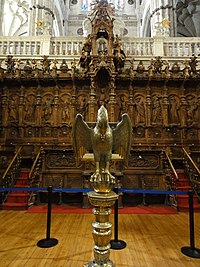}
\hspace{1pt}}{}
\jsubfig{\includegraphics[height=2.26cm]{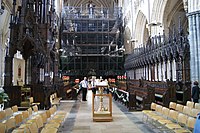}
\hspace{1pt}}{}
\jsubfig{\includegraphics[height=2.26cm]{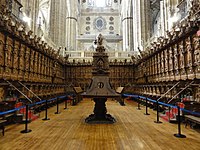}
\hspace{1pt}}{}
\jsubfig{\includegraphics[height=2.26cm]{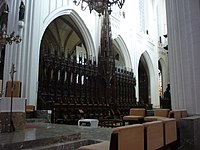}
\hspace{1pt}}{}
\\ \vspace{1pt}
\rotatebox{90}{\whitetxt{sss} ``choir''}
\jsubfig{\includegraphics[height=2.26cm]{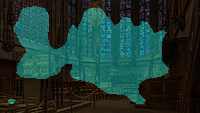}
\hspace{1pt}}{}
\jsubfig{\includegraphics[height=2.26cm]{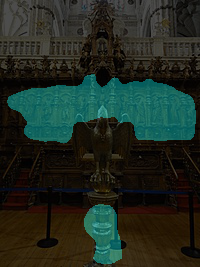}
\hspace{1pt}}{}
\jsubfig{\includegraphics[height=2.26cm]{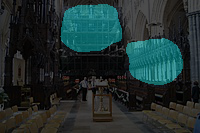}
\hspace{1pt}}{}
\jsubfig{\includegraphics[height=2.26cm]{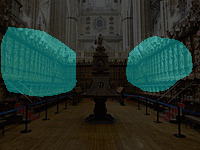}
\hspace{1pt}}{}
\jsubfig{\includegraphics[height=2.26cm]{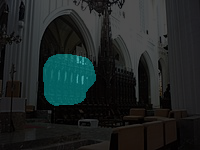}
\hspace{1pt}}{}
\\
\vspace{1pt} %
\rotatebox{90}{\whitetxt{``tower''} }
\jsubfig{\includegraphics[height=2.23cm]{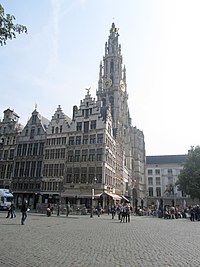}
\hspace{1pt}}{}
\jsubfig{\includegraphics[height=2.23cm]{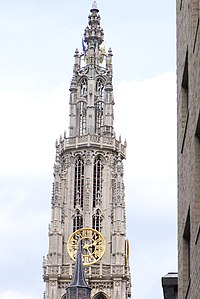}
\hspace{1pt}}{}
\jsubfig{\includegraphics[height=2.23cm]{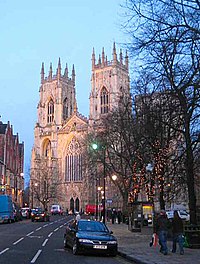}
\hspace{1pt}}{}
\jsubfig{\includegraphics[height=2.23cm]{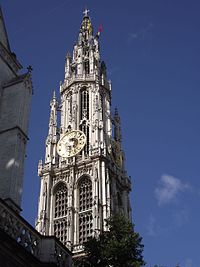}
\hspace{1pt}}{}
\jsubfig{\includegraphics[height=2.23cm]{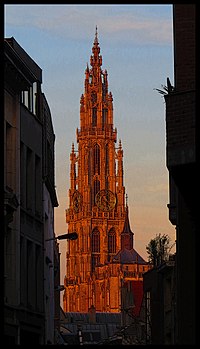}
\hspace{1pt}}{}
\jsubfig{\includegraphics[height=2.23cm]{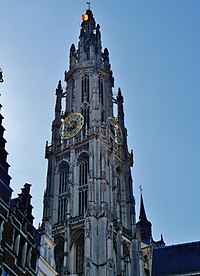}
\hspace{1pt}}{}
\jsubfig{\includegraphics[height=2.23cm]{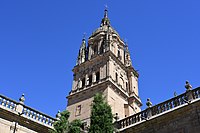}
\hspace{1pt}}{}
\jsubfig{\includegraphics[height=2.23cm]{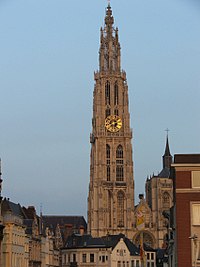}
\hspace{1pt}}{}
\\ \vspace{1pt} %
\rotatebox{90}{ \whitetxt{sas}``tower''}
\jsubfig{\includegraphics[height=2.23cm]{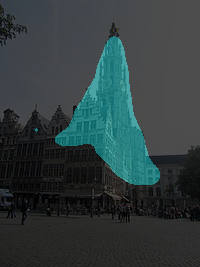}
\hspace{1pt}}{}
\jsubfig{\includegraphics[height=2.23cm]{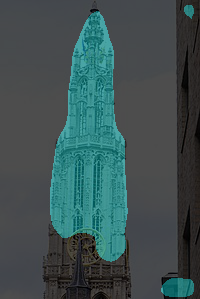}
\hspace{1pt}}{}
\jsubfig{\includegraphics[height=2.23cm]{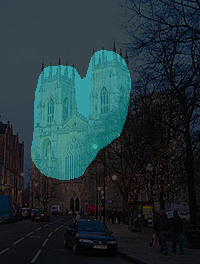}
\hspace{1pt}}{}
\jsubfig{\includegraphics[height=2.23cm]{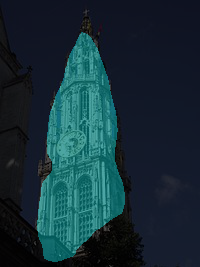}
\hspace{1pt}}{}
\jsubfig{\includegraphics[height=2.23cm]{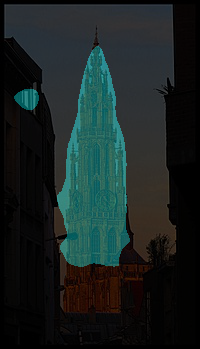}
\hspace{1pt}}{}
\jsubfig{\includegraphics[height=2.23cm]{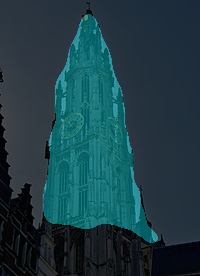}
\hspace{1pt}}{}
\jsubfig{\includegraphics[height=2.23cm]{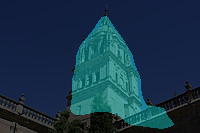}
\hspace{1pt}}{}
\jsubfig{\includegraphics[height=2.23cm]{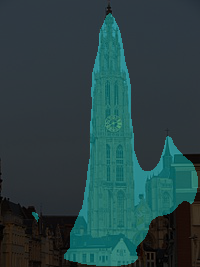}
\hspace{1pt}}{}
\\
\vspace{1pt} %
\rotatebox{90}{\whitetxt{``chapel''} }
\jsubfig{\includegraphics[height=2.26cm]{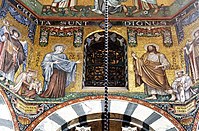}
\hspace{1pt}}{}
\jsubfig{\includegraphics[height=2.26cm]{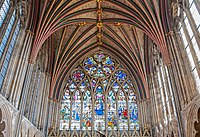}
\hspace{1pt}}{}
\jsubfig{\includegraphics[height=2.26cm]{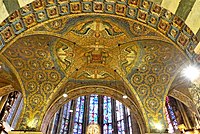}
\hspace{1pt}}{}
\jsubfig{\includegraphics[height=2.26cm]{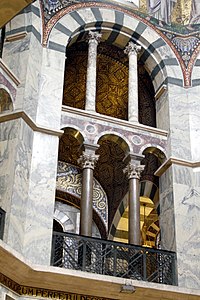}
\hspace{1pt}}{}
\jsubfig{\includegraphics[height=2.26cm]{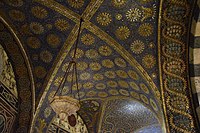}
\hspace{1pt}}{}
\\ \vspace{1pt} %
\rotatebox{90}{\whitetxt{ss} ``chapel''}
\jsubfig{\includegraphics[height=2.26cm]{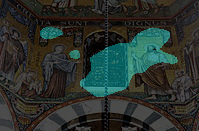}
\hspace{1pt}}{}
\jsubfig{\includegraphics[height=2.26cm]{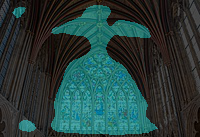}
\hspace{1pt}}{}
\jsubfig{\includegraphics[height=2.26cm]{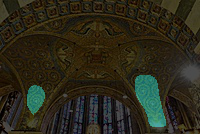}
\hspace{1pt}}{}
\jsubfig{\includegraphics[height=2.26cm]{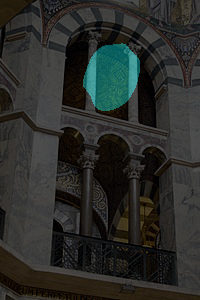}
\hspace{1pt}}{}
\jsubfig{\includegraphics[height=2.26cm]{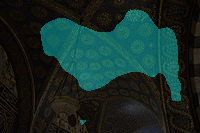}
\hspace{1pt}}{}
\\
\vspace{1pt} %
\rotatebox{90}{\whitetxt{``nave''} }
\jsubfig{\includegraphics[height=2.17cm]{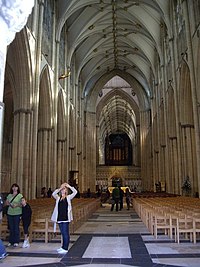}
\hspace{1pt}}{}
\jsubfig{\includegraphics[height=2.17cm]{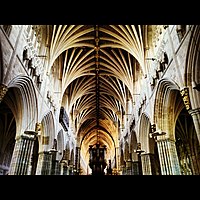}
\hspace{1pt}}{}
\jsubfig{\includegraphics[height=2.17cm]{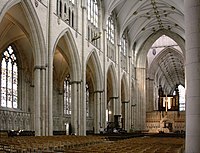}
\hspace{1pt}}{}
\jsubfig{\includegraphics[height=2.17cm]{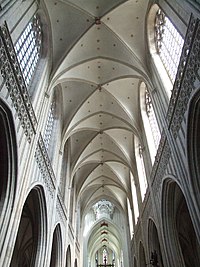}
\hspace{1pt}}{}
\jsubfig{\includegraphics[height=2.17cm]{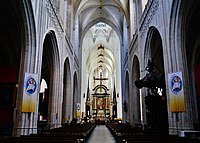}
\hspace{1pt}}{}
\jsubfig{\includegraphics[height=2.17cm]{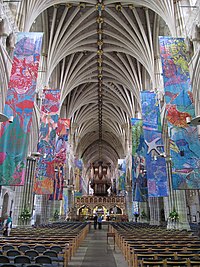}
\hspace{1pt}}{}
\jsubfig{\includegraphics[height=2.17cm]{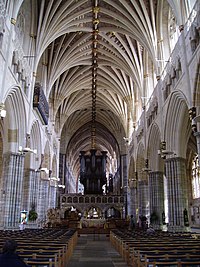}
\hspace{1pt}}{}
\\ \vspace{1pt} 
\rotatebox{90}{\whitetxt{sss} ``nave''}
\jsubfig{\includegraphics[height=2.17cm]{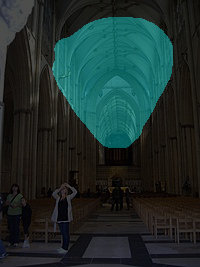}
\hspace{1pt}}{}
\jsubfig{\includegraphics[height=2.17cm]{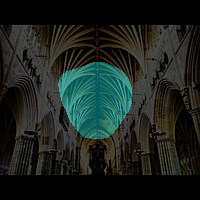}
\hspace{1pt}}{}
\jsubfig{\includegraphics[height=2.17cm]{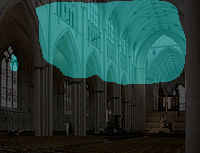}
\hspace{1pt}}{}
\jsubfig{\includegraphics[height=2.17cm]{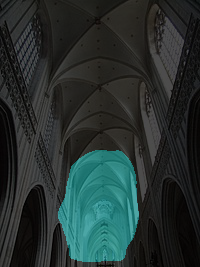}
\hspace{1pt}}{}
\jsubfig{\includegraphics[height=2.17cm]{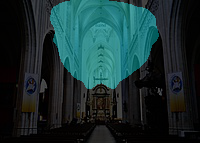}
\hspace{1pt}}{}
\jsubfig{\includegraphics[height=2.17cm]{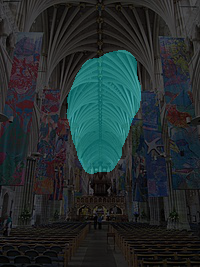}
\hspace{1pt}}{}
\jsubfig{\includegraphics[height=2.17cm]{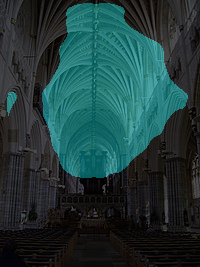}
\hspace{1pt}}{}
\vspace{-8pt}
\caption{\textbf{2D Segmentations on correctly classified unseen images,} segmented as ``portal'', ``choir'', ``tower'', ``chapel'' and ``nave''. }
    \label{fig:seg_supp}
\end{figure*}
\begin{figure*}
    \centering
    \hspace{1pt}
    \rotatebox{90}{\whitetxt{``facade''} }
\jsubfig{\includegraphics[height=2.05cm]{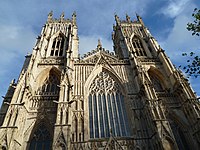}
\hspace{1pt}}{}
\jsubfig{\includegraphics[height=2.05cm]{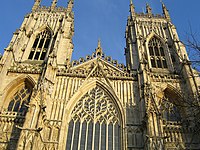}
\hspace{1pt}}{}
\jsubfig{\includegraphics[height=2.05cm]{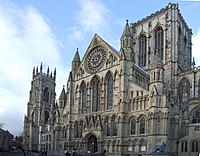}
\hspace{1pt}}{}
\jsubfig{\includegraphics[height=2.05cm]{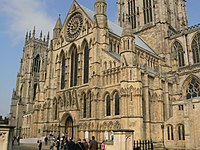}
\hspace{1pt}}{}
\jsubfig{\includegraphics[height=2.05cm]{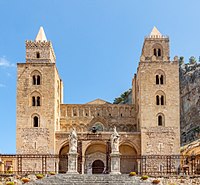}
\hspace{1pt}}{}
\jsubfig{\includegraphics[height=2.05cm]{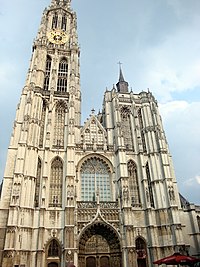}
\hspace{1pt}}{}
\\ \vspace{1pt}
\hspace{2.5pt}
\rotatebox{90}{ \whitetxt{ss}``facade''}
\jsubfig{\includegraphics[height=2.05cm]{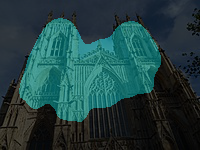}
\hspace{1pt}}{}
\jsubfig{\includegraphics[height=2.05cm]{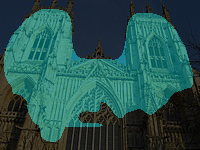}
\hspace{1pt}}{}
\jsubfig{\includegraphics[height=2.05cm]{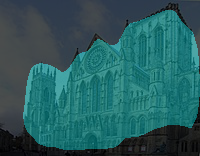}
\hspace{1pt}}{}
\jsubfig{\includegraphics[height=2.05cm]{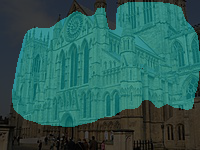}
\hspace{1pt}}{}
\jsubfig{\includegraphics[height=2.05cm]{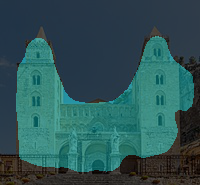}
\hspace{1pt}}{}
\jsubfig{\includegraphics[height=2.05cm]{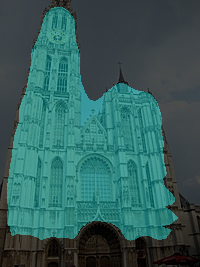}
\hspace{1pt}}{}
\vspace{1pt} \\ 
\hspace{0.5pt}
\rotatebox{90}{\whitetxt{``altar''} }
\jsubfig{\includegraphics[height=2.16cm]{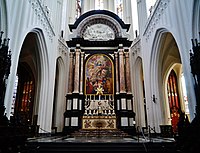}
\hspace{1pt}}{}
\jsubfig{\includegraphics[height=2.16cm]{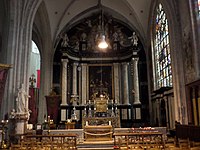}
\hspace{1pt}}{}
\jsubfig{\includegraphics[height=2.16cm]{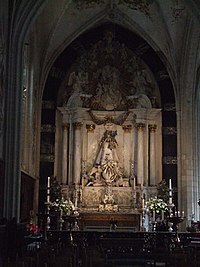}
\hspace{1pt}}{}
\jsubfig{\includegraphics[height=2.16cm]{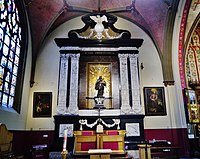}
\hspace{1pt}}{}
\jsubfig{\includegraphics[height=2.16cm]{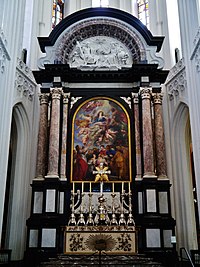}
\hspace{1pt}}{}
\jsubfig{\includegraphics[height=2.16cm]{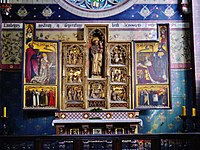}
\hspace{1pt}}{}
\\ 
\vspace{1pt} 
\rotatebox{90}{\whitetxt{sss} ``altar''}
\jsubfig{\includegraphics[height=2.16cm]{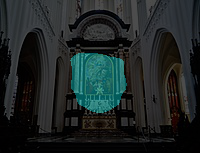}
\hspace{1pt}}{}
\jsubfig{\includegraphics[height=2.16cm]{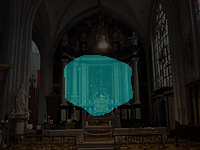}
\hspace{1pt}}{}
\jsubfig{\includegraphics[height=2.16cm]{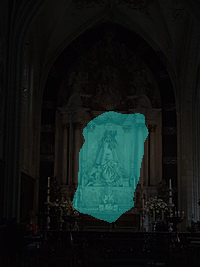}
\hspace{1pt}}{}
\jsubfig{\includegraphics[height=2.16cm]{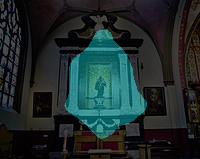}
\hspace{1pt}}{}
\jsubfig{\includegraphics[height=2.16cm]{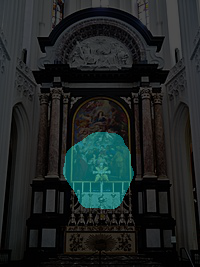}
\hspace{1pt}}{}
\jsubfig{\includegraphics[height=2.16cm]{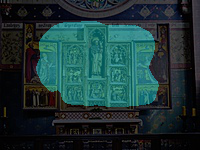}}{}
\\
\vspace{1pt} %
\rotatebox{90}{\whitetxt{``organ''} }
\jsubfig{\includegraphics[height=2.07cm]{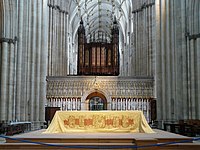}
\hspace{1pt}}{}
\jsubfig{\includegraphics[height=2.07cm]{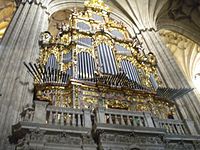}
\hspace{1pt}}{}
\jsubfig{\includegraphics[height=2.07cm]{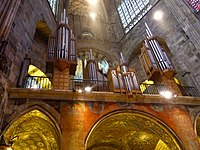}
\hspace{1pt}}{}
\jsubfig{\includegraphics[height=2.07cm]{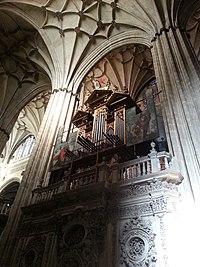}
\hspace{1pt}}{}
\jsubfig{\includegraphics[height=2.07cm]{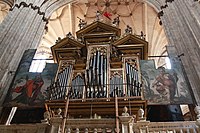}
\hspace{1pt}}{}
\jsubfig{\includegraphics[height=2.07cm]{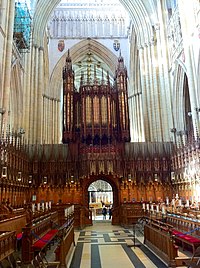}
\hspace{1pt}}{}
\\ \vspace{1pt}%
\rotatebox{90}{ \whitetxt{sss}``organ''}
\jsubfig{\includegraphics[height=2.07cm]{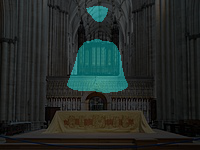}
\hspace{1pt}}{}
\jsubfig{\includegraphics[height=2.07cm]{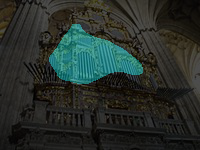}
\hspace{1pt}}{}
\jsubfig{\includegraphics[height=2.07cm]{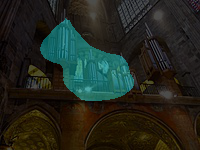}
\hspace{1pt}}{}
\jsubfig{\includegraphics[height=2.07cm]{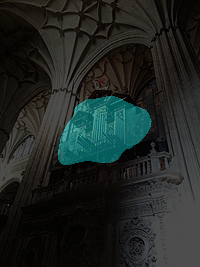}
\hspace{1pt}}{}
\jsubfig{\includegraphics[height=2.07cm]{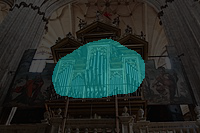}
\hspace{1pt}}{}
\jsubfig{\includegraphics[height=2.07cm]{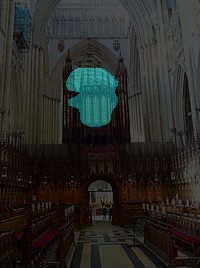}
\hspace{1pt}}{}
\\
\vspace{1pt} %
\hspace{0.5pt}
\rotatebox{90}{\whitetxt{``statue''} }
\jsubfig{\includegraphics[height=2.05cm]{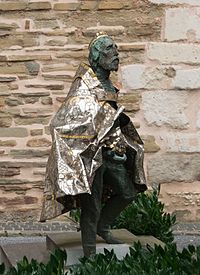}
\hspace{1pt}}{}
\jsubfig{\includegraphics[height=2.05cm]{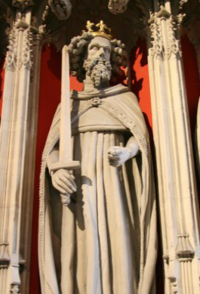}
\hspace{1pt}}{}
\jsubfig{\includegraphics[height=2.05cm]{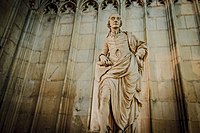}
\hspace{1pt}}{}
\jsubfig{\includegraphics[height=2.05cm]{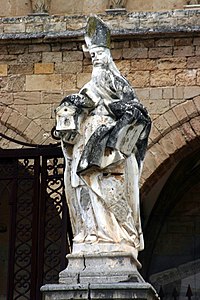}
\hspace{1pt}}{}
\jsubfig{\includegraphics[height=2.05cm]{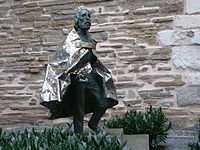}
\hspace{1pt}}{}
\jsubfig{\includegraphics[height=2.05cm]{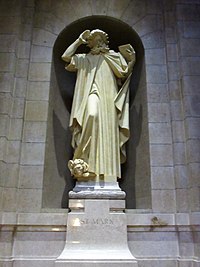}
\hspace{1pt}}{}
\jsubfig{\includegraphics[height=2.05cm]{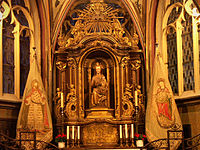}
\hspace{1pt}}{}
\\ \vspace{1pt} \hspace{0.5pt}
\rotatebox{90}{ \whitetxt{ss}``statue''}
\jsubfig{\includegraphics[height=2.05cm]{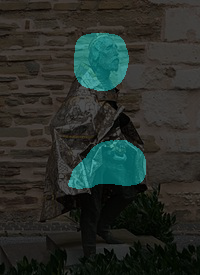}
\hspace{1pt}}{}
\jsubfig{\includegraphics[height=2.05cm]{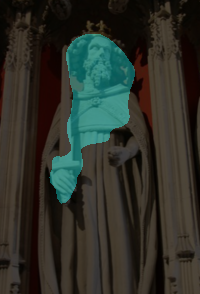}
\hspace{1pt}}{}
\jsubfig{\includegraphics[height=2.05cm]{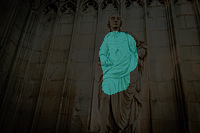}
\hspace{1pt}}{}
\jsubfig{\includegraphics[height=2.05cm]{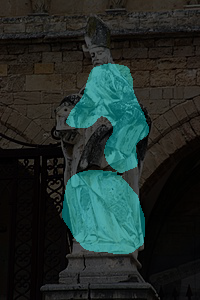}
\hspace{1pt}}{}
\jsubfig{\includegraphics[height=2.05cm]{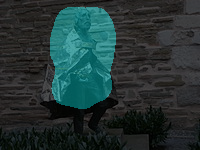}
\hspace{1pt}}{}
\jsubfig{\includegraphics[height=2.05cm]{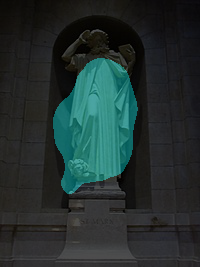}
\hspace{1pt}}{}
\jsubfig{\includegraphics[height=2.05cm]{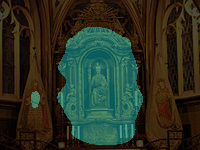}
\hspace{1pt}}{}
\\
\vspace{1pt} %
\rotatebox{90}{\whitetxt{``window''} }
\jsubfig{\includegraphics[height=2.08cm]{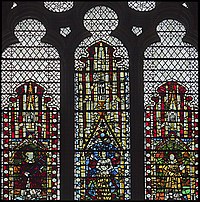}
\hspace{1pt}}{}
\jsubfig{\includegraphics[height=2.08cm]{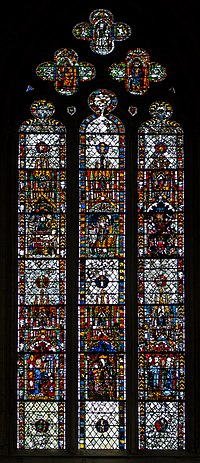}
\hspace{1pt}}{}
\jsubfig{\includegraphics[height=2.08cm]{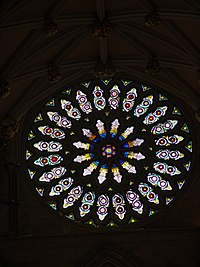}
\hspace{1pt}}{}
\jsubfig{\includegraphics[height=2.08cm]{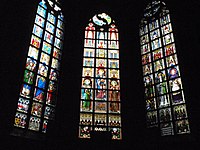}
\hspace{1pt}}{}
\jsubfig{\includegraphics[height=2.08cm]{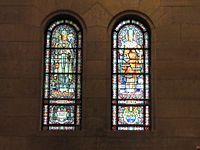}
\hspace{1pt}}{}
\jsubfig{\includegraphics[height=2.08cm]{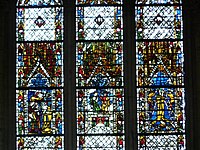}
\hspace{1pt}}{}
\jsubfig{\includegraphics[height=2.08cm]{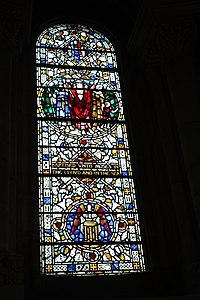}
\hspace{1pt}}{}
\\ \vspace{1pt} %
\rotatebox{90}{\whitetxt{ss} ``window''}
\jsubfig{\includegraphics[height=2.08cm]{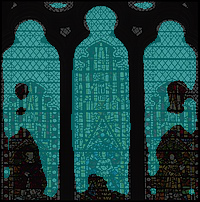}
\hspace{1pt}}{}
\jsubfig{\includegraphics[height=2.08cm]{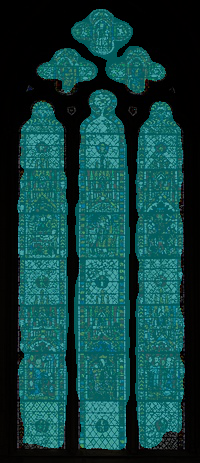}
\hspace{1pt}}{}
\jsubfig{\includegraphics[height=2.08cm]{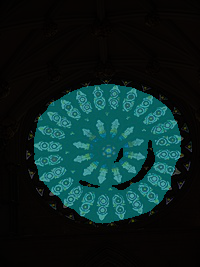}
\hspace{1pt}}{}
\jsubfig{\includegraphics[height=2.08cm]{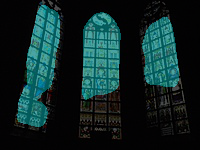}
\hspace{1pt}}{}
\jsubfig{\includegraphics[height=2.08cm]{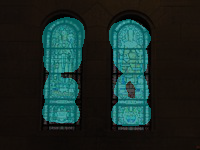}
\hspace{1pt}}{}
\jsubfig{\includegraphics[height=2.08cm]{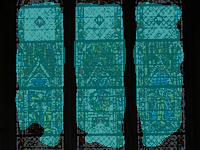}
\hspace{1pt}}{}
\jsubfig{\includegraphics[height=2.08cm]{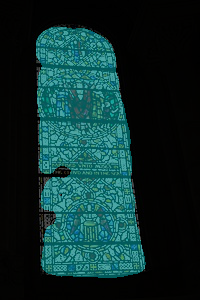}
\hspace{1pt}}{}
\caption{\textbf{2D Segmentations on correctly classified unseen images.} Hightlighted pixels are segmented as ``facade'', ``altar'', ``organ'', ``statue'' and ``window''.}
    \label{fig:seg_supp2}
\end{figure*}
\definecolor{facade}{RGB}{0, 0, 255}%
\definecolor{window}{RGB}{255,104, 0}%
\definecolor{chapel}{RGB}{0,153, 0}%
\definecolor{organ}{RGB}{255, 0, 0} %
\definecolor{nave}{RGB}{101,0,204}%
\definecolor{tower}{RGB}{102, 51, 0}%
\definecolor{choir}{RGB}{255,51, 255}%
\definecolor{portal}{RGB}{255,153, 153}%
\definecolor{altar}{RGB}{236, 227, 102} %
\definecolor{statue}{RGB}{57,218,250}%

\begin{figure*}
    \centering
\jsubfig{\includegraphics[height=3.5cm]{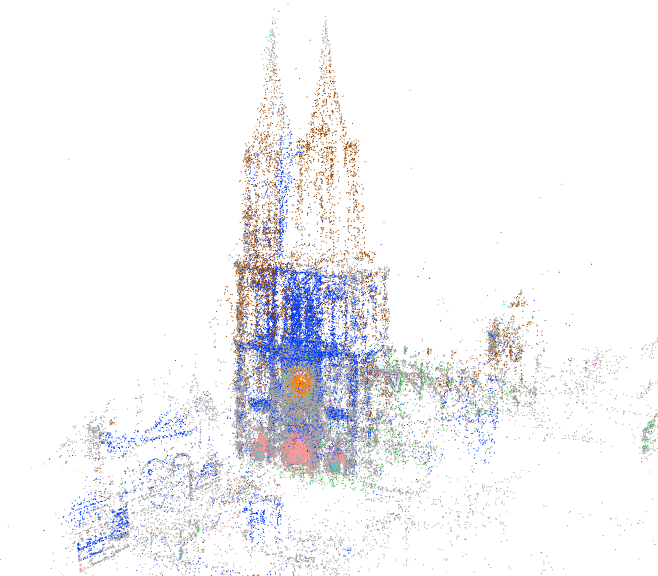}}{Notre-Dame de Strasbourg (exterior)}
\jsubfig{\includegraphics[height=3.5cm]{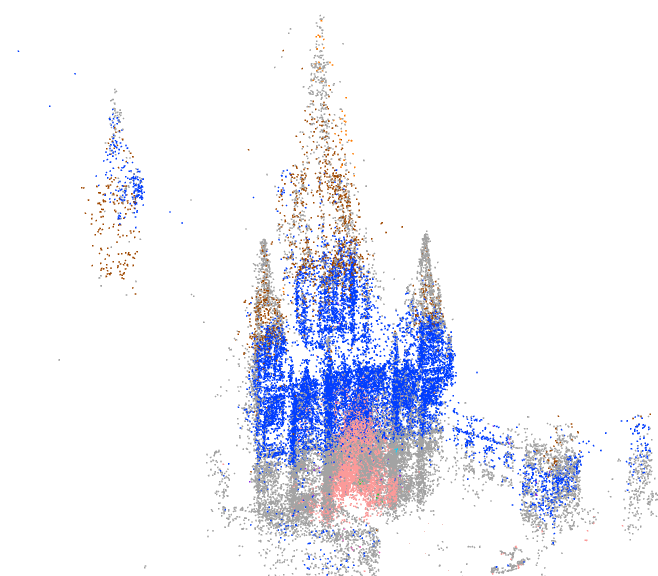}}{Cathedral of Barcelona (exterior)}
\jsubfig{\includegraphics[height=3.5cm]{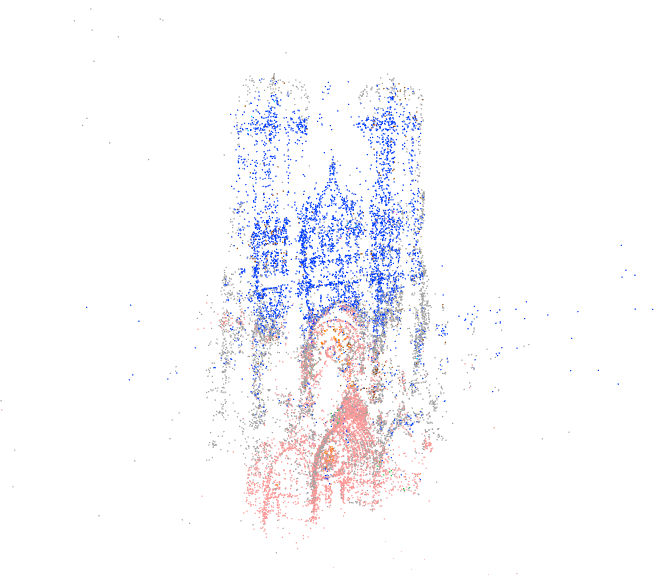}}{Notre-Dame de Reims  (exterior)}
\jsubfig{\includegraphics[height=3.5cm]{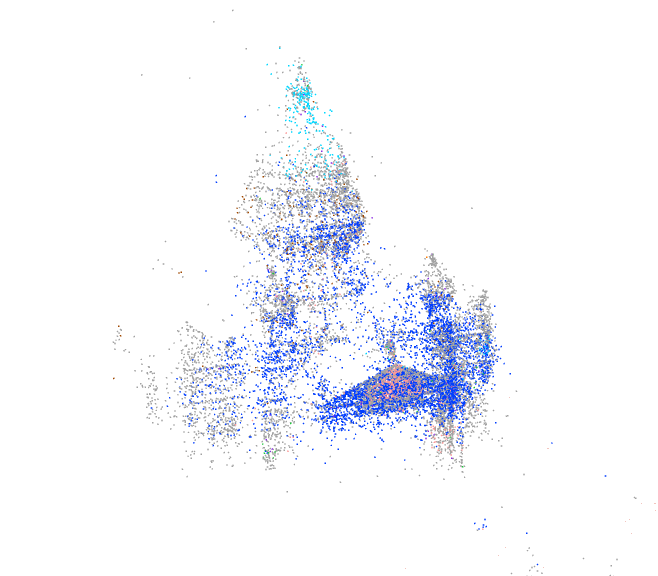}}{Saint Isaac's Cathedral  (exterior)}
\jsubfig{\includegraphics[height=3.5cm]{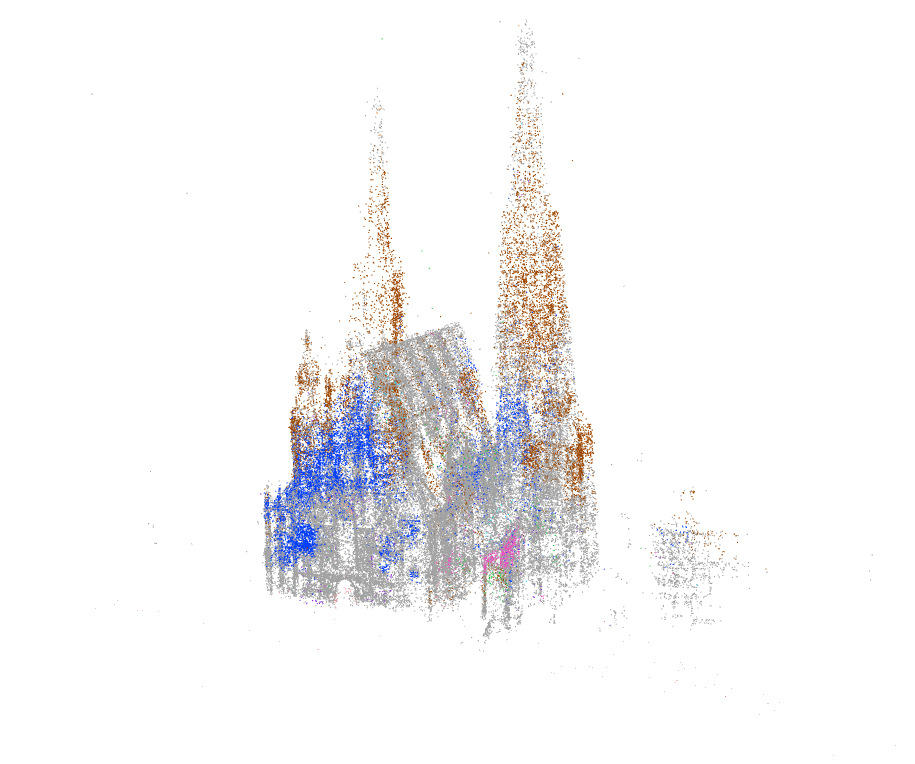}}{St. Stephen's Cathedral (exterior) }
\jsubfig{\includegraphics[height=3.5cm]{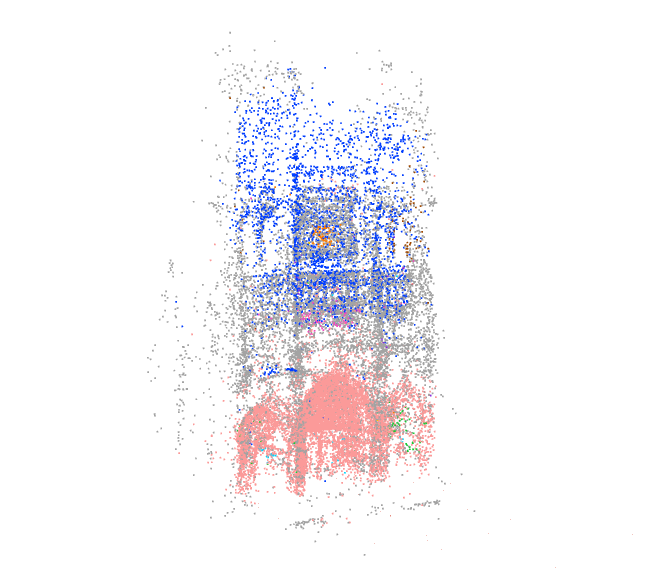}}{Amiens Cathedral  (exterior)}
\jsubfig{\includegraphics[height=3.5cm]{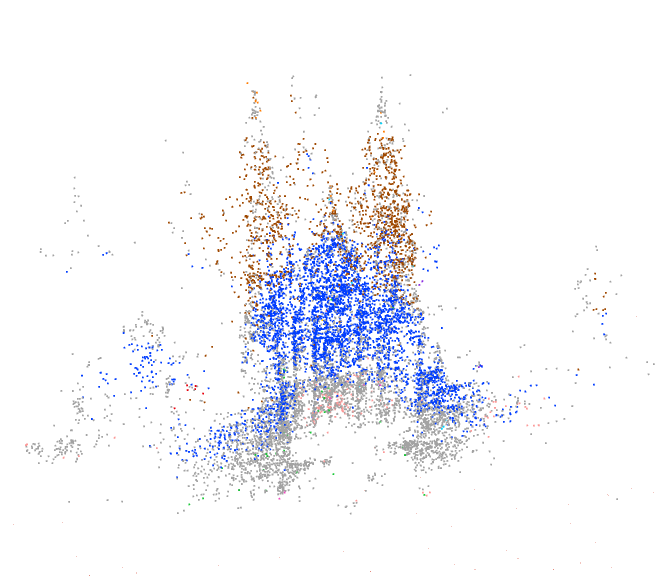}}{Santiago de Compostela Cathedral (exterior)}
\jsubfig{\includegraphics[height=3.5cm]{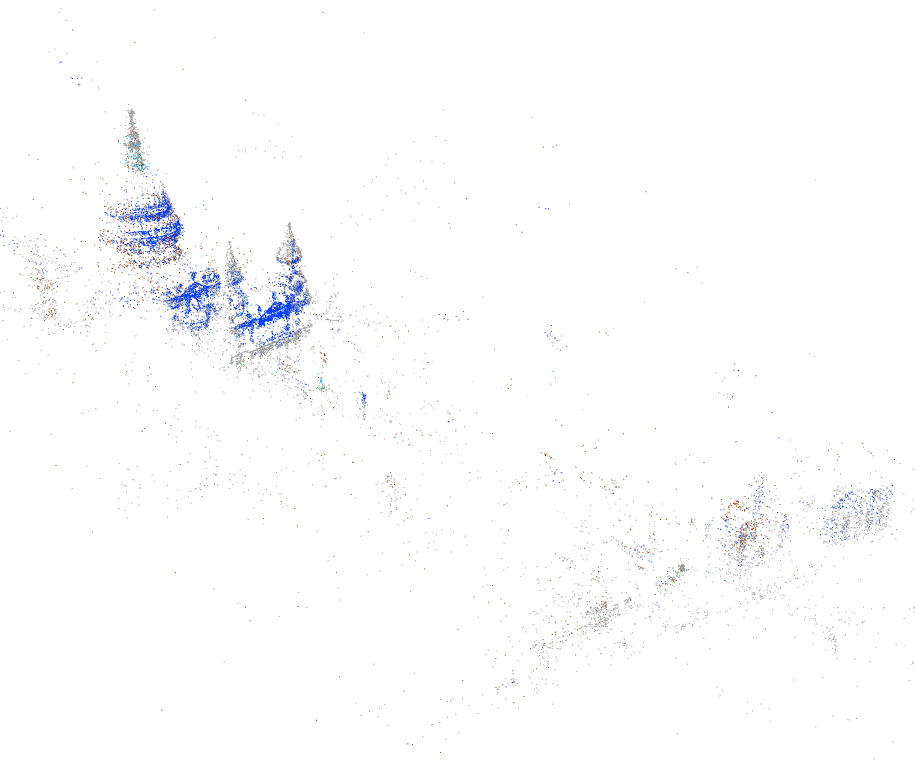}}{St Paul's Cathedral (exterior)}
\jsubfig{\includegraphics[height=3.5cm]{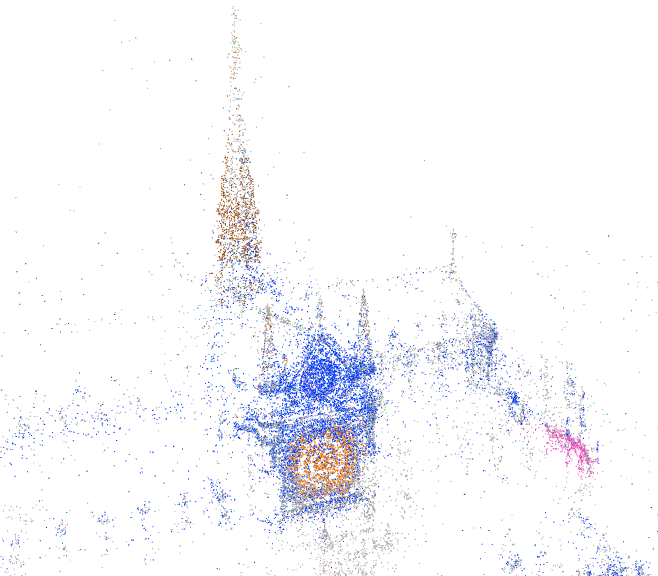}}{Notre-Dame de Paris (exterior 1)}
\jsubfig{\includegraphics[height=3.5cm]{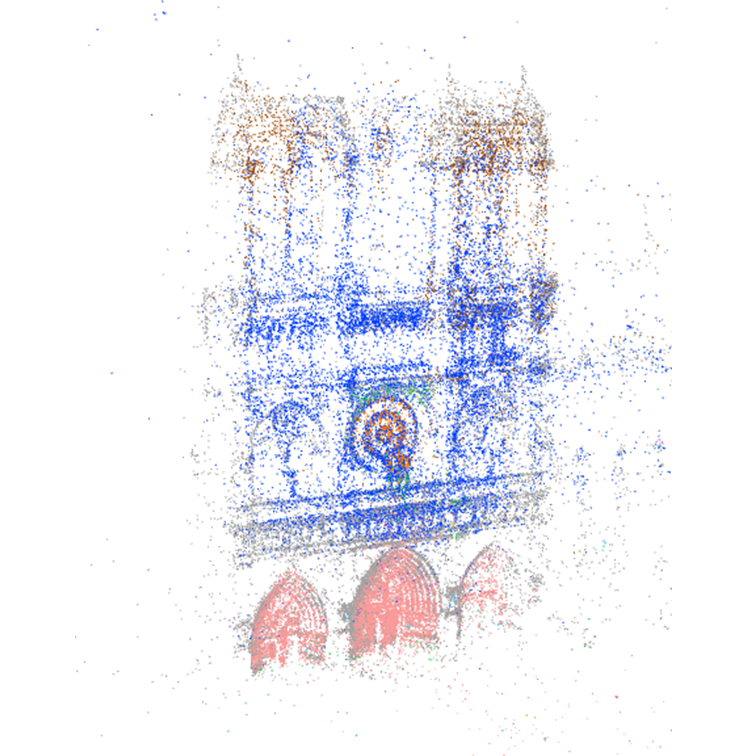}}{Notre-Dame de Paris (exterior 2)}
\jsubfig{\includegraphics[height=3.5cm]{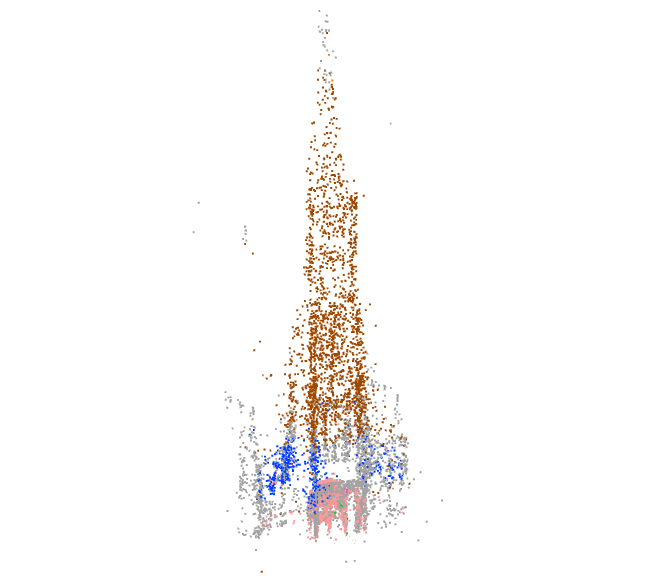}}{Ulm Minster (exterior)}
\jsubfig{\includegraphics[height=3.5cm]{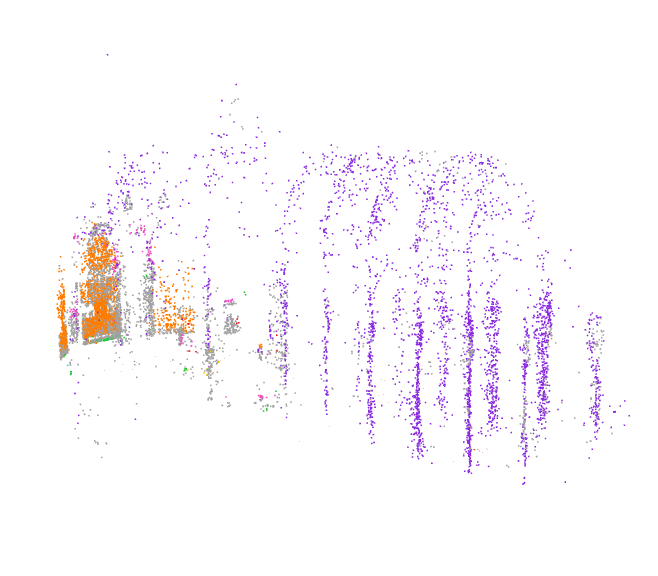}}{Duomo di Milano (interior)}
\jsubfig{\includegraphics[height=3.5cm]{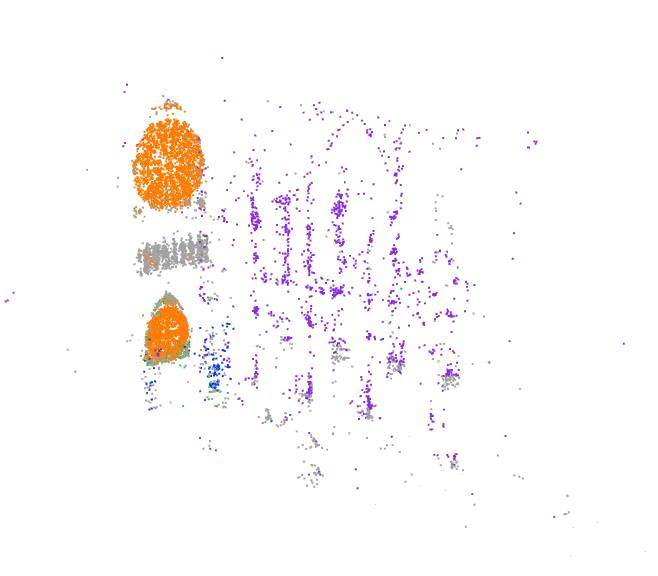}}{Notre-Dame de Reims (interior)}
\jsubfig{\includegraphics[height=3.5cm]{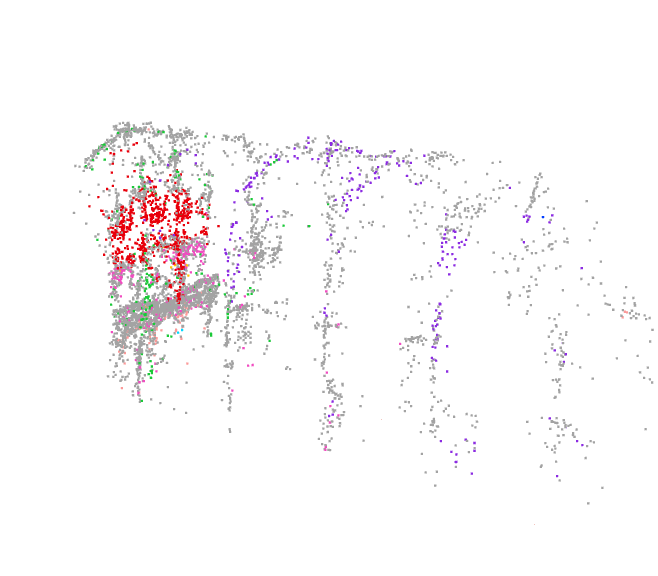}}{Cathédrale Saint-André de Bordeaux (interior)}
\jsubfig{\includegraphics[height=3.5cm]{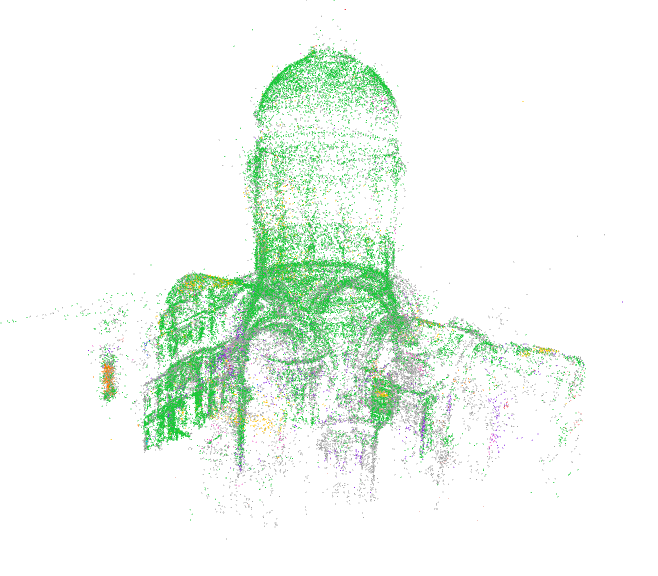}}{Saint Isaac's Cathedral (interior)}
\jsubfig{\includegraphics[height=3.5cm]{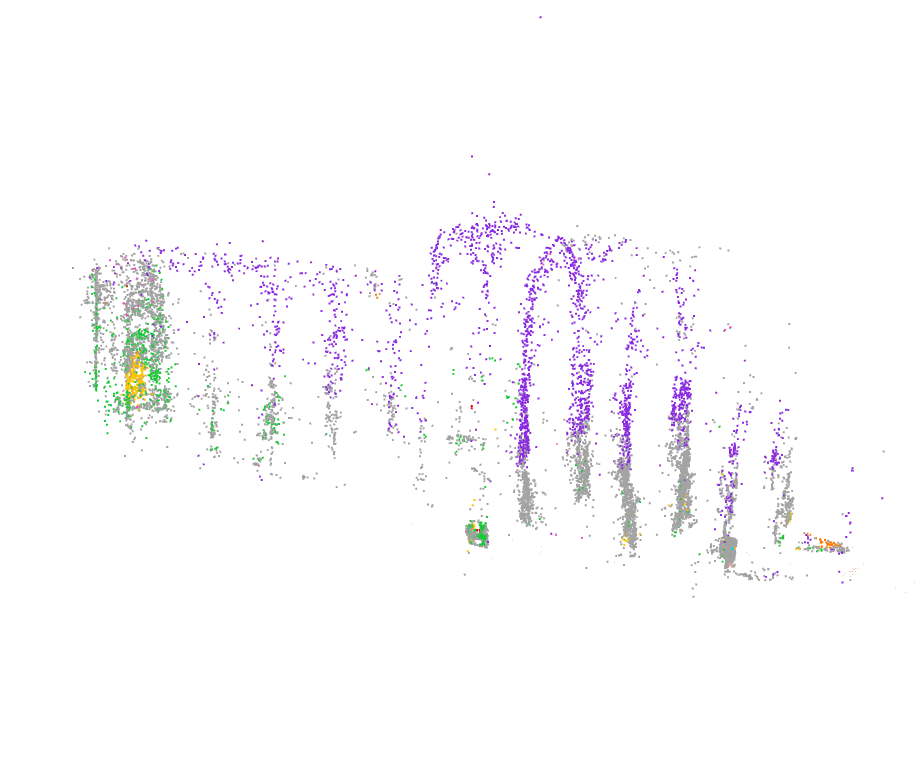}}{St. Stephen's Cathedral (interior)}
\jsubfig{\includegraphics[height=3.5cm]{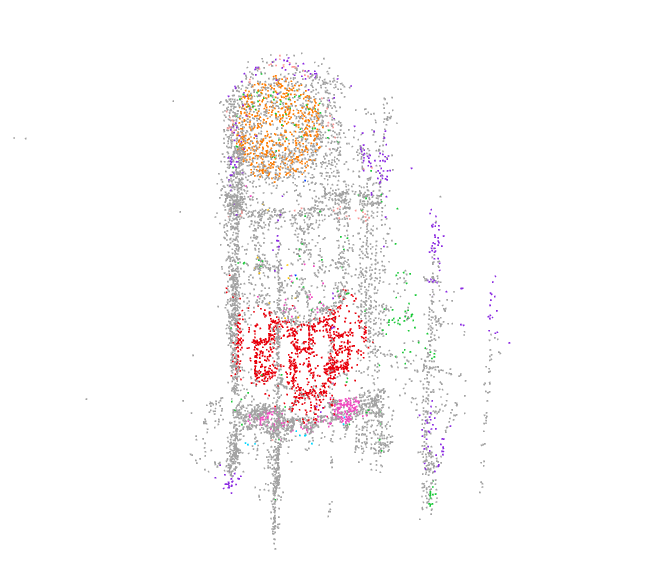}}{Amiens Cathedral (interior)}
\jsubfig{\includegraphics[height=3.5cm]{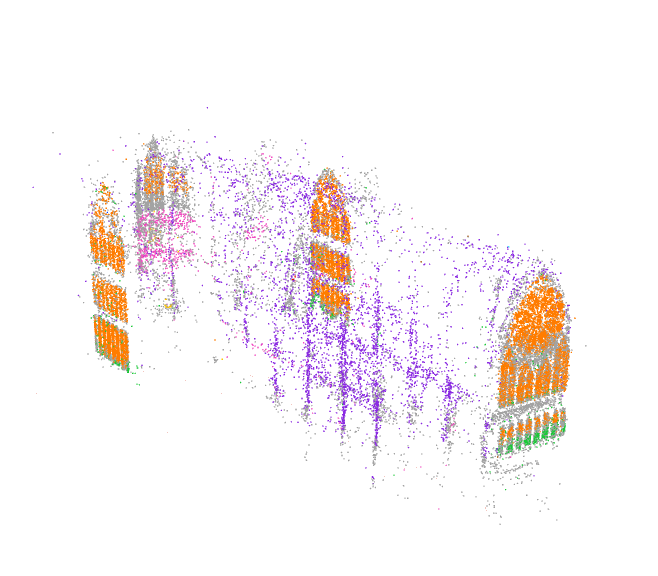}}{Metz Cathedral (interior)}
\jsubfig{\includegraphics[height=3.5cm]{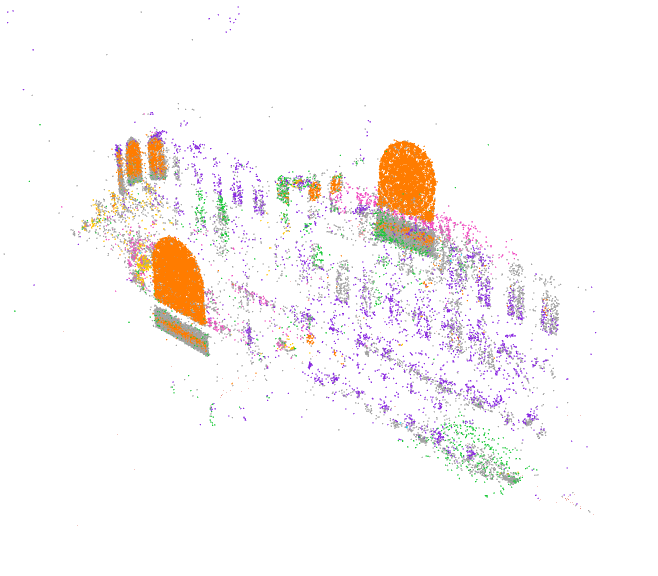}}{Notre-Dame de Paris (interior)}
\jsubfig{\includegraphics[height=3.5cm]{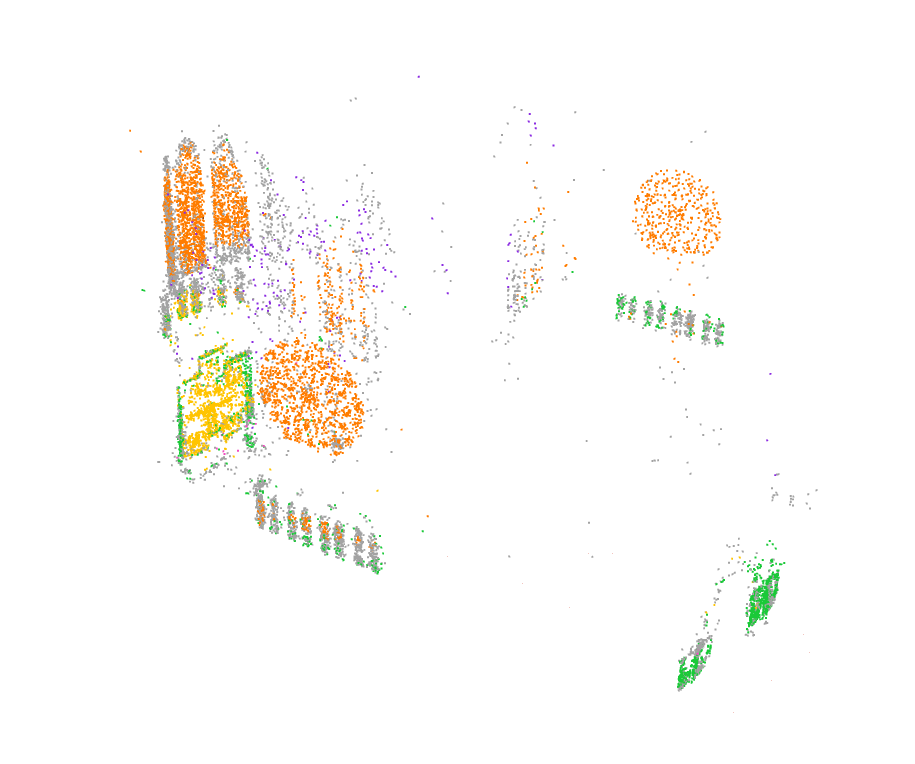}}{León Cathedral (interior)}

    \caption{Segmenting 3D reconstructions. Above we show segmentation results for landmarks seen during training. 3D points not associated with concepts are colored in gray. 
    Color legend of segmented points: \color{nave}{\emph{nave}}, \color{chapel}{\emph{chapel}}, \color{organ}{\emph{organ}}, \color{altar}{\emph{altar}},  \color{choir}{\emph{choir}}, \color{statue}{\emph{statue}}, \color{portal}{\emph{portal}}, \color{facade}{\emph{facade}},
    \color{tower}{\emph{tower}},
    \color{window}{\emph{window}}.}
    \label{fig:coloring_supp}
\end{figure*}

We show additional image segmentation results on test images from the WS-K test set in Figure \ref{fig:seg_supp} and Figure \ref{fig:seg_supp2}. As illustrated in the figures, the model is more successful with segmenting certain concepts, such as ``tower'', ``portal'' or ``window''. Some concepts, such as ``chapel'' yield noisier segmentation results. 
We show 3D segmentation results for landmarks in WS-K in Figure \ref{fig:coloring_supp}.

We visualize the learned features for two image pairs in Figure \ref{fig:features}. As the figure illustrates, distances in feature space are more semantically meaningful on the model trained with the 3D contrastive loss. For example, only pixels on the windows yield small distances using our model (left image pair). Our model is also more robust against large motion and appearance variations between the images.

We show additional caption-based image retrieval results in Figure \ref{fig:retrieval_supp}, mostly for images not-labeled with one of the semantic concepts we compute according to the method described in the main paper. As demonstrated in the figure, the model
can also align more generic semantic concepts to our images. However, as we perform 3D-augmentations, the model is less aware of appearance-based differences. For example, see the bottom row in the figure, where the retrieved images are not captured ``at night''.

\end{document}